\documentclass[10pt,twocolumn,letterpaper]{article}

\usepackage[pagenumbers]{paper} %

\usepackage{cuted} %

\usepackage{currfile} %

\usepackage{caption} %

\usepackage{multirow}
\usepackage{placeins}
\newcommand{\berke}[1]{} %
\newcommand{\ours}{CDS\xspace} %
\newcommand{\pizero}{$\pi_0$\xspace}
\newcommand{\pifive}{$\pi_{0.5}$\xspace}
\newcommand{\vect}[1]{\boldsymbol{#1}}
\DeclareMathOperator*{\argmax}{arg\,max}

\usepackage{colortbl}
\usepackage{tabularx}
\definecolor{firstcell}{HTML}{B9E0B5}  %
\definecolor{secondcell}{HTML}{E3F2E1} %
\newcommand{\fst}[1]{\cellcolor{firstcell}#1}
\newcommand{\snd}[1]{\cellcolor{secondcell}#1}
\newcommand{\hardsix}{Hard-62\xspace}

\definecolor{linkblue}{rgb}{0.21,0.49,0.74}
\usepackage[pagebackref,breaklinks,colorlinks,allcolors=linkblue]{hyperref}

\makeatletter
\patchcmd{\@maketitle}{\vspace*{24pt}}{\vspace*{18pt}}{}{\PackageWarning{main}{title spacing patch failed}}
\patchcmd{\@maketitle}{\vspace*{12pt}}{\vspace*{-8.5pt}}{}{\PackageWarning{main}{title spacing patch failed}}
\makeatother

\title{Unlocking Hidden Skills of Generalist Robot Policies by Steering Them in the Right Direction}

\author{Ahmet Berke G\"okmen \enspace Atakan Topcu \enspace Fiona Ryan \enspace Yutong Bai \enspace Sergey Levine \enspace Trevor Darrell\\
University of California, Berkeley
}

\begin{document}
\maketitle
\begin{strip}
\centering
\vspace{-1.5em}
\includegraphics[width=\linewidth]{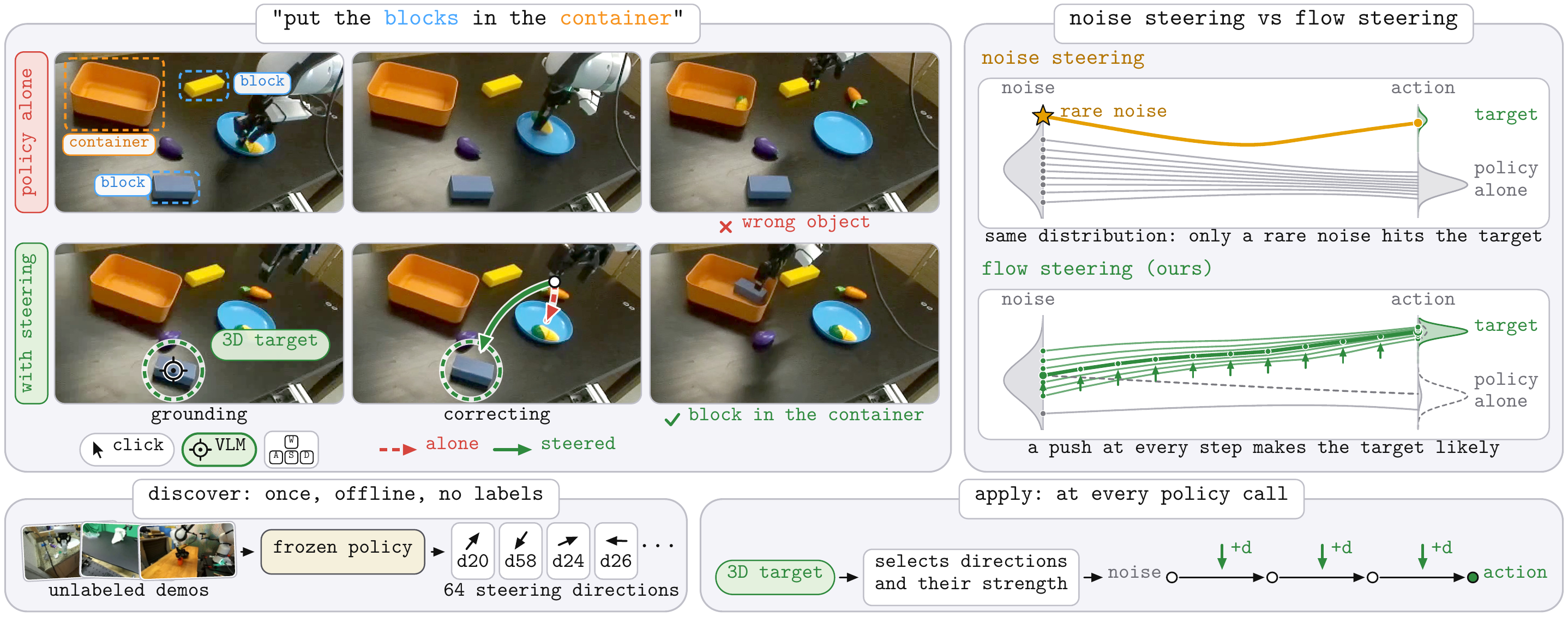}
\captionsetup{hypcap=false}
\captionof{figure}{\textbf{Steering a frozen policy toward the right motion.} Top left: the policy takes the wrong object; with a target from a click, VLM, or direction keys, the motion is steered toward it. Right: noise steering changes only the noise and needs a rare sample, while flow steering adds a push at every denoising step. Bottom: directions are learned once from unlabeled demonstrations and reused at every policy call.}
\captionsetup{hypcap=true}
\label{fig:teaser}%
\end{strip}

\begin{abstract}
Generalist robot policies learn manipulation skills across many scenes, but can repeatedly produce the wrong motion when a task requires reaching a different object or destination. Often the policy can  produce the needed motion but does not choose it. Steering adjusts a policy toward a desired outcome at test time, interactively or autonomously, but earlier methods act on the initial noise of the policy’s sampler and struggle to change the action distribution sufficiently. We show that a correct action that is unlikely in the policy's distribution is reached from the noise only by a rare sample, while a small push at every denoising step reaches it directly. We introduce Contrastive Direction Steering (CDS), an unsupervised method that learns reusable steering directions in the flow matching space from existing training data, without labels or weight updates. Each direction is an offset to the policy’s action representations that changes its predictions consistently across scenes. At inference, a target position from a person or a vision-language model determines which directions are applied. Across vision-language-action policies and world action models, CDS achieves 3.7 to 7.8 times the base-policy success rate on held-out LIBERO tasks, improves task success by 25\% relative on RoboLab-120, and raises the score on a real Franka arm from 32\% to 79\%. Project page: {\small\def\UrlBreaks{\do\/}\url{https://berkegokmen1.github.io/contrastive-direction-steering/}}
\end{abstract}

\section{Introduction}
\label{sec:intro}
\begin{figure*}[!t]
\centering
\includegraphics[width=\textwidth]{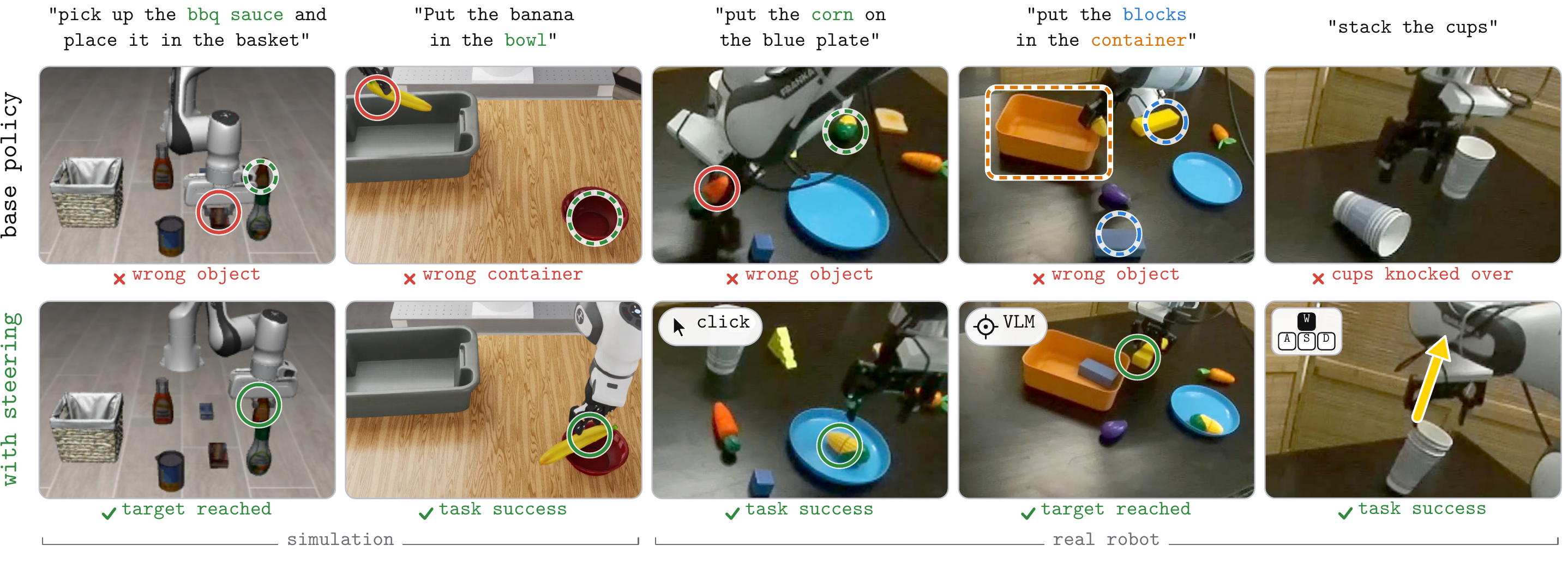}
\vspace{-2em}
\caption{\textbf{Steering corrects object selection and placement.} Base policy (top) and grad steering (bottom): two rollouts in simulation and three on the real robot, with the target from a click, a VLM or direction keys (badges). Solid rings indicate end effector positions; dashed marks show the target in the color of its corresponding word in the prompt; yellow arrow is the pressed key direction.}
\label{fig:behavior}
\end{figure*}

Training on diverse demonstrations gives robot policies experience with many objects and manipulation tasks~\cite{black2024pi0,intelligence2025pi05}. Some of that experience can be reused in tasks that were never demonstrated: a policy that has learned to pick up and place household objects may already have the manipulation skills needed to set a table in a new arrangement. However, giving the policy a new instruction does not necessarily make it use those skills as intended. In many cases, it incorrectly continues reaching for an object or placing it at a location associated with familiar training scenes~\cite{zhou2025liberopro,hou2026langgap,fang2026cf,xing2025shortcut}, even when it can execute the right grasping and placing motions. Such failures suggest an opportunity to reuse the policy's existing skills by guiding its choice of actions, rather than collecting more demonstrations for each new setting.

Often, what the policy lacks is not the ability to perform the correct motion, but ability to identify \textit{where} in the scene to direct it. Vision language models (VLMs) therefore offer a way to assist such policies with spatial guidance: a VLM can interpret the scene and instruction and point to a relevant object~\cite{deitke2024molmo}. For example, when setting a table, a VLM could identify the plate requested by the instruction and provide its location as a target, helping redirect a policy that would otherwise reach for a nearby cup. However, existing generalist policies lack an input channel for spatial targets. As a result, a mechanism for steering the policy in a given direction is needed to apply spatial guidance~\cite{nakamoto2024vgps,wang2025inference}.

Many generalist robot policies generate actions using diffusion or flow matching~\cite{chi2023diffusionpolicy,black2024pi0,kim2026cosmospolicy}. They start from random noise and gradually turn it into a sequence of robot commands using the scene and instruction. The noise is therefore a natural handle for steering, and prior work has steered these policies by choosing or adjusting it~\cite{tang2026frs,kao2026flowcontrol,wagenmaker2025dsrl}. In many of the failures we study, however, nearly all noise samples produce the same incorrect reach toward the cup, even when noise-based steering is applied. Optimizing the initial noise can produce a reach toward the target, but each step requires running the sampler again to predict where the gripper will end up, which is costly at every policy call (\cref{sec:alternative_controls}). We trace the limited effect of noise-based steering to how the noise reaches the action: its effect must persist through every denoising step, and along the way the policy pulls most noise samples toward the same familiar motion, so a strong redirection requires a large change to the noise. This motivates us to explore steering in the action space during denoising as a more powerful alternative to adjusting the initial noise.

We propose Contrastive Direction Steering (\ours), which steers the frozen policy by adding learned directions to its representation of the action sequence at every denoising step (\cref{fig:teaser}). We learn a small set of directions with a contrastive objective, using only existing demonstrations: each direction must produce an effect on the policy’s predictions that is consistent across scenes and distinct from the others. Each learned direction moves the gripper in a particular way, and adding it reweights the policy toward the actions that move that way. This does not give the policy new motions; it changes which of its existing ones it picks. Because the effects are consistent across scenes and learned over many actions, one fixed set of directions can be reused for new targets. Learning them requires no motion labels, task rewards, or changes to the policy weights. At inference, an external source such as a human or a VLM specifies where the gripper should go (\cref{fig:behavior}), and we set the direction strengths from each direction’s measured effect on the gripper, without searching over the noise.

We evaluate \ours\ on \pizero~\cite{black2024pi0}, \pifive~\cite{intelligence2025pi05}, FastWAM~\cite{yuan2026fastwam} and Cosmos Policy~\cite{kim2026cosmospolicy} on held-out LIBERO~\cite{liu2023libero} tasks and LIBERO-PRO~\cite{zhou2025liberopro} perturbations, and on \pifive-DROID on RoboLab-120~\cite{yang2026robolab}. Steering raises \pifive{} task success from $5.3\%$ to $20.6\%$ on LIBERO \hardsix{} and from $26.4\%$ to $33.1\%$ on RoboLab-120 (\cref{tab:libero,tab:robolab}). In a closed-loop LIBERO comparison, optimizing direction strengths takes less time per call than optimizing the initial noise (\cref{tab:noise_search}). Steering inside the policy's denoising process corrects a frozen generalist policy at deployment, without retraining.

\section{Related Work}
\label{sec:related}

\paragraph{Foundation models for robot control.}
Large robot datasets support generalist policies~\cite{brohan2023rt1,ghosh2024octo,oneill2024openx}.
RT-2~\cite{zitkovich2023rt2} and OpenVLA~\cite{kim2025openvla} adapt pretrained vision-language models to predict discretized robot actions as tokens.
Diffusion Policy~\cite{chi2023diffusionpolicy} instead models continuous action chunks through conditional denoising.
Vision-language-action (VLA) models such as \pizero and \pifive~\cite{black2024pi0,intelligence2025pi05} combine a vision-language backbone with a flow-matching action expert.
World action models (WAMs) build on video models: FastWAM~\cite{yuan2026fastwam} uses video co-training with a separate action expert, while Cosmos Policy~\cite{kim2026cosmospolicy} represents actions within the latents of a video diffusion model.
Rather than adapting these policies to each new task, we keep their weights fixed and change which of their learned behaviors they produce, with one discovery procedure applied to the action-expert representation or the action latent frame of each family.

\paragraph{Steering robot policies.}
Methods that steer a robot policy differ in the information supplied by a user or external model and in where they act.
LILAC~\cite{cui2023lilac} incorporates online language corrections into a learned low-dimensional action interface for shared autonomy, and diffusion-based shared autonomy~\cite{yoneda2023noise} moves user actions toward demonstrated behaviors by partial noising and denoising.
VoxPoser~\cite{huang2023voxposer} builds 3D value maps with language and vision models and plans trajectories on them, while AFI~\cite{xu2025afi} takes spatial guidance from affordance fields and recovers from stalls by waypoint search.
Other methods rank sampled actions~\cite{nakamoto2024vgps,wu2025foresight} or fine-tune the denoising process with reinforcement learning, as in DPPO~\cite{ren2025dppo}.
A frozen generator can be steered through its initial noise: with reward-trained noise policies in DSRL~\cite{wagenmaker2025dsrl}, approximate reversal of reference actions in FRS~\cite{tang2026frs}, simple real-time commands such as keyboard inputs in Flow Control~\cite{kao2026flowcontrol}, and, outside robotics, optimization through the frozen flow in D-Flow~\cite{ben-hamu2024dflow}.
Sampling can also be guided by alignment losses on point goals, trajectory sketches or physical corrections in Inference-Time Policy Steering (ITPS)~\cite{wang2025inference}, learned dynamics in DynaGuide~\cite{du2025dynaguide}, VLM-generated rewards in VLS~\cite{liu2026vls}, or Cartesian energy fields in OmniGuide~\cite{song2026omniguide}.
We take only a 3D target, leave the motion to the frozen policy, and change the denoiser's computation rather than its initial noise or sampling objective, using directions learned before deployment.

\paragraph{Steering internal representations.}
Internal representations provide another place to steer, in language models and more recently in VLAs.
Activation Addition~\cite{turner2023activation} and Contrastive Activation Addition~\cite{rimsky2024caa} use representation differences between contrasting examples to steer language models.
H\"aon et al.~\cite{haon2025mechanistic} identify concept-related neuron groups in VLAs and modulate them to control speed or direction.
Other methods learn steerable features with sparse autoencoders~\cite{swann2026sae}, estimate feature values with linear observers~\cite{buurmeijer2026observing}, or filter activations using conceptors fitted to rollout outcomes, as in COAST~\cite{miao2026coast}.
Unlike these methods, we learn directions without concept labels, rewards or success/failure partitions, and we measure the physical motion each one produces.

\paragraph{Generative controls and unsupervised discovery.}
Unsupervised discovery for generative control has been explored in the context of image generation.
Concept Sliders~\cite{gandikota2024sliders} learns controls for image attributes that are specified by text or image pairs during training.
Earlier work discovers controls in GANs without such labels~\cite{voynov2020unsupervised,harkonen2020ganspace,shen2021sefa}, and LatentCLR~\cite{yuksel2021latentclr} does so with a contrastive objective on the changes each direction causes.
NoiseCLR~\cite{dalva2024noiseclr} extends this to diffusion models: the same direction's effects on the denoising prediction should agree across images, and different directions' effects should differ.
We adapt this principle to temporally structured robot actions and evaluate the controls through measured end-effector motion and closed-loop task completion.

\section{Contrastive Direction Steering}
\label{sec:method}
\begin{figure*}[!t]
\centering
\includegraphics[width=\linewidth]{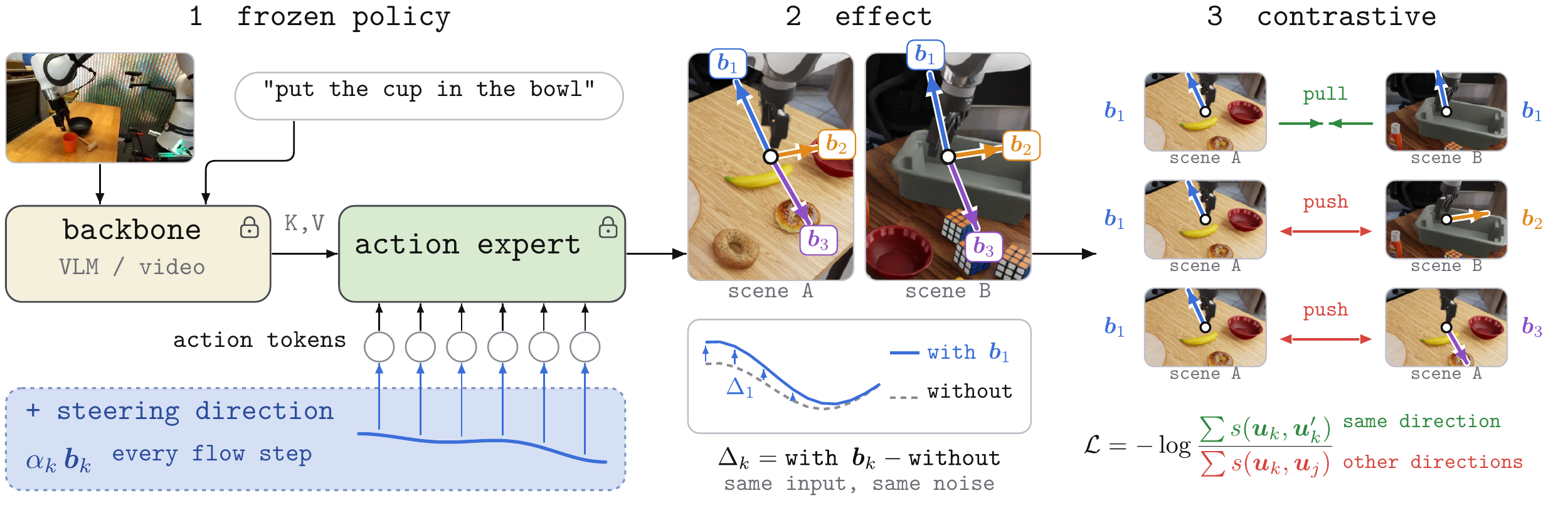}
\caption{\textbf{Unsupervised discovery of steering directions.} Offsets are added to the frozen action expert at each denoising step. Their effects are the differences between predictions with and without an offset, same input and noise. The objective encourages effects of the same direction to agree across scenes and noise samples, while separating effects of different directions. Only the directions are trained. The diagram shows the action-expert formulation for the VLA policies and FastWAM; Cosmos Policy uses its action latent frame.}
\label{fig:method}
\end{figure*}

\ours\ steers a frozen flow-matching policy by adding steering directions to its action representation during denoising, which reweights the policy's own action distribution (\cref{sec:setup}).
We do not know in advance which offsets produce which motions, so we learn a dictionary of directions from existing training data (\cref{sec:objective}, \cref{fig:method}) and measure the end-effector motion each one induces.
At inference, an externally supplied target sets their strengths, by lookup or by optimizing through the frozen sampler (\cref{sec:selectors}).

\subsection{Preliminaries}
\label{sec:prelim}
A flow-matching policy generates a chunk of $H$ actions, each with $d$ command components, conditioned on an observation $o$ and an instruction $\ell$~\cite{lipman2023flow,liu2023rectified,black2024pi0}.
Its action expert predicts a velocity $\vect{v}_\theta(\vect{x}_\tau,\tau\mid o,\ell)$ for a noisy action chunk $\vect{x}_\tau\in\mathbb{R}^{H\times d}$.
Sampling integrates this field from Gaussian noise at $\tau=1$ to an action chunk $\vect{a}=\vect{x}_0$ at $\tau=0$, which defines the policy's distribution $p(\vect{a}\mid o,\ell)$ over chunks for the call.
The expert receives one hidden token per action step, $\vect{h}\in\mathbb{R}^{H\times D}$.

\subsection{Steering a frozen policy}
\label{sec:setup}
A steering direction is a fixed offset $\vect{b}_k\in\mathbb{R}^{H\times D}$ in the space of the expert's action tokens, with one vector per action step.
We keep a dictionary of $N$ such directions for each policy, learned once from existing data (\cref{sec:objective}) and reused across scenes and tasks.
To steer the frozen policy, we add a weighted sum of directions to the action tokens at each denoising step $\tau$:
\begin{equation}
\vect{h}_\tau\leftarrow\vect{h}_\tau+\textstyle\sum_{k=1}^{N}\alpha_k\vect{b}_k.
\label{eq:steering direction}
\end{equation}
Here $\alpha_k$ sets the strength of direction $k$.
The same direction is used at every denoising step, with its strength held fixed for that policy call.
The policy's weights and its vision and language inputs remain unchanged.

For the VLA policies we study and FastWAM~\cite{yuan2026fastwam}, we steer the tokens of a flow-matching action expert. The VLA experts use vision-language features, while FastWAM's expert uses video features. Cosmos Policy~\cite{kim2026cosmospolicy} encodes the action chunk in a latent frame, so we add the directions directly to that frame during diffusion. Each policy uses its native sampler and reuses its learned directions across observations. Only their strengths change at inference.

\paragraph{Steering reweights the policy's distribution.}
\label{sec:theory}
Adding offsets to the action tokens changes which chunks the frozen policy samples.
Steering only moves probability among the chunks the policy can already produce, and because the offsets act inside the flow, the amount it moves has a closed form.
Let $p_{\vect{\alpha}}(\vect{a}\mid o,\ell)$ be the distribution of chunks sampled with the offsets of \cref{eq:steering direction}, and let $\Delta\vect{v}_\tau$ be the resulting change in the predicted velocity.
For a continuous flow with strengths fixed per call,
\begin{equation}
\begin{aligned}
p_{\vect{\alpha}}(\vect{a}\mid o,\ell)&=p(\vect{a}\mid o,\ell)\,f_{\vect{\alpha}}(\vect{a}),\\
\log f_{\vect{\alpha}}(\vect{a})&=\int_0^1\frac{\nabla\!\cdot\big(p_\tau\,\Delta\vect{v}_\tau\big)}{p_\tau}(\vect{x}_\tau)\,d\tau,
\end{aligned}
\label{eq:tilt}
\end{equation}
where $p_\tau$ is the unsteered density at flow time $\tau$ and $\vect{x}_\tau$ is the steered path ending at $\vect{a}$.
The weight $f_{\vect{\alpha}}(\vect{a})$ is how much probability the velocity change moves onto the path ending at $\vect{a}$.
Since $f_{\vect{\alpha}}$ is finite, a chunk the policy assigns zero probability keeps zero probability, so steering reuses motions the policy already has and does not invent new ones.
The derivation of \cref{eq:tilt} and its verification on a 2D toy policy and on \pifive-LIBERO are in \cref{sec:supp_theory}.

\paragraph{Why steer in the flow and not in the noise.}
Changing the initial noise also reweights the policy's distribution, but little of that change reaches the action.
Let $\Psi_\tau=\partial\vect{a}/\partial\vect{x}_\tau$ measure how much the final chunk moves when the sample at flow time $\tau$ is moved.
For small changes, a change $\vect{\delta}$ in the initial noise and a change $\Delta\vect{v}_\tau$ in the velocity move the chunk by
\begin{equation}
\Delta\vect{a}_{\mathrm{noise}}=\Psi_1\vect{\delta},
\Delta\vect{a}_{\mathrm{steer}}
=-\int_0^1\Psi_\tau\,\Delta\vect{v}_\tau(\vect{x}_\tau)\,d\tau.
\label{eq:gain}
\end{equation}
A confident policy maps most noise draws to a small set of chunks, so $\Psi_1$ is small: for \pifive-LIBERO, a unit change in the noise moves the chunk by about $10^{-3}$ per dimension on average (\cref{sec:supp_gain}).
At the last denoising steps the sample is already close to the action, so $\Psi_\tau\to I$ and a small push moves the action directly.
The same move made through the noise needs a large change, to a noise sample the policy rarely draws (\cref{fig:teaser}, right).
Steering in the flow is not more expressive than optimizing the noise: both reach the same actions, but the flow needs a much smaller intervention, which is faster to find (\cref{sec:alternative_controls}).

\subsection{Unsupervised direction discovery}
\label{sec:objective}
\label{sec:naming}
We learn the dictionary with a contrastive objective on the changes that directions cause in the policy's predictions.
This allows us to learn directions that (i) have a consistent effect across scenes, so that an effect measured once predicts what the direction will do in a new scene, and (ii) cover different motions, so that together they can move the policy in any direction.
Both are required at inference, where steering combines directions according to their measured effect in order to reach a specified target (\cref{sec:selectors}).
We represent each direction using the first $K$ temporal cosine components~\cite{ahmed1974dct}, reducing the number of learned coefficients and limiting variation across action steps.

Following NoiseCLR~\cite{dalva2024noiseclr}, we compare changes in the model's predictions rather than the direction coefficients themselves. For flow-matching experts, we measure a direction by the change it causes in the predicted velocity:
\begin{equation}
\Delta\vect{v}_k = \vect{v}_\theta(\vect{x}_\tau,\tau\mid o,\ell;\vect{b}_k) - \vect{v}_\theta(\vect{x}_\tau,\tau\mid o,\ell;\vect{0}).
\label{eq:direction_effect}
\end{equation}
The two predictions use the same observation, noisy chunk, and flow time, so the difference is due to the steering direction.
During discovery for these experts, we construct $\vect{x}_\tau$ by linearly mixing a recorded action chunk from the policy's training data with standard Gaussian noise, using $\tau$ as the weight on the noise. The policy retains the observation's own instruction. Each batch contains $R$ combinations of observations and noise samples, evaluated at a shared flow time.
Let $\vect{u}_{k,r}=\Delta\vect{v}_{k,r}/\|\Delta\vect{v}_{k,r}\|$ be the normalized effect of direction $k$ on row $r$, with the action chunk flattened into one vector.
Effects of the same direction on two different rows form positive pairs.
Effects of different directions form negative pairs, including pairs from different scenes.
We know which direction was applied to each prediction, so its identity determines these pair assignments without any annotation of its motion.
Every direction is used as an anchor:
\begin{equation}
\mathcal{L}_{\mathrm{clr}}=-\frac{1}{N}\sum_{k=1}^{N}
\log\frac{\sum_{r<s}s(\vect{u}_{k,r},\vect{u}_{k,s})}
{\sum_{j\ne k}\sum_{r,s}s(\vect{u}_{k,r},\vect{u}_{j,s})},
\label{eq:clr}
\end{equation}
where $s(\vect{u},\vect{v})=\exp(\vect{u}^{\top}\vect{v}/T)$ and $T$ is the temperature.
Because cosine similarity leaves the magnitude of each prediction change unconstrained, we rescale each direction after every update to bring its relative effect $\rho_k=\|\Delta\vect{v}_k\|/\|\vect{v}_\theta\|$ toward a common target.
No motion labels or task rewards enter this objective, so we do not know in advance which motion each direction produces.
After discovery we measure it: on held-out observations we sample one chunk without steering and one with direction $k$ at strength $1$, from the same noise, and take the mean shift of the predicted end-effector endpoint as the direction's measured effect $\vect{m}_k\in\mathbb{R}^{3}$.

The two terms of \cref{eq:clr} target these two properties: the positive term makes each direction's effect consistent across scenes (i), and the negative term spreads the effects of different directions apart (ii)~\cite{wang2020understanding}.
For a target unknown when the set is learned, these are the properties that make the distance left after lookup smallest (Proposition 2, \cref{sec:supp_why_clr}).

\subsection{Target-guided steering}
\label{sec:selectors}
\begin{figure}[!t]
\centering
\includegraphics[width=\linewidth]{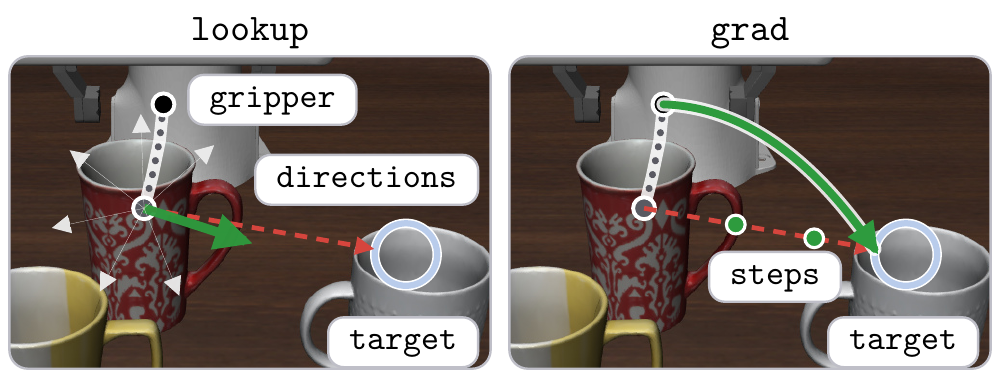}
\caption{\textbf{Choosing the direction strengths.} The policy alone (dotted) heads to the red mug, while the target is the white mug. Lookup picks the one direction whose stored effect best matches the needed move (red dashed). Grad takes gradient steps on the distance to the target through the frozen sampler (dots: chunk end after each step).}
\label{fig:demo_lookup_grad}
\end{figure}

An externally supplied target $\vect{p}^{\star}$ sets the direction strengths at each policy call (\cref{fig:demo_lookup_grad}).
The dictionary remains fixed, and the target is not used during direction discovery.
The target can come from three sources.
A person can click on the object or destination in the camera image.
A vision-language model such as Molmo~\cite{deitke2024molmo} or Qwen~\cite{qwen38} can point to the object or destination named in the instruction.
In both cases the camera calibration converts the pixel into a 3D position.
A person can also press direction keys, each of which places the target one step beyond the end of the policy's own chunk along the chosen axis.
In order to isolate the effectiveness of the steering method, our LIBERO and RoboLab benchmark comparisons instead use simulator object and placement positions, so that all target-based methods receive the same spatial information.
\Cref{sec:supp_targets} describes each source.

\begin{table*}[!t]
\centering
\scriptsize
\setlength{\tabcolsep}{1.5pt}
\begin{tabularx}{\textwidth}{@{}l*{24}{>{\centering\arraybackslash}X}@{}}
\toprule
 & \multicolumn{6}{c}{\pifive} & \multicolumn{6}{c}{\pizero} & \multicolumn{6}{c}{FastWAM} & \multicolumn{6}{c}{Cosmos Policy} \\
\cmidrule(lr){2-7}\cmidrule(lr){8-13}\cmidrule(lr){14-19}\cmidrule(lr){20-25}
 & & \multicolumn{5}{c}{LIBERO-PRO} & & \multicolumn{5}{c}{LIBERO-PRO} & & \multicolumn{5}{c}{LIBERO-PRO} & & \multicolumn{5}{c}{LIBERO-PRO} \\
\cmidrule(lr){3-7}\cmidrule(lr){9-13}\cmidrule(lr){15-19}\cmidrule(lr){21-25}
Method & H-62 & Swap & Obj. & Lang. & Task & Mean & H-62 & Swap & Obj. & Lang. & Task & Mean & H-62 & Swap & Obj. & Lang. & Task & Mean & H-62 & Swap & Obj. & Lang. & Task & Mean \\
\midrule
Base policy & 5.3 & 26.6 & \snd{86.5} & \snd{96.4} & 26.0 & 58.9 & 2.2 & 5.0 & \fst{79.0} & \fst{88.1} & 21.1 & \fst{48.3} & 1.3 & 0.6 & \fst{79.2} & \snd{81.2} & 14.6 & 43.9 & 2.7 & 3.2 & \fst{83.4} & 95.0 & 23.5 & \snd{51.3} \\
+ FRS~\cite{tang2026frs} & 7.2 & 31.6 & 80.6 & 91.5 & 31.2 & 58.8 & 3.5 & \snd{8.6} & 50.3 & 49.5 & 19.9 & 32.1 & 1.6 & 1.0 & 76.6 & \fst{83.8} & 15.1 & \snd{44.1} & 2.5 & 3.0 & \snd{82.6} & \snd{95.1} & 24.0 & 51.2 \\
+ FC~\cite{kao2026flowcontrol} & 7.3 & 30.4 & 85.2 & \fst{97.8} & 28.6 & \snd{60.5} & 3.9 & 6.1 & 73.9 & \snd{87.0} & \snd{23.9} & \snd{47.7} & 1.2 & 0.9 & \snd{77.8} & 80.7 & 14.8 & 43.5 & 3.9 & 3.8 & 81.0 & 93.1 & 26.4 & 51.1 \\
+ AFI~\cite{xu2025afi} & 8.1 & 27.1 & 78.2 & 90.8 & 23.8 & 55.0 & 3.7 & 5.6 & \snd{74.0} & 83.5 & 19.2 & 45.6 & 3.6 & 0.6 & 72.8 & 76.8 & 12.4 & 40.6 & 3.4 & 2.6 & 77.6 & 90.0 & 23.8 & 48.5 \\
+ COAST~\cite{miao2026coast} & 6.5 & 26.1 & \fst{87.9} & 96.2 & 25.9 & 59.0 & 3.5 & 5.1 & 69.9 & 82.1 & 20.5 & 44.4 & 1.7 & 0.6 & 77.0 & 80.0 & 13.2 & 42.7 & 2.3 & 3.5 & 82.3 & \fst{95.5} & 23.5 & 51.2 \\
+ \ours (lookup) & \snd{17.4} & \snd{37.4} & 71.9 & 79.9 & \snd{45.9} & 58.8 & \snd{5.8} & 8.5 & 43.2 & 38.6 & 21.4 & 27.9 & \snd{8.1} & \snd{15.5} & 61.7 & 64.6 & \snd{26.9} & 42.2 & \snd{6.1} & \snd{13.4} & 62.1 & 71.1 & \snd{31.0} & 44.4 \\
+ \ours (grad) & \fst{20.6} & \fst{45.6} & 73.2 & 84.1 & \fst{52.1} & \fst{63.8} & \fst{8.2} & \fst{20.5} & 64.2 & 71.6 & \fst{34.2} & \snd{47.7} & \fst{10.1} & \fst{20.9} & 66.9 & 69.8 & \fst{31.8} & \fst{47.3} & \fst{10.3} & \fst{24.2} & 69.5 & 79.5 & \fst{39.5} & \fst{53.2} \\
\bottomrule
\end{tabularx}
\caption{\textbf{Task success across four policies.} Success rates (\%) on LIBERO Hard-62 (H-62) and LIBERO-PRO. Mean averages the four perturbations. Dark and light green mark the best and second-best values per column.}
\label{tab:libero}
\end{table*}

In order to select the weights of the learned directions based on the target, we define two strategies: Lookup and Grad. Let $\vect{p}(\vect{\alpha})$ be the predicted end-effector position at the last action step to be executed, and $\vect{e}_0=\vect{p}(\vect{0})$ be the unsteered prediction under the same noise.

\paragraph{Lookup.}
Among the directions with sufficiently large, consistent measured effects, we select the direction
$k^{\star}=\argmax_k\cos(\vect{d},\vect{m}_k)$
whose measured effect best matches the required correction
$\vect{d}=\vect{p}^{\star}-\vect{e}_0$, and set
\begin{equation}
\alpha_{k^{\star}}=\operatorname{clip}\bigl(\vect{d}^{\top}\vect{m}_{k^{\star}}/\|\vect{m}_{k^{\star}}\|^{2},\,0,\,\alpha_{\max}\bigr).
\label{eq:lookup}
\end{equation}
All other strengths are zero.
Selecting the direction uses the stored effects and the unsteered prediction; the chosen direction is then used to sample the steered chunk.

\paragraph{Grad.}
We optimize the strengths through the frozen sampler to reduce
\begin{equation}
\mathcal{L}_{\mathrm{target}}(\vect{\alpha})=
\|\vect{p}(\vect{\alpha})-\vect{p}^{\star}\|^2+\lambda\|\vect{\alpha}\|^2.
\label{eq:target}
\end{equation}
Gradients pass through the sampling steps and the map from actions to end-effector position. The second term penalizes large direction strengths. The VLM or video conditioning is held fixed. The selected strengths weight the directions applied at each denoising step.

\begin{figure}[!htbp]
\centering
\includegraphics[width=\linewidth]{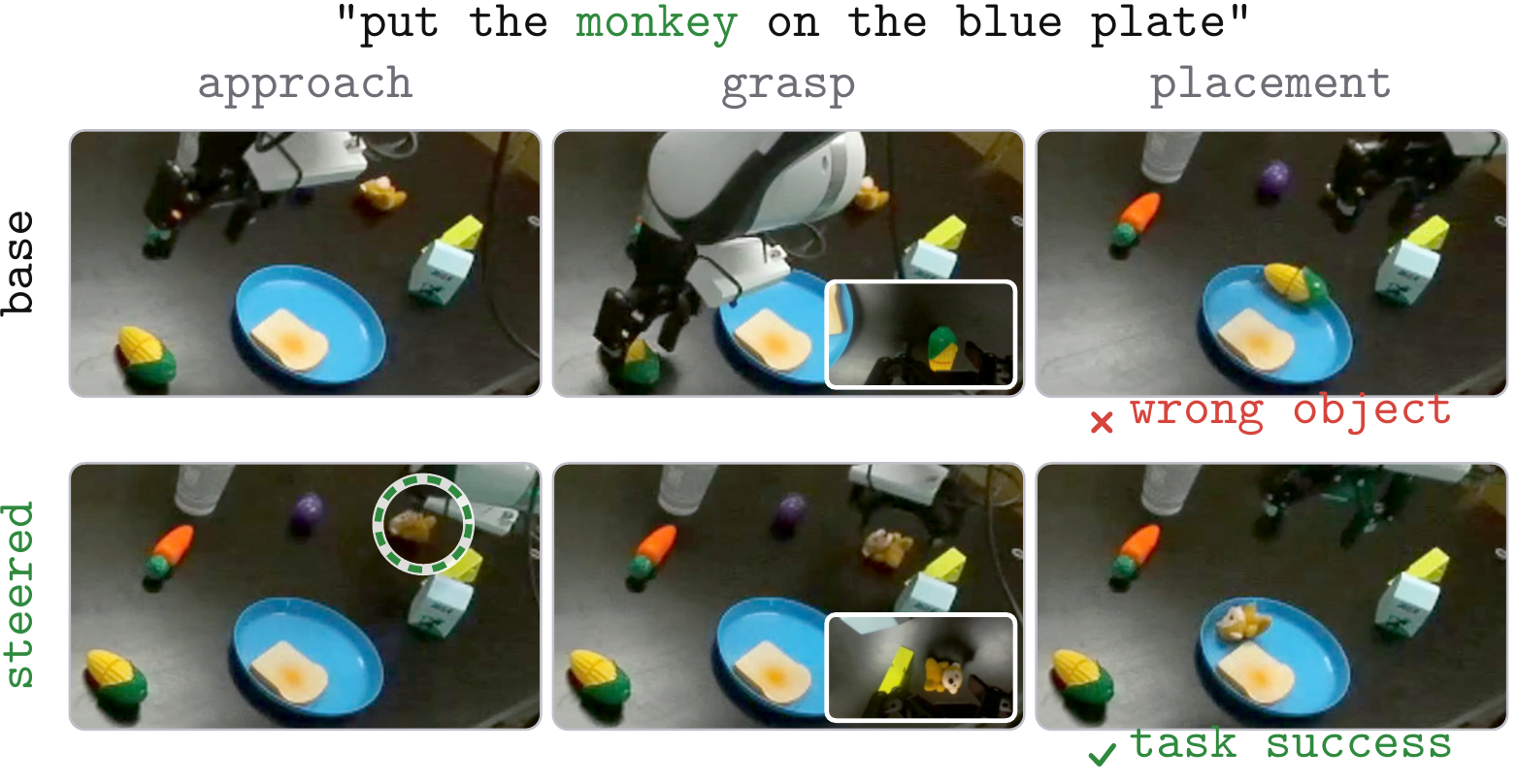}
\vspace{-2em}
\caption{\textbf{Real-robot rollout, base vs.\ steered} (same layout). The base policy takes the corn; steered toward the clicked monkey (dashed ring), it completes the task.}
\label{fig:demo_libero}
\end{figure}

\begin{figure*}[!t]
\centering
\includegraphics[width=\linewidth]{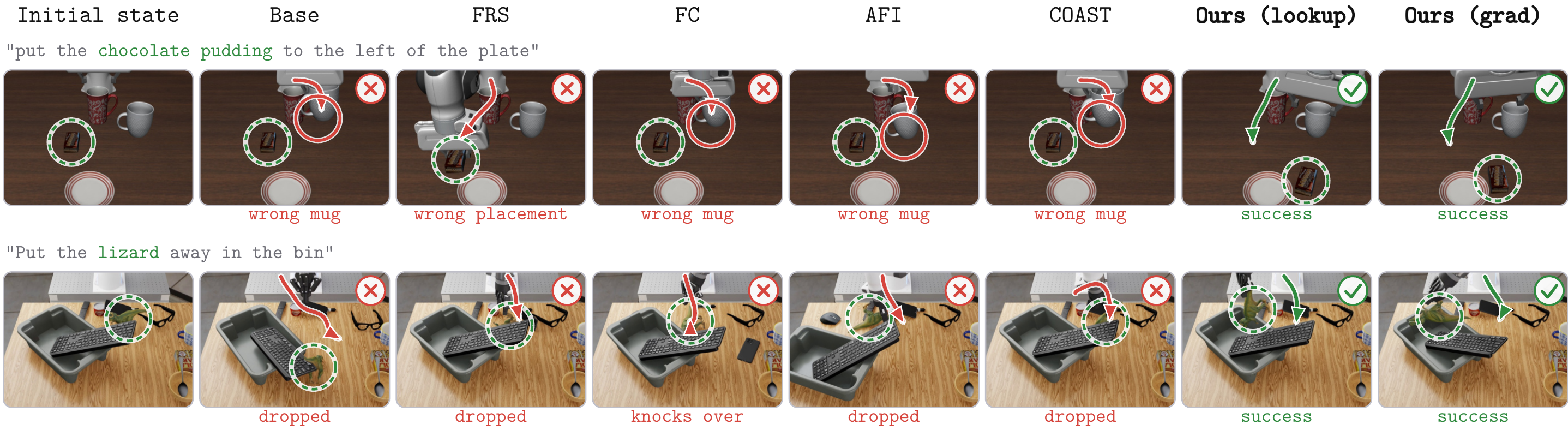}
\caption{\textbf{Methods compared from the same initial state.} Top: \pifive-LIBERO; bottom: \pifive-DROID on RoboLab. Arrows trace the gripper's path; dashed green circles mark the instructed object, red circles a wrong object.}
\label{fig:method_comparison}
\end{figure*}

\section{Experiments}
\label{sec:experiments}

\subsection{Experimental setup}
\label{sec:exp_setup}

\paragraph{Policies and tasks.}
We learn a steering direction dictionary for each released checkpoint: \pizero-LIBERO, \pifive-LIBERO, \pifive-DROID~\cite{black2024pi0,intelligence2025pi05}, FastWAM~\cite{yuan2026fastwam} and Cosmos Policy~\cite{kim2026cosmospolicy}.
Following FRS~\cite{tang2026frs}, we evaluate $50$ episodes on each of $62$ LIBERO-90 tasks~\cite{liu2023libero} where the reference \pifive succeeds at most $40\%$ of the time (listed in \cref{tab:supp_hard62}).
LIBERO-PRO~\cite{zhou2025liberopro} varies object positions (Swap), appearance (Object), instructions (Language) and goals (Task). We average success across these four perturbations and four suites, with $10$ tasks per suite and $20$ episodes per task.
RoboLab-120~\cite{yang2026robolab} comprises $64$ Simple, $39$ Moderate and $17$ Complex tabletop tasks for a Franka arm with DROID cameras. We report objects placed (OBJ) and tasks completed (SR) over six shared seeds per task.

\paragraph{Baselines.}
All methods share the observation pipeline and seeded noise stream.
In our comparison to existing steering methods, we study how well a method redirects a frozen policy once the target is known, not how the target is found. We therefore give FRS, Flow Control and \ours the same simulator location derived target and the same rule for when to steer, and evaluate the base policy as a reference, not as a competitor with equal information.
FRS~\cite{tang2026frs} reverses the sampler from a reference action toward the target, using approximate reversal for Cosmos. Flow Control (FC)~\cite{kao2026flowcontrol} writes the target direction into the starting noise. We reimplement AFI~\cite{xu2025afi}, which detects stalled or incorrect approaches and searches for waypoints using simulator geometry, and COAST~\cite{miao2026coast}, which filters hidden states using conceptors fitted to rollout outcomes.

\subsection{Task completion across policy families}
\label{sec:libero}

Lookup and grad improve Hard-62 success across all four backbones (\cref{tab:libero}).
For \pifive, grad reaches $20.6\%$, compared with $8.1\%$ for AFI and $5.3\%$ for the base policy.
\Cref{fig:method_comparison,fig:demo_libero} show how steering changes the approach to the instructed object.

\begin{table}[!tb]
\centering
\footnotesize
\setlength{\tabcolsep}{1pt}
\begin{tabularx}{\linewidth}{@{}l*{8}{>{\centering\arraybackslash}X}}
\toprule
 & \multicolumn{2}{c}{Simple} & \multicolumn{2}{c}{Moderate} & \multicolumn{2}{c}{Complex} & \multicolumn{2}{c}{All 120} \\
\cmidrule(lr){2-3}\cmidrule(lr){4-5}\cmidrule(lr){6-7}\cmidrule(lr){8-9}
Method & OBJ & SR & OBJ & SR & OBJ & SR & OBJ & SR \\
\midrule
Base policy & 37.9 & 28.4 & 48.5 & 27.4 & 43.5 & 16.7 & 42.1 & 26.4 \\
+ FRS~\cite{tang2026frs} & 37.9 & 28.1 & 48.3 & 27.8 & \fst{45.3} & \snd{22.5} & 42.3 & 27.2 \\
+ FC~\cite{kao2026flowcontrol} & 29.7 & 19.5 & 43.6 & 25.6 & 31.1 & 8.8 & 34.4 & 20.0 \\
+ AFI~\cite{xu2025afi} & 33.3 & 23.7 & \snd{49.6} & \snd{29.1} & 38.9 & 12.7 & 39.4 & 23.9 \\
+ COAST~\cite{miao2026coast} & 38.2 & 28.6 & 49.1 & 26.1 & 43.1 & 18.6 & 42.4 & 26.4 \\
+ \ours (lookup) & \snd{41.1} & \snd{31.2} & 49.4 & 28.6 & 41.6 & 18.6 & \snd{43.9} & \snd{28.6} \\
+ \ours (grad) & \fst{44.9} & \fst{35.7} & \fst{53.5} & \fst{32.5} & \snd{44.7} & \fst{24.5} & \fst{47.7} & \fst{33.1} \\
\bottomrule
\end{tabularx}
\caption{\textbf{RoboLab-120 with \pifive-DROID.} Objects placed (OBJ) and tasks completed (SR), in percent. Dark and light green mark the best and second-best values per column.}
\label{tab:robolab}
\end{table}

\paragraph{Perturbations.}
\label{sec:perturbations}
Object and Language perturbations preserve the task goal, so a familiar action can remain successful despite changes in appearance or instruction wording~\cite{zhou2025liberopro}.
Strong performance on these variants is compatible with reliance on familiar scene configurations.
The base policies already perform well ($86.5\%$ and $96.4\%$ for \pifive), and most steering methods reduce success (\cref{tab:libero}), a cost we discuss in \cref{sec:conclusion}.
Our gains concentrate on Swap and Task, where success requires redirecting the policy toward a different location or goal.
With grad, \pifive improves from $26.6\%$ to $45.6\%$ on Swap and from $26.0\%$ to $52.1\%$ on Task, and both world action models show similar gains.
Averaged over the four perturbations, \ours improves on the base policy for \pifive and both world action models.

\paragraph{Task complexity.}
\label{sec:robolab}

Grad reaches $33.1\%$ task success, compared with $26.4\%$ for the base policy and $27.2\%$ for FRS, and is best or second best in every complexity group (\cref{tab:robolab}).
AFI improves on the base policy on Moderate tasks but falls below it on Simple and Complex tasks. COAST matches the base policy's overall success rate.

\subsection{Alternatives to learned steering directions}
\label{sec:alternative_controls}

\begin{figure}[!b]
\centering
\includegraphics[width=\columnwidth]{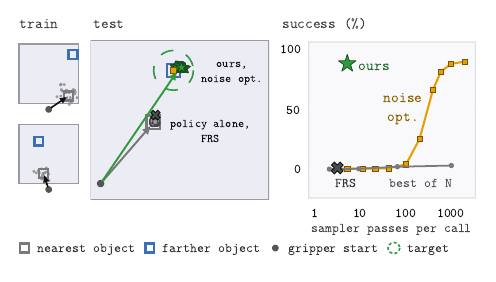}
\vspace{-2em}
\caption{\textbf{A familiar action in a new configuration.} Training demonstrations reach the nearest object. Test: a target toward the farther object (dashed circle, $5\times$ zoom). Right: success against the number of sampler passes per call.}
\label{fig:toy2d}
\end{figure}

\begin{table}
\centering
\small
\begin{tabularx}{\linewidth}{@{}l*{3}{>{\centering\arraybackslash}X}@{}}
\toprule
Method & Success (\%) & Opt. steps & Time (s) \\
\midrule
Base policy (\pifive) & 4.6 & 0 & 0.15 \\
+ \ours (lookup) & 15.1 & 0 & 0.34 \\
+ \ours (grad) & \fst{19.9} & 2 & 0.52 \\
+ Noise search & \snd{15.6} & 100 & 9.82 \\
\bottomrule
\end{tabularx}
\caption{\textbf{Optimizing the initial noise.} \pifive-LIBERO success on Hard-62 and time per call on one RTX 3090.}
\label{tab:noise_search}
\end{table}

\paragraph{Initial noise.}
To study the difficulty of steering in noise space, we construct a 2D example where a flow policy is instructed to reach one of two objects, but only is trained on examples that reach the nearest object.  At test time, the instruction names the farther object and an external target specifies the next $6$\,cm toward it, with a $2$\,cm success tolerance. Across $300$ calls (\cref{fig:toy2d}), best-of-$1{,}024$ sampling and FRS succeed in $3.0\%$ and $1.0\%$ of cases, while searching for the optimal noise reaches $89.7\%$ after $1{,}000$. In contrast, our learned steering directions reach $88.0\%$ after two steps. The policy rarely reaches the farther object when instructed, even though it reached objects at that location in other training scenes (\cref{sec:supp_toy2d}).

\begin{figure}[!tb]
\centering
\includegraphics[width=\linewidth]{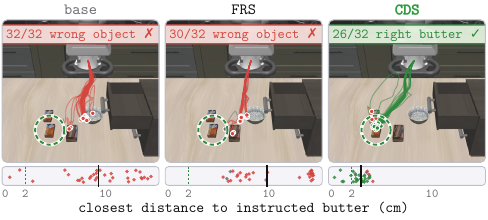}
\vspace{-2em}
\caption{\textbf{Resampling the noise is not enough.} $32$ noise seeds per method, each path drawn up to the grasp; the instruction asks for the butter at the front (dashed ring). Strips: closest distance of each path to that butter (black line: median; dashed: $2$\,cm).}
\label{fig:traj}
\end{figure}

The robot experiments show a similar difficulty: in the scene of \cref{fig:traj}, the base policy picks up another object in all $32$ seeds and FRS in $30$, while \ours\ picks up the instructed butter in $26$. For closed-loop \pifive-LIBERO (\cref{tab:noise_search}), we compare $100$ Adam steps on initial noise with two steps on direction strengths, using two initial states per Hard-62 task and three shared seeds for the steered methods (ten for base). Noise optimization improves on base but takes $18.9\times$ as long per steered call as grad. A noise solution is specific to its call: it must be searched again at every step, whereas a learned direction keeps its effect across scenes (\cref{sec:supp_transfer}).

\paragraph{Direct reach.}
We define a Reach baseline that moves the gripper straight to the target with a position controller and returns control to the policy within $8$\,cm. Reach improves on the base policy on Hard-62 ($16.7\%$ vs.\ $5.3\%$) but stays below grad ($20.6\%$). On LIBERO-PRO and RoboLab-120 it falls below the base policy ($36.6\%$ vs.\ $58.9\%$ and $17.9\%$ vs.\ $26.4\%$), since a straight-line motion does not use what the policy has learned about approaching objects and avoiding obstacles, which steering keeps (\cref{sec:supp_reach}).

\subsection{Ablations}
\label{sec:ablations}

We ablate four choices in direction discovery on \pifive-LIBERO: learned versus random directions, the number of directions, the number of temporal components per direction, and the motion regularizer.
Each variant is trained separately and evaluated under the protocol of \cref{sec:libero}, and \cref{tab:ablations} reports all results.

\begin{table}[!htbp]
\centering
\footnotesize
\renewcommand{\arraystretch}{0.92}
\setlength{\tabcolsep}{4pt}
\newcommand{\ablsep}{\arrayrulecolor{black!25}\specialrule{0.3pt}{1.5pt}{1.5pt}\arrayrulecolor{black}}
\begin{tabularx}{\linewidth}{@{}l*{4}{>{\centering\arraybackslash}X}}
\toprule
 & \multicolumn{2}{c}{\hardsix} & \multicolumn{2}{c}{LIBERO-PRO} \\
\cmidrule(lr){2-3}\cmidrule(lr){4-5}
 & grad & lookup & grad & lookup \\
\midrule
Base policy & \multicolumn{2}{c}{5.3} & \multicolumn{2}{c}{58.9} \\
\textbf{+ \ours} ($N{=}64$, $K{=}4$) & 20.6 & 17.4 & 63.8 & 58.8 \\
\quad random directions & 4.9 & n/a & 58.7 & n/a \\
\quad w/o regularizer & 19.2 & 13.8 & 63.3 & 53.8 \\
\ablsep
\multicolumn{5}{@{}l}{\textit{Number of steering directions ($N$)}} \\
\quad $N{=}16$ & 14.0 & 13.9 & 63.4 & \snd{55.0} \\
\quad $N{=}32$ & 17.0 & 13.8 & \snd{63.5} & 53.9 \\
\quad $\boldsymbol{N{=}64}$ & \snd{20.6} & \fst{17.4} & \fst{63.8} & \fst{58.8} \\
\quad $N{=}128$ & \fst{21.5} & \snd{16.4} & 61.3 & 53.3 \\
\ablsep
\multicolumn{5}{@{}l}{\textit{Temporal cosine components per direction ($K$)}} \\
\quad $K{=}2$ & \fst{21.3} & \snd{16.3} & 61.3 & \fst{59.2} \\
\quad $\boldsymbol{K{=}4}$ & \snd{20.6} & \fst{17.4} & \fst{63.8} & \snd{58.8} \\
\quad $K{=}8$ & 20.1 & 16.1 & \snd{62.5} & 55.2 \\
\bottomrule
\end{tabularx}
\caption{\textbf{Ablations on \pifive-LIBERO, task success (\%).} Rows below the default change one choice each.}
\label{tab:ablations}
\end{table}

\paragraph{Learned or random directions.}
To test whether contrastive discovery matters, we replace the dictionary with $64$ random directions in the same format, each scaled to the norm of a learned direction. Random directions barely move the gripper, so none is usable for lookup (\cref{tab:supp_coverage}), and with grad Hard-62 success stays at the base policy's level. Learning effective directions is thus important for the success of the inference-time search.

\paragraph{Dictionary size and temporal components.}
Grad combines all directions, and its Hard-62 success grows with the dictionary up to $128$; its LIBERO-PRO mean and both lookup scores peak at $N=64$.
$K=4$ gives the highest LIBERO-PRO mean with grad and the highest Hard-62 success with lookup.
We use $N=64$ and $K=4$ (\cref{sec:supp_ablations}).

\paragraph{Motion regularization.}
\label{sec:ablations_regularizer}
Without a constraint, discovery can find directions whose effect includes motions that do not occur in demonstrated chunks. We penalize the fraction of each prediction change outside a PCA subspace fitted to demonstrated actions. Removing this penalty lowers lookup's success on both benchmarks and raises its predicted path roughness from $5.8\times$ to $14.0\times$ that of the base policy. The regularizer is defined in \cref{eq:motion_regularizer}, and the full ablation, including a stronger weight, is in \cref{tab:supp_abl_w}.

\subsection{Real-robot experiments}
\label{sec:real}

\begin{figure}[h]
\centering
\includegraphics[width=\linewidth]{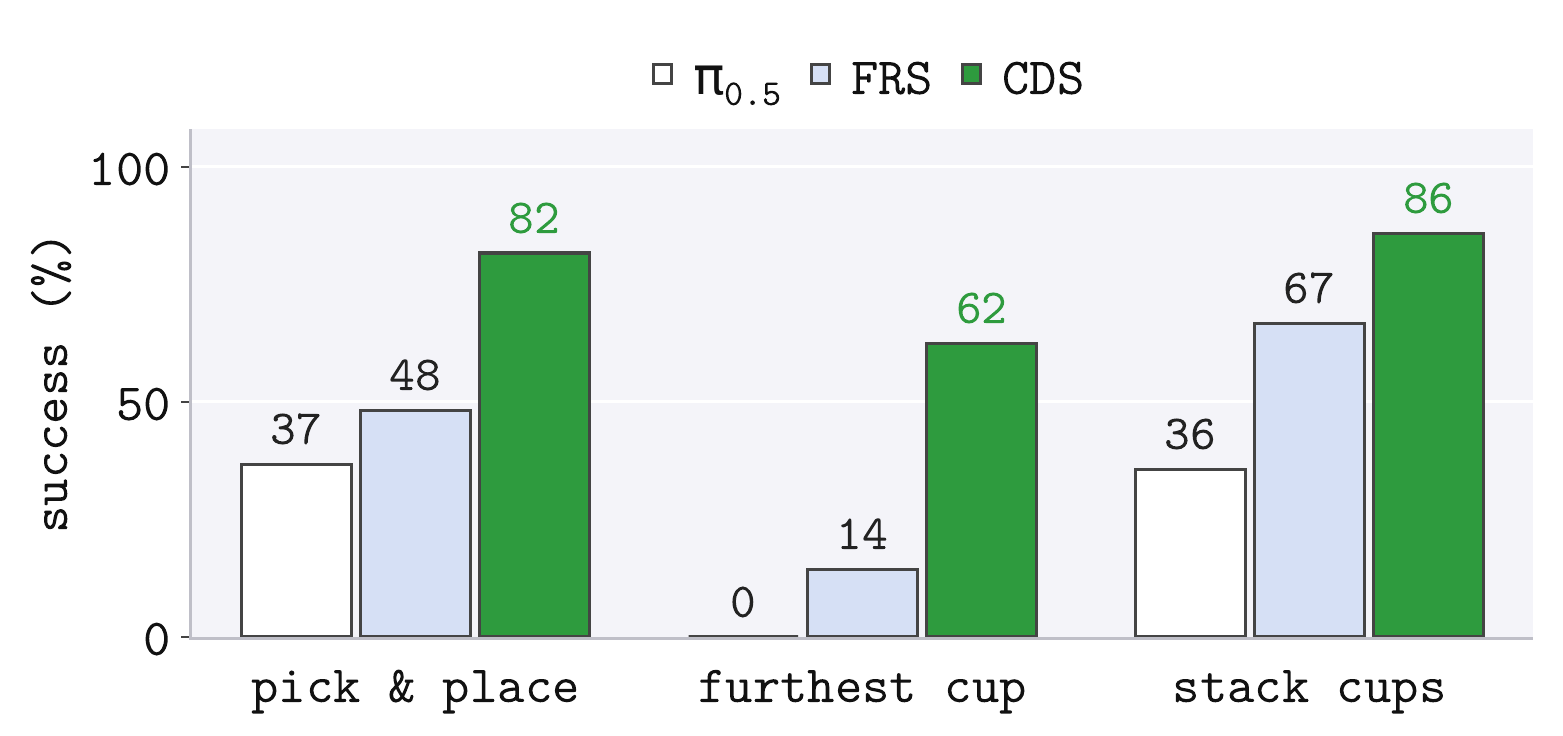}
\vspace{-2em}
\caption{\textbf{Real-robot success.} \pifive-DROID on a Franka arm, alone, with FRS and with \ours, on the same object layouts. A partial success counts as half.}
\label{fig:real_bars}
\end{figure}

We run \pifive-DROID on a Franka arm with one exterior and one wrist camera, with the policy weights unchanged. The tasks are eight pick-and-place instructions, picking up the furthest of several cups, and stacking cups. FRS and \ours\ receive the same targets, given by a click on the camera image, a VLM, or direction keys (\cref{sec:supp_targets}). We compare the three methods on the same object layouts. The policy alone often takes the nearest or wrong object (\cref{fig:demo_libero}) and succeeds in $32\%$ of the trials, FRS in $45\%$ and \ours\ in $79\%$ (\cref{fig:real_bars}). \Cref{sec:supp_real} gives the details per task.

\section{Discussion and Conclusion}
\label{sec:conclusion}

Generalist policies often fail on new arrangements not because they lack the motion, but because they choose the wrong one. \ours corrects this choice in a frozen policy: directions learned once from unlabeled demonstrations are added during denoising and steer the motion toward a given target. Our analysis and experiments in simulation and on a real robot show that steering inside the denoising process is more effective than on the initial noise, which changes the policy little or takes far longer per call. \ours has limitations: it changes which of its motions the policy picks (\cref{eq:tilt}) and cannot create a motion the policy has not learned. It steers only the end-effector position, so grasping and release still depend on the frozen policy. Success also depends on the externally supplied target: a wrong target steers the policy to the wrong place, and even a correct target can lower success when the base policy already acts correctly (\cref{sec:libero}). Within these limits, our results suggest that generalist policies hold more skills than their default behavior shows. Because steering needs no retraining and adds little time per call, these skills can be reached at deployment by a person or a VLM. Extending the directions to orientation and grasping is a natural next step.

\clearpage

{
    \small
    \bibliographystyle{ieeenat_fullname}
    \bibliography{main}
}

\clearpage
\setcounter{page}{1}
\maketitlesupplementary
\appendix

\noindent\textbf{\large Contents}\par\smallskip
\noindent\hyperref[sec:supp_impl]{\makebox[1.6em][l]{\ref*{sec:supp_impl}.}Models and training details}\dotfill\pageref{sec:supp_impl}\par
\noindent\hyperref[sec:supp_theory]{\makebox[1.6em][l]{\ref*{sec:supp_theory}.}Steering as a reweighting of the action distribution}\dotfill\pageref{sec:supp_theory}\par
\noindent\hyperref[sec:supp_why_clr]{\makebox[1.6em][l]{\ref*{sec:supp_why_clr}.}From the contrastive objective to the steering error}\dotfill\pageref{sec:supp_why_clr}\par
\noindent\hyperref[sec:supp_protocol]{\makebox[1.6em][l]{\ref*{sec:supp_protocol}.}Steering protocol and target construction}\dotfill\pageref{sec:supp_protocol}\par
\noindent\hyperref[sec:supp_real]{\makebox[1.6em][l]{\ref*{sec:supp_real}.}Real-robot evaluation}\dotfill\pageref{sec:supp_real}\par
\noindent\hyperref[sec:supp_vlm]{\makebox[1.6em][l]{\ref*{sec:supp_vlm}.}Vision-language model queries}\dotfill\pageref{sec:supp_vlm}\par
\noindent\hyperref[sec:supp_molmo]{\makebox[1.6em][l]{\ref*{sec:supp_molmo}.}Targets from a pointing model in simulation}\dotfill\pageref{sec:supp_molmo}\par
\noindent\hyperref[sec:supp_baselines]{\makebox[1.6em][l]{\ref*{sec:supp_baselines}.}Baseline implementations}\dotfill\pageref{sec:supp_baselines}\par
\noindent\hyperref[sec:supp_initial_noise]{\makebox[1.6em][l]{\ref*{sec:supp_initial_noise}.}Initial-noise and transfer studies}\dotfill\pageref{sec:supp_initial_noise}\par
\noindent\hyperref[sec:supp_readout]{\makebox[1.6em][l]{\ref*{sec:supp_readout}.}Direction readout and motion metrics}\dotfill\pageref{sec:supp_readout}\par
\noindent\hyperref[sec:supp_benchmarks]{\makebox[1.6em][l]{\ref*{sec:supp_benchmarks}.}Full benchmark results}\dotfill\pageref{sec:supp_benchmarks}\par
\noindent\hyperref[sec:supp_ablations]{\makebox[1.6em][l]{\ref*{sec:supp_ablations}.}Full ablation results}\dotfill\pageref{sec:supp_ablations}\par
\begin{figure*}[p]
\section{Additional qualitative comparisons}
\label{sec:supp_qualitative}
\noindent \Cref{fig:supp_comparison_pi0,fig:supp_comparison_robolab,fig:supp_comparison_fastwam,fig:supp_comparison_cosmos} compare the methods from the same initial state for each policy, with marks as in \cref{fig:method_comparison}. Each panel shows the last frame of the rollout. \Cref{fig:supp_comparison_real} does the same on the real robot for the policy alone, FRS and \ours. \Cref{fig:supp_real_ours} follows four real-robot runs of \ours\ from start to end.
\medskip

\begin{minipage}{\textwidth}
\centering
\includegraphics[width=0.9\textwidth]{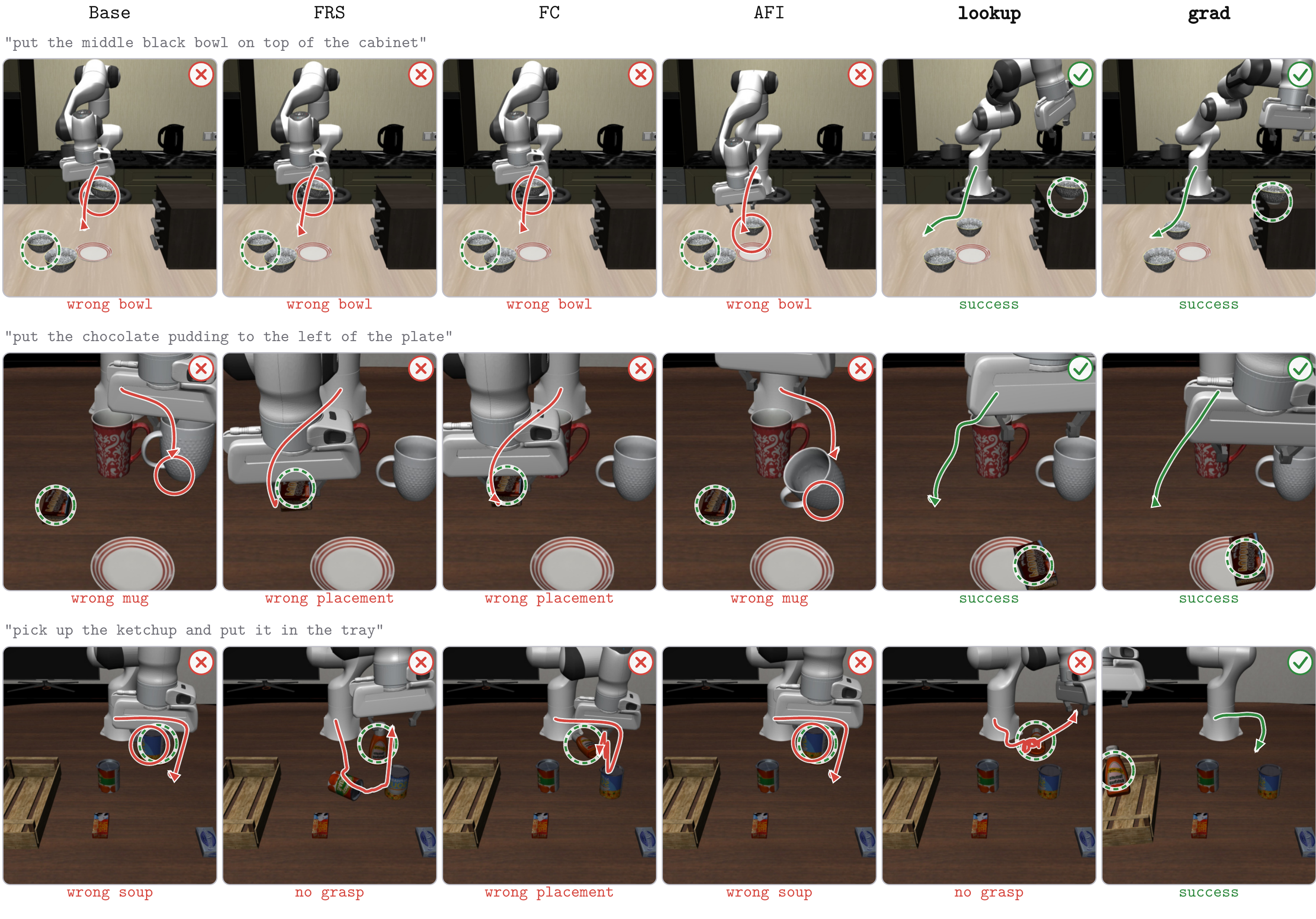}
\captionof{figure}{\textbf{Additional LIBERO comparisons.} Three Hard-62 tasks, all methods from the same initial state: \pifive-LIBERO (top two rows) and \pizero-LIBERO (bottom row).}
\label{fig:supp_comparison_pi0}
\end{minipage}
\medskip

\begin{minipage}{\textwidth}
\centering
\includegraphics[width=\textwidth]{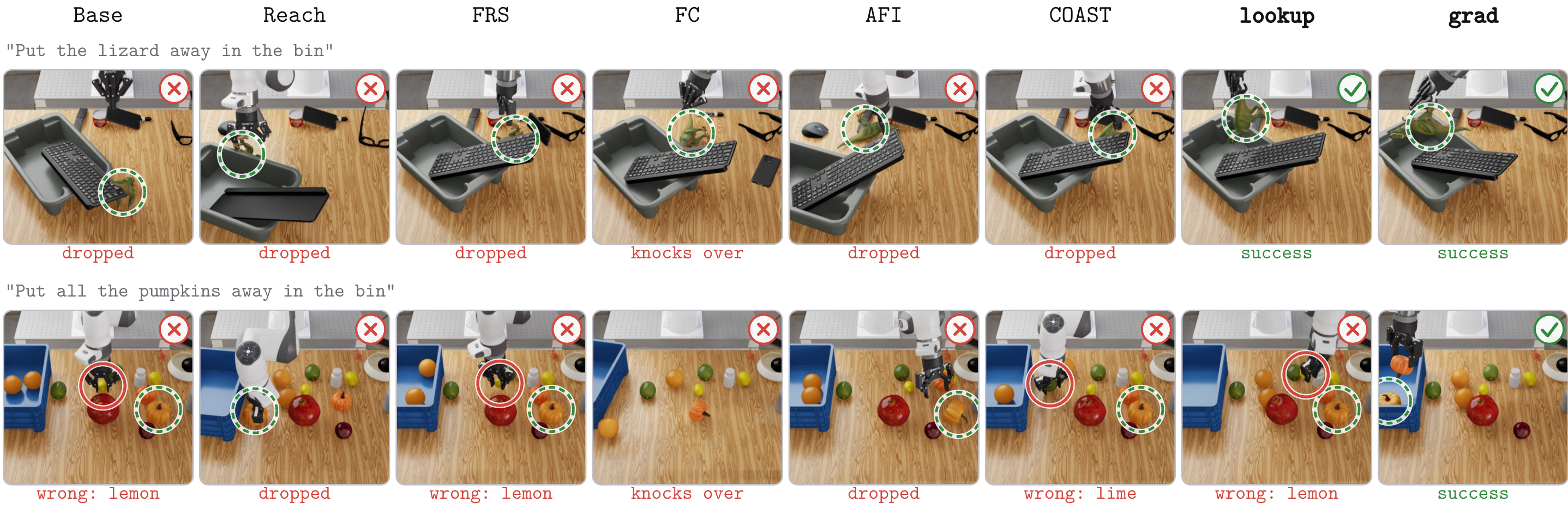}
\captionof{figure}{\textbf{Additional RoboLab comparisons.} \pifive-DROID on two RoboLab tasks, including the Reach baseline. The first row is the scene of \cref{fig:method_comparison}.}
\label{fig:supp_comparison_robolab}
\end{minipage}
\end{figure*}

\noindent\hyperref[sec:supp_qualitative]{\makebox[1.6em][l]{\ref*{sec:supp_qualitative}.}Additional qualitative comparisons}\dotfill\pageref{sec:supp_qualitative}\par
\medskip

\section{Models and training details}
\label{sec:supp_impl}
\label{sec:supp_models}

\begin{table*}[t]
\centering
\small
\setlength{\tabcolsep}{5pt}
\begin{tabular}{@{}lcll@{}}
\toprule
Policy & Action chunk & Where directions are added & Sampling \\
\midrule
\pifive-DROID & $15\times8$ & Projected action tokens, width $1024$ & $10$ Euler steps \\
\pifive-LIBERO & $10\times7$ & Projected action tokens, width $1024$ & $10$ Euler steps \\
\pizero-LIBERO & $50\times7$ & Tokens after the action-time MLP & $10$ Euler steps \\
FastWAM & $32\times7$ & Projected action tokens, width $1024$ & $10$ Euler steps, shifted grid \\
Cosmos Policy & $16\times7$ & Preconditioned action latent frame & EDM, $5$ denoiser calls \\
\bottomrule
\end{tabular}
\caption{\textbf{Policy-specific steering.} The action chunk shape is horizon by command dimension, excluding padding. Seven-dimensional LIBERO commands contain translation, rotation and gripper values; the DROID commands contain joint positions (joint velocities on the real robot) and a gripper value. Each model retains its own action normalization and sampler.}
\label{tab:supp_models}
\end{table*}

\paragraph{VLA checkpoints.}
The LIBERO evaluations use the released openpi \pifive-LIBERO and \pizero-LIBERO checkpoints.
RoboLab uses \pifive-DROID with joint-position commands.
The real-robot experiments use the \pifive-DROID checkpoint with joint-velocity commands, for which we learn a separate dictionary.
For the DROID checkpoint, the $15$-step chunk spans one second at $15$\,Hz, and the action expert's input width is $1024$.
Each step contains eight commands: seven joint positions and a gripper command.
For \pizero, directions are added after the action and time embeddings are combined; for \pifive, they are added after the action input projection (\cref{tab:supp_models}).
The backbone cache is shared across all steering directions evaluated on the same observation.
The vision-language backbone is unchanged, so its key-value cache is computed once and reused for every direction and noise sample on that observation.

\paragraph{FastWAM.}
We use the released \texttt{libero\_\allowbreak uncond\_\allowbreak 2cam224.pt} checkpoint~\cite{yuan2026fastwam}, which predicts actions from the current frame rather than a generated future video.
The exterior and wrist images are resized to $224\times224$ and concatenated horizontally.
The video backbone processes these observations once per policy call; the resulting key-value cache conditions the action expert throughout sampling.
Directions are added after the action input projection, giving a $32\times1024$ offset.
The sampler uses $10$ Euler updates on FastWAM's shifted time grid, with shift $5$.
The evaluation queue uses steering directions saved after $2{,}000$ training steps.

\paragraph{Cosmos Policy.}
We use \nolinkurl{nvidia/Cosmos-Policy-LIBERO-Predict2-2B}~\cite{kim2026cosmospolicy}.
The $16\times7$ action chunk is tiled into a latent frame of shape $16\times28\times28$ and read out by averaging the copies.
A direction has separate temporal DCT coefficients for each of the $112$ full copies and seven action dimensions.
It is added to the action frame after EDM input preconditioning, before patch embedding; the same direction is used at every denoiser call.
The effect used for contrastive training is the change in the decoded clean-action prediction.
We retain the deterministic EDM sampler with noise levels from $80$ to $4$, four integration updates and a final clean prediction, giving five denoiser calls.
The model jointly denoises future observations, so its conditioning cannot be reused as an independent backbone cache in the same way as the other policies.
Training therefore evaluates a subset of the $64$ directions per update.

\paragraph{Training observations and optimization.}
DROID observations are drawn from episodes with a successful demonstration and an instruction.
We exclude idle chunk starts; $34{,}352$ of $34{,}378$ episodes contain retained frames.
The episode's own instruction is used as the prompt.
LIBERO directions use LIBERO training observations.
VLA directions are trained for $10$k steps on DROID~\cite{khazatsky2024droid} or LIBERO~\cite{liu2023libero} observations, matching the policy's training domain; the WAM settings are given above.
These runs include the motion regularizer defined with its ablation in \cref{sec:ablations_regularizer}.
The reference \pifive-LIBERO configuration uses Adam with learning rate $0.01$ and initial coefficient scale $0.65$; VLA training distributes the scene/noise rows across eight GPUs.
For \pizero, initial coefficients and learning rate are scaled according to the measured sensitivity of the expert input.
FastWAM uses initial scale $1.25$ and learning rate $0.02$ to obtain a similar starting effect size.

\paragraph{Contrastive batches.}
We train $N=64$ directions per policy.
For the VLA runs, each update samples $S=8$ observations and $M=8$ noise draws per observation at a shared flow time, giving $R=SM$ rows in \cref{eq:clr}.
The temperature is $T=0.1$.

\paragraph{Motion-subspace fit.}
For direction $k$ and sample $r$, the regularized training loss is
\begin{equation}
\begin{split}
\mathcal{L}&=\mathcal{L}_{\mathrm{clr}}+w\mathcal{L}_{\mathrm{real}},\\
\mathcal{L}_{\mathrm{real}}&=1-\frac{1}{NR}\sum_{k,r}
\frac{\|\vect{U}\Delta\vect{v}_{k,r}\|^2}{\|\Delta\vect{v}_{k,r}\|^2}.
\end{split}
\label{eq:motion_regularizer}
\end{equation}
The regularizer in \cref{eq:motion_regularizer} uses a PCA fitted to $20{,}000$ real action chunks in the policy's normalized action coordinates.
The \pifive-DROID fit retains $17$ components; the basis is refitted for LIBERO and for each WAM's action normalization.
The retained basis has orthonormal rows.
The default regularization weight is $5$.

\paragraph{Effect-size calibration.}
The effect-size rule rescales directions toward a relative effect of $0.10$, calibrated against the effect of an instruction relative to an uninformative prompt.
An exponential moving average is used to estimate each direction's relative effect, and each per-update rescaling factor is clipped to $[0.9,1.1]$.

\paragraph{Temporal basis.}
\label{sec:dct}
Each direction acts over a sequence of action tokens.
To constrain variation between adjacent offsets, we use the first $K$ components of a temporal DCT-II basis~\cite{ahmed1974dct}:
\begin{equation}
\vect{b}_k=\vect{\Phi}^{\top}\vect{C}_k,
\qquad \vect{\Phi}\in\mathbb{R}^{K\times H},\quad
\vect{C}_k\in\mathbb{R}^{K\times D}.
\label{eq:dct_directions}
\end{equation}
The rows of $\vect{\Phi}$ are normalized cosine waves along the action chunk, including the constant component, and we learn the coefficient matrices $\vect{C}_k$.
The constant component controls the direction's mean, while the other components vary its temporal shape.
The basis varies across actions within a chunk, and the resulting direction is reapplied at each denoising step.
This constrains the offsets' temporal variation. The generated end-effector path also depends on the nonlinear policy.
We use $K=4$ components unless stated otherwise.
For \pifive-DROID, this reduces the number of learned values per direction from $15\times1024$ to $4\times1024$.
For the $15$-step, $15$\,Hz DROID chunk, the first four DCT waves have frequencies $0$, $0.5$, $1$ and $1.5$\,Hz.
They can change the mean and broad shape of a direction without assigning an independent coefficient to every step.
On $1{,}500$ random DROID chunks, waves 0--3 retain $99.8\%$ of the energy of joint-position actions and $98.0\%$ of joint velocities, with each joint standardized.
Rebuilding the joint-position chunks from these waves moves the tool point at the chunk end by $2.7$\,mm at the median and $8.8$\,mm at the 95th percentile, against a median chunk travel of $92$\,mm.
For the learned \pifive-DROID directions at strength $1$, $99.9\%$ of the energy of the tool-path change lies in the same waves.

\section{Steering as a reweighting of the action distribution}
\label{sec:supp_theory}
\paragraph{Setting.}
Fix the observation and instruction.
Write $\vect{v}_\tau$ for the unsteered velocity field and $\vect{v}_\tau+\Delta\vect{v}_\tau$ for the steered one, where $\Delta\vect{v}_\tau$ is the change caused by the offsets of \cref{eq:steering direction}.
Let $p_\tau$ and $q_\tau$ be the unsteered and steered densities at flow time $\tau$, so that $p_0=p(\cdot\mid o,\ell)$ and $q_0=p_{\vect{\alpha}}(\cdot\mid o,\ell)$.

\paragraph{Assumptions.}
(i) Both samplers start from the same Gaussian noise, $p_1=q_1$.
(ii) Both velocity fields are smooth and generate invertible maps with differentiable inverses.
(iii) The strengths $\vect{\alpha}$ are fixed for the call and do not depend on the sampled noise.
(iv) The integral in \cref{eq:tilt} is finite.

\paragraph{Proposition 1 (steering reweights the policy's distribution).}
Under (i)--(iv), for every chunk $\vect{a}$ and the steered path $\vect{x}_\tau$ that ends at it,
\begin{equation*}
\begin{aligned}
p_{\vect{\alpha}}(\vect{a}\mid o,\ell)&=p(\vect{a}\mid o,\ell)\,f_{\vect{\alpha}}(\vect{a}),\\
\log f_{\vect{\alpha}}(\vect{a})&=\int_0^1\frac{\nabla\!\cdot\big(p_\tau\,\Delta\vect{v}_\tau\big)}{p_\tau}(\vect{x}_\tau)\,d\tau,
\end{aligned}
\end{equation*}
and $\mathbb{E}_{\vect{a}\sim p}[f_{\vect{\alpha}}(\vect{a})]=1$.
In particular, a set of chunks with zero probability under $p$ has zero probability under $p_{\vect{\alpha}}$.

\paragraph{Proof.}
Let $s=1-\tau$, so that sampling runs forward in $s$ with velocity $\vect{w}_s=-\vect{v}_{1-s}$ for the unsteered flow and $\vect{w}_s+\vect{d}_s$ for the steered one, with $\vect{d}_s=-\Delta\vect{v}_{1-s}$.
A density $\rho_s$ carried by a field $\vect{w}_s$ obeys the continuity equation $\partial_s\rho_s=-\nabla\!\cdot(\rho_s\vect{w}_s)$, that is, $\partial_s\log\rho_s=-\nabla\!\cdot\vect{w}_s-\vect{w}_s^{\top}\nabla\log\rho_s$.
Along the steered path $\dot{\vect{x}}_s=(\vect{w}_s+\vect{d}_s)(\vect{x}_s)$, the chain rule gives
\begin{equation*}
\begin{aligned}
\frac{d}{ds}\log q_s(\vect{x}_s)&=-\nabla\!\cdot(\vect{w}_s+\vect{d}_s),\\
\frac{d}{ds}\log p_s(\vect{x}_s)&=-\nabla\!\cdot\vect{w}_s+\vect{d}_s^{\top}\nabla\log p_s.
\end{aligned}
\end{equation*}
The second has an extra term because the unsteered density is evaluated along a path that is not its own.
Subtracting,
\begin{equation*}
\frac{d}{ds}\log\frac{q_s(\vect{x}_s)}{p_s(\vect{x}_s)}
=-\nabla\!\cdot\vect{d}_s-\vect{d}_s^{\top}\nabla\log p_s
=-\frac{\nabla\!\cdot(p_s\vect{d}_s)}{p_s}(\vect{x}_s).
\end{equation*}
By (i) the log-ratio is zero at $s=0$.
Integrating over $s\in[0,1]$ and substituting $\tau=1-s$ gives \cref{eq:tilt}.
The weight averages to one because $p_{\vect{\alpha}}$ integrates to one, and the last claim follows since $f_{\vect{\alpha}}$ is finite by (iv). \hfill$\square$

\paragraph{Relation to initial-noise steering.}
Noise-initialization methods~\cite{wagenmaker2025dsrl,tang2026frs} change the distribution of the input noise and keep the generator fixed, which reweights $p$ by the ratio of the new noise density to the old one.
For a continuous flow the noise-to-action map is invertible, so every chunk the policy can produce has a noise that produces it, even if Gaussian draws almost never do.
This does not give an efficient way to find that noise.
Our offsets instead change the generator's computation during sampling, and \cref{sec:supp_gain} compares how far the two kinds of change move the chunk.

\paragraph{Scope.}
The proposition is for a continuous flow.
The deployed samplers take a finite number of Euler steps, and grad chooses strengths that depend on the sampled noise, so neither is covered by the proof.
The checks below measure how well \cref{eq:tilt} describes the samplers we use.

\paragraph{Numerical checks.}
We check it numerically on the two-dimensional toy policy, where every density can be drawn, and on \pifive-LIBERO, where an action chunk has $320$ dimensions.
In both cases the steered and unsteered densities are computed exactly by the instantaneous change of variables, independently of \cref{eq:tilt}, and compared with the prediction $p\,f_{\vect{\alpha}}$.

\subsection{Two-dimensional toy policy}
\label{sec:supp_theory_toy}
We use the toy policy of \cref{sec:supp_toy2d}.
Its demonstrations always reach for the object nearer the robot, so the policy reaches for the wrong object when the instruction names the other one.
We steer it toward the instructed object with the learned directions at the strengths selected by grad, and integrate the flow finely (RK4, $64$ steps) so that it is close to continuous.

\begin{figure*}[t]
\centering
\includegraphics[width=\linewidth]{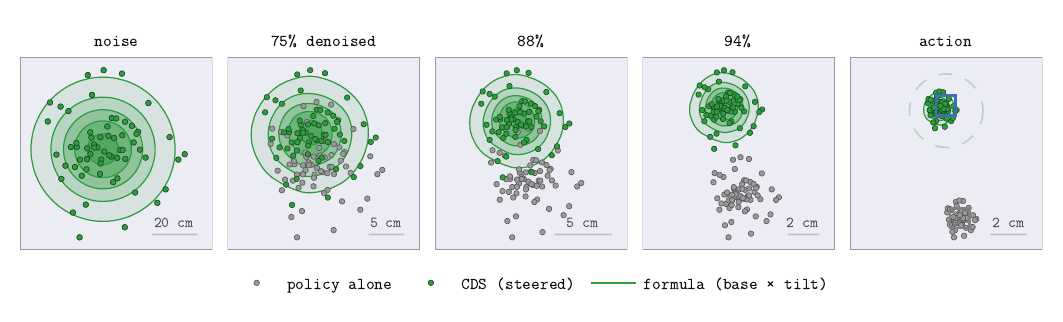}
\caption{\textbf{Steering as a reweighting on the 2D toy policy.} One sampling run, from noise (left) to the final action (right); each panel zooms on its own samples (scale bars). Grey: unsteered samples. Green dots: samples from the steered policy. Green contours: the prediction of \cref{eq:tilt} at that step, the unsteered density reweighted by the tilt accumulated so far. Final panel: grey square, the object the unsteered policy reaches for; blue square, the instructed object; dashed circle, $2$\,cm around the steering target.}
\label{fig:supp_tilt_toy}
\end{figure*}

\Cref{fig:supp_tilt_toy} follows one sampling run from noise to the final action.
At every step, the steered samples lie inside the density predicted by \cref{eq:tilt}, evaluated with the tilt accumulated up to that step.
Most of the movement happens in the last quarter of denoising.
Over four test scenes, the predicted and the actual steered action distributions differ in $1.7$--$7.1\%$ of their probability mass (total variation), while the steered and unsteered distributions do not overlap (total variation $1.00$).
Across eight scenes and three steering magnitudes ($0.1\times$, $0.3\times$ and $1\times$ the selected strengths), the formula accounts for more than $99\%$ of the change in log-probability.

\subsection{\pifive-LIBERO}
\label{sec:supp_theory_pi05}
In $320$ dimensions the densities cannot be drawn, so we compare them chunk by chunk.
We take four held-out LIBERO calls from different tasks and fix the lookup strengths selected for each call.
From each call we sample $6$ chunks from the unsteered and $6$ from the steered policy with a continuous-time sampler.
For each chunk we compute the unsteered and steered log-densities exactly, taking the divergence as the full trace of the $320\times320$ Jacobian, and compare their difference with $\log f_{\vect{\alpha}}$ from \cref{eq:tilt}, where $\nabla\log p_\tau$ is obtained by differentiating through the unsteered flow.
Integration uses RK4 with $16$ steps on a grid that is approximately uniform in $\log\tau$, because the integrand is concentrated near the end of denoising.

\begin{figure*}[t]
\centering
\includegraphics[width=0.85\linewidth]{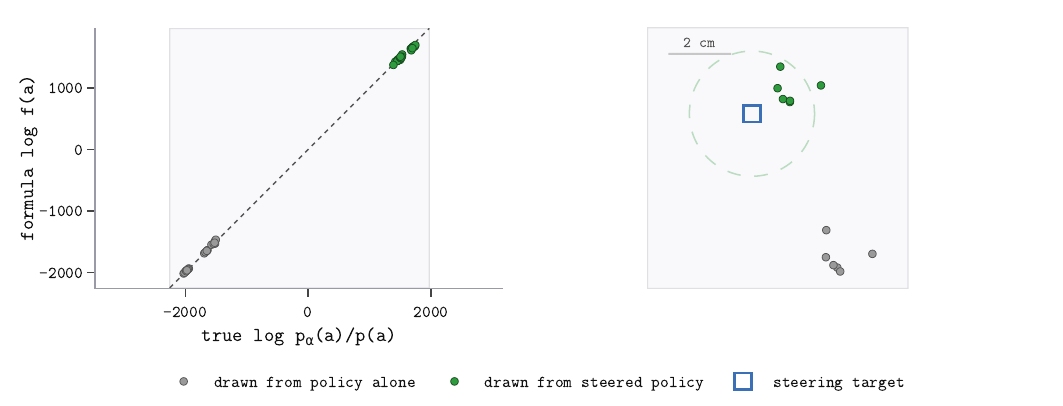}
\caption{\textbf{Steering as a reweighting on \pifive-LIBERO.} Left: for each of $48$ sampled $320$-dimensional action chunks, the change in log-probability computed from the exact steered and unsteered densities, against the prediction of \cref{eq:tilt}; dotted, equality. Grey: chunks from the unsteered policy. Green: chunks from the steered policy. Right: chunk ends seen from above for one call; blue square, the steering target; dashed circle, $2$\,cm around it.}
\label{fig:supp_tilt_pi05}
\end{figure*}

Across the $48$ chunks, \cref{eq:tilt} accounts for $99.98\%$ of the variation in the log-ratio (\cref{fig:supp_tilt_pi05}).
The median absolute error is $10$ nats, against a median log-ratio magnitude of $1{,}675$ nats.
The ratios are this large because \pifive{} is nearly deterministic for a given observation, so steering moves almost all probability to a different set of chunks.
The error decreases as the integration grid is refined ($1.9$, $1.0$ and $0.34$ nats with $16$, $24$ and $32$ steps on one chunk), so it reflects numerical integration rather than the formula.
Chunks from the continuous-time sampler differ from those of the deployed $10$-step sampler by at most $0.24$ in any normalized action component; on the first call, steering moves the chunk by $3.6$ in Euclidean norm.

\subsection{Where a change enters the flow}
\label{sec:supp_gain}
We derive \cref{eq:gain} and relate it to \cref{eq:tilt}.
Fix an observation and a sampling path $\vect{x}_\tau$ of the unsteered flow from noise $\vect{x}_1=\vect{\varepsilon}$ to the chunk $\vect{a}=\vect{x}_0$, with density $p_\tau$ at flow time $\tau$.
Let $\Psi_\tau=\partial\vect{a}/\partial\vect{x}_\tau$ and $n=Hd$.

\paragraph{Displacements.}
The first identity in \cref{eq:gain} is the definition of $\Psi_1$.
For the second, the steered path obeys $\dot{\vect{x}}=\vect{v}_\theta+\Delta\vect{v}$, integrated from $\tau=1$ down to $0$.
Linearizing around the unsteered path, the deviation $\vect{\xi}_\tau$ satisfies $\dot{\vect{\xi}}=(\partial\vect{v}_\theta/\partial\vect{x})\,\vect{\xi}+\Delta\vect{v}_\tau$ with $\vect{\xi}_1=\vect{0}$, and variation of constants gives $\vect{\xi}_0=-\int_0^1\Psi_\tau\,\Delta\vect{v}_\tau(\vect{x}_\tau)\,d\tau$.

\paragraph{Relation to \cref{eq:tilt}.}
To first order the steered chunk is $\vect{a}+\Delta\vect{a}(\vect{a})$, and the change of variables gives $\log p_{\vect{\alpha}}(\vect{a}+\Delta\vect{a})=\log p(\vect{a})-\nabla\!\cdot\Delta\vect{a}$, hence $\log f_{\vect{\alpha}}=-\Delta\vect{a}^{\top}\nabla\log p-\nabla\!\cdot\Delta\vect{a}$ to first order.
Substituting $\Delta\vect{a}_{\mathrm{steer}}$ recovers the integrand of \cref{eq:tilt}.
The score at time $\tau$ is $\nabla\log p_\tau=\Psi_\tau^{\top}\nabla\log p+\nabla\log|\det\Psi_\tau|$, and the divergence of the transported field $\Psi_\tau\Delta\vect{v}_\tau$ with respect to $\vect{a}$ equals $\nabla\!\cdot\Delta\vect{v}_\tau+\Delta\vect{v}_\tau^{\top}\nabla\log|\det\Psi_\tau|$, so the two expressions agree term by term.

\paragraph{Volume change.}
The flow from time $\tau$ to $0$ is invertible and carries $p_\tau$ to $p$, so $|\det\Psi_\tau|=p_\tau(\vect{x}_\tau)/p(\vect{a}\mid o,\ell)$, which at $\tau=1$ is $\mathcal{N}(\vect{\varepsilon})/p(\vect{a}\mid o,\ell)$.
The geometric mean of the $n$ singular values of $\Psi_1$ is therefore $g=\big(\mathcal{N}(\vect{\varepsilon})/p(\vect{a}\mid o,\ell)\big)^{1/n}$, while $\Psi_0$ is the identity.
In the check of \cref{sec:supp_theory_pi05}, the $24$ unsteered chunks have a median log-density of $1{,}715$ nats (range $1{,}658$ to $1{,}783$), and a standard Gaussian draw in $320$ dimensions has a log-density of about $-454$ nats, which gives $g\approx1.1\times10^{-3}$.
This is an average over all $n$ axes and not a bound on each one: a noise change can have a larger effect along a few axes, but which axes these are depends on the observation and on the noise draw.

\section{From the contrastive objective to the steering error}
\label{sec:supp_why_clr}
We relate the two terms of \cref{eq:clr} to the error of lookup (\cref{eq:lookup}).
Let $\vect{m}_k(o)$ be the endpoint shift that direction $k$ produces at unit strength in scene $o$, so that the measured effect of \cref{sec:naming} is its mean $\vect{m}_k=\mathbb{E}_o\,\vect{m}_k(o)$.
Let
\begin{equation}
V_k=\frac{\mathbb{E}_o\|\vect{m}_k(o)-\vect{m}_k\|^2}{\|\vect{m}_k\|^2}
\label{eq:cross_scene_variation}
\end{equation}
be the variation of the effect across scenes relative to its mean, and let $\theta_k$ be the angle between the required correction $\vect{d}$ and $\vect{m}_k$.
We assume that the endpoint shift is linear in the strength and that the strength is not clipped.

\paragraph{Error of lookup.}
With $\alpha_k=\vect{d}^{\top}\vect{m}_k/\|\vect{m}_k\|^2$, the expected squared distance left to the target is
\begin{equation}
\mathbb{E}_o\|\vect{d}-\alpha_k\vect{m}_k(o)\|^2=\|\vect{d}\|^2\big(\sin^2\theta_k+\cos^2\theta_k\,V_k\big).
\label{eq:lookup_error}
\end{equation}
To see this, write $\vect{d}-\alpha_k\vect{m}_k(o)=(\vect{d}-\alpha_k\vect{m}_k)-\alpha_k(\vect{m}_k(o)-\vect{m}_k)$.
The second term has zero mean, so the expected squared norm is $\|\vect{d}-\alpha_k\vect{m}_k\|^2+\alpha_k^2\,\mathbb{E}_o\|\vect{m}_k(o)-\vect{m}_k\|^2$.
The first part is the squared distance from $\vect{d}$ to the line through $\vect{m}_k$, which is $\|\vect{d}\|^2\sin^2\theta_k$, and $\alpha_k^2\|\vect{m}_k\|^2=\|\vect{d}\|^2\cos^2\theta_k$ gives the second.
The error thus has two sources: how far the nearest measured effect points from the required correction, and how much that direction's effect changes from scene to scene.

\paragraph{Positive pairs.}
Let $\tilde V_k=\mathbb{E}\|\vect{u}_k-\bar{\vect{u}}_k\|^2/\|\bar{\vect{u}}_k\|^2$ be the same relative variation, measured on the unit prediction changes that enter \cref{eq:clr}.
Since $\|\vect{u}_k\|=1$, we have $\mathbb{E}\|\vect{u}_k-\bar{\vect{u}}_k\|^2=1-\|\bar{\vect{u}}_k\|^2$, and for two rows with independently drawn observations and noise the expected similarity of a positive pair is
\begin{equation}
\mathbb{E}\big[\vect{u}_{k,r}^{\top}\vect{u}_{k,s}\big]=\|\bar{\vect{u}}_k\|^2=\frac{1}{1+\tilde V_k}.
\label{eq:positive_pairs}
\end{equation}
Raising the similarity of positive pairs is therefore the same as lowering $\tilde V_k$.
This identity holds for the prediction changes on which the objective acts; endpoint shifts follow from them to first order through \cref{eq:gain} and the map from actions to end-effector position, so $\tilde V_k$ and $V_k$ are related but not equal.

\paragraph{Negative pairs.}
For two different directions the expected similarity is $\bar{\vect{u}}_k^{\top}\bar{\vect{u}}_j$, so the negative term pushes the mean effects of different directions apart.
A pairwise penalty of this form is minimized by points spread uniformly over the sphere~\cite{wang2020understanding}.
The more spread out the mean effects are, the closer the nearest one lies to a required correction, so the direction that lookup selects has a smaller $\theta_k$.

\paragraph{Regret.}
The target is not known when the directions are learned.
We therefore measure a dictionary by the fraction of the squared correction that lookup leaves, averaged over scenes and over corrections $\vect{d}$ whose direction is uniform on the sphere.
By \cref{eq:lookup_error} this is
\begin{equation}
R_N=\mathbb{E}_{\vect{d}}\big[\sin^2\theta_{k^\star}+\cos^2\theta_{k^\star}\,V_{k^\star}\big],
\label{eq:regret}
\end{equation}
where $k^\star$ is the direction whose mean effect is nearest in angle to $\vect{d}$.
An oracle that could apply any correction would leave nothing, so $R_N$ is the regret of the dictionary against that oracle.

\paragraph{Proposition 2 (what a dictionary of $N$ directions should look like).}
For a fixed number of directions $N$, the regret $R_N$ is smallest when (i) each direction's effect is the same in every scene, $V_k=0$, and (ii) the mean effects are spread so that the nearest one is as close as possible to a correction of unknown direction.

\paragraph{Proof.}
Since $V_k\ge0$ and $\cos^2\theta\ge0$,
\begin{equation*}
R_N\;\ge\;\mathbb{E}_{\vect{d}}\,\min_k\sin^2\theta_k(\vect{d}),
\end{equation*}
with equality exactly when $V_k=0$ for every direction that lookup selects.
The right-hand side depends only on where the mean effects point: it is the average, over corrections, of how far the nearest mean effect lies from the correction.
It is smallest for the set of $N$ points on the sphere that leaves no correction far from its nearest point, which is the most even spread. \hfill$\square$

\paragraph{How the objective relates to the regret.}
The two conditions of Proposition 2 are the two terms of \cref{eq:clr}.
By \cref{eq:positive_pairs}, raising the similarity of positive pairs lowers the variation of an effect across scenes, which is condition (i).
The negative term pushes the mean effects of different directions apart and is minimized by points spread uniformly over the sphere, which is condition (ii).
If the objective could be optimized fully, the set it gives would therefore be, approximately, the minimum-regret set of size $N$.
The approximation is that the objective acts on normalized changes of the predicted velocity while \cref{eq:regret} is defined on end-effector shifts, so the statement is exact only if one maps to the other without distortion.

\paragraph{Scope.}
The argument covers lookup with one direction per call, without its strength cap, and assumes an effect that is linear in the strength.
Grad combines several directions and is not covered.
The relation between the objective and the regret is an argument about its two terms; it does not show that training reaches the smallest regret.

\section{Steering protocol and target construction}
\label{sec:supp_protocol}
\label{sec:sequencer}

\paragraph{Per-call target.}
During approach, the target is $5$\,cm above the current object in LIBERO and $6$\,cm above it in RoboLab.
During carry, the target is $10$\,cm above the placement location.
The target for one call is capped at $6$\,cm from the current end-effector position:
\begin{equation}
\vect{p}^{\star}_{\mathrm{call}}=\vect{p}_{\mathrm{EE}}+
\operatorname{clip}_{6\,\mathrm{cm}}(\vect{p}^{\star}-\vect{p}_{\mathrm{EE}}),
\end{equation}
where clipping limits the vector norm.
This capped target is the $\vect{p}^{\star}$ used by lookup and grad in \cref{sec:selectors}.
The simulation experiments use object and placement positions from the simulator.

\paragraph{Task phases.}
Each object-and-place subtask proceeds through approach, grasp, carry and release.
Among interchangeable objects, the nearest is selected at each new approach.
An object is considered held when it is lifted by at least $3$\,cm and is within $8$\,cm of the tool point or in contact.
A subtask completes when the benchmark's per-object success flag fires for an object previously observed as held.
A failed grasp or a drop outside the placement region returns the subtask to approach.
On RoboLab, lookup and grad steer during approach only.

\paragraph{Execution and path prediction.}
The LIBERO client executes $10$ actions before replanning for all four policies, even when the generated chunk is longer.
Lookup and grad score the endpoint of this executed prefix, not the end of a longer unused suffix.
For joint-position actions, forward kinematics gives the tool path.

\paragraph{Lookup and grad settings.}
Lookup uses a maximum strength of $20$ in the main LIBERO comparisons.
Grad uses two unit-normalized gradient steps of size $1$ on the strengths, bounds each strength to $[-3,3]$, and uses $\lambda=10^{-4}$ in \cref{eq:target}.
It starts from half of the previous call's strengths.
Every evaluated candidate is compared with the unsteered chunk using the same sampling noise; the lowest-objective candidate is used only if it improves on zero strengths.
The backbone or video conditioning is computed once per call and is not differentiated during this optimization.
For Cosmos Policy, gradients pass through the full joint denoiser, since there is no independent prefix cache.

\subsection{Target sources}
\label{sec:supp_targets}
On the real robot the target comes from a click, a vision-language model or direction keys.

\paragraph{Click.}
A person clicks the object or the destination in the exterior camera image.
The pixel is lifted to 3D by intersecting its camera ray with the table plane, and the target is placed a fixed height above that point.

\begin{center}
\includegraphics[width=\linewidth]{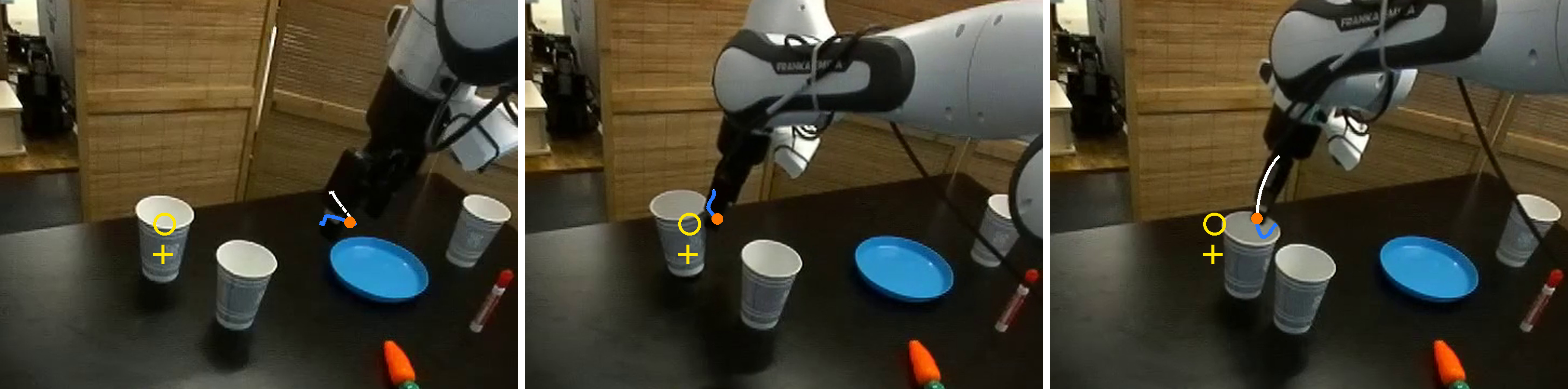}
\captionsetup{hypcap=false}
\captionof{figure}{\textbf{Target from a click.} Three moments of one run (``pick up the furthest cup''). Cross: the clicked pixel. Yellow ring: the target above it. Orange dot: the gripper. White: the policy's own chunk. Blue: the steered chunk.}
\captionsetup{hypcap=true}
\label{fig:supp_src_click}
\end{center}

\Cref{fig:supp_click_base} compares the base policy with the steered one for a clicked target.

\begin{center}
\includegraphics[width=\linewidth]{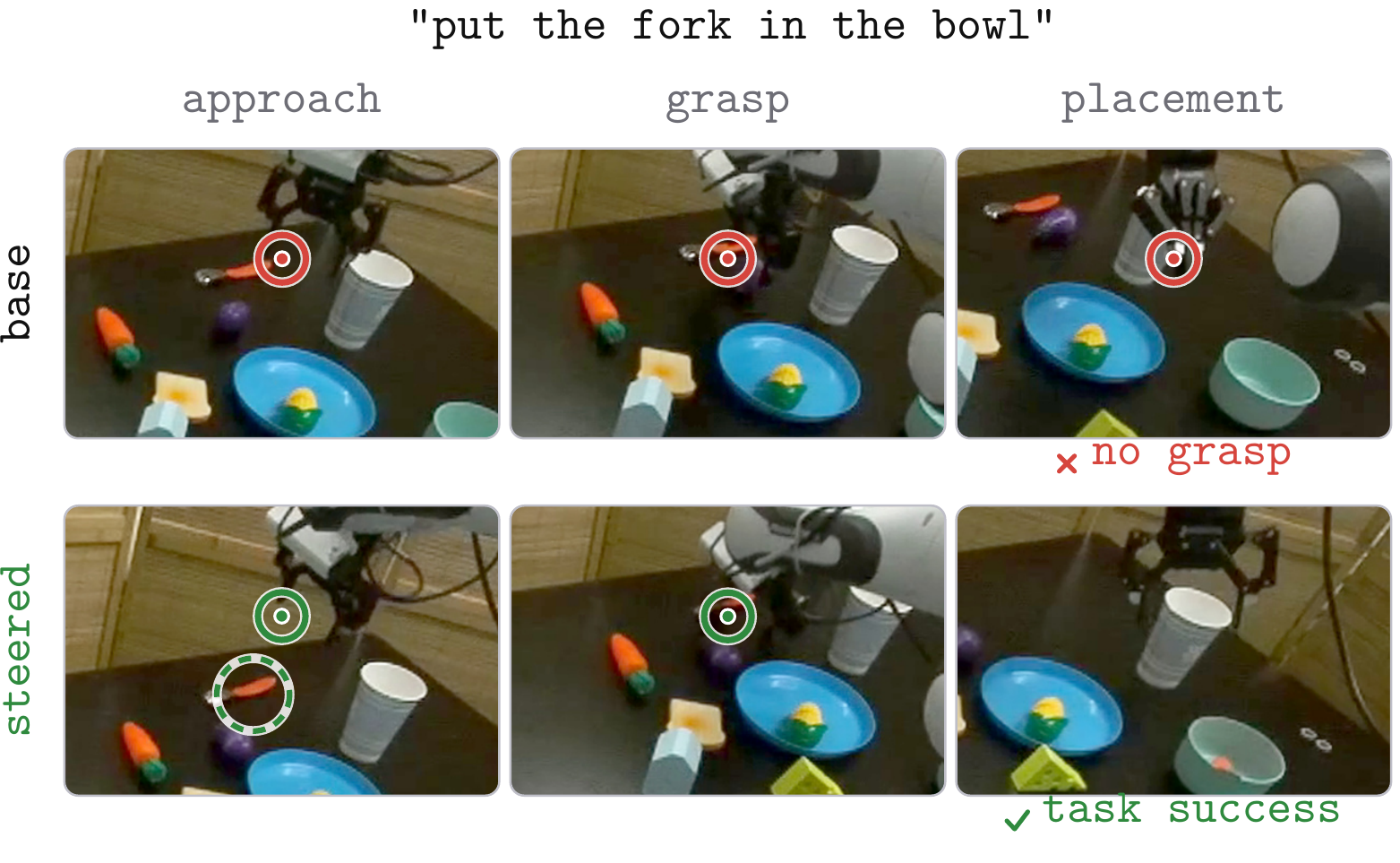}
\captionsetup{hypcap=false}
\captionof{figure}{\textbf{Base vs.\ steered with a clicked target} (same layout). Each panel is zoomed on the gripper (ring with dot). The base policy does not pick up the fork; steered toward the clicked fork (dashed ring), it picks it up and puts it in the bowl.}
\captionsetup{hypcap=true}
\label{fig:supp_click_base}
\end{center}

\paragraph{Vision-language model.}
A VLM splits the instruction into ``object $\to$ place'' steps and points to each in the image; the pointed pixel is lifted like a click.
We use Qwen3.8-27B~\cite{qwen38} as the VLM (\cref{sec:supp_vlm}).

\begin{center}
\includegraphics[width=\linewidth]{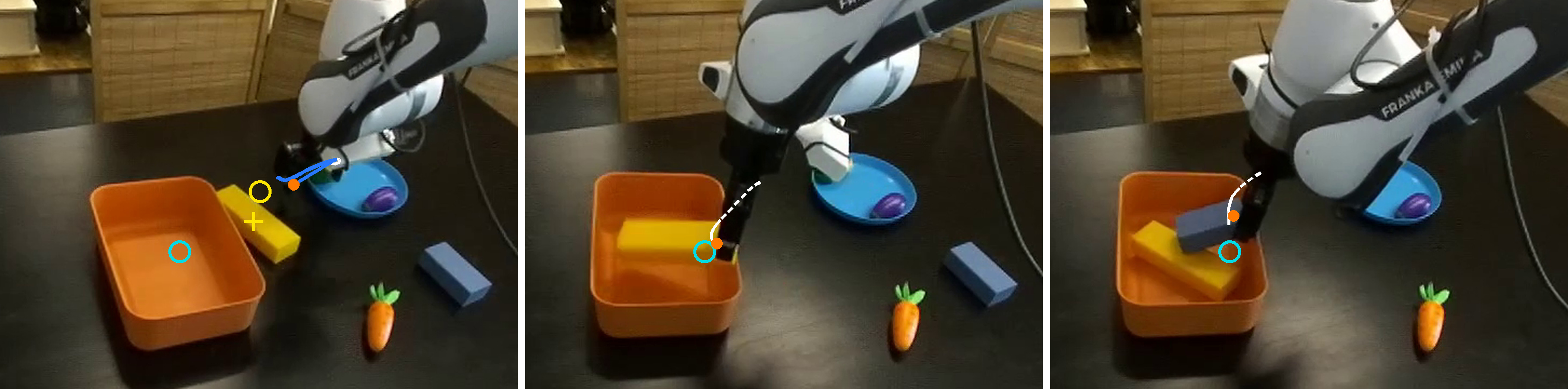}
\captionsetup{hypcap=false}
\captionof{figure}{\textbf{Targets from a vision-language model.} ``Put the blocks in the container'', marks as in \cref{fig:supp_src_click}. Cross: the block the VLM points to. Cyan ring: the pointed container. The blocks are placed in the planned order.}
\captionsetup{hypcap=true}
\label{fig:supp_src_vlm}
\end{center}

\begin{center}
\includegraphics[width=\linewidth]{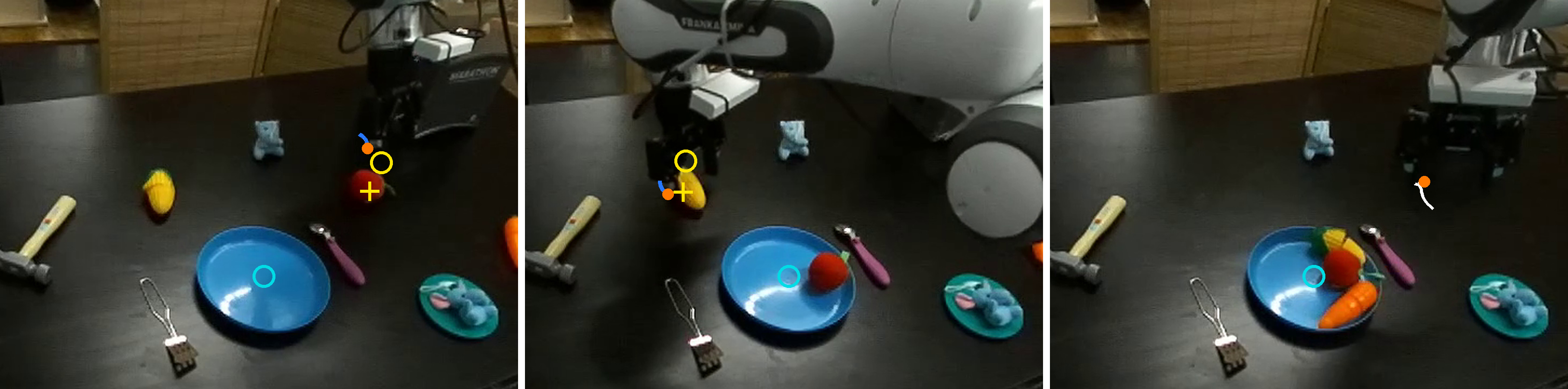}
\captionsetup{hypcap=false}
\captionof{figure}{\textbf{A second example.} ``Put the vegetables on the blue plate.'' The VLM names the tomato, the pepper and the carrot and points to each in turn.}
\captionsetup{hypcap=true}
\label{fig:supp_src_vlm2}
\end{center}

\noindent The planning and pointing query is a single call. For the run of \cref{fig:supp_src_vlm}, abridged:
\par\smallskip{\footnotesize\ttfamily\raggedright\noindent
Task: "put the blocks in the container". Break the task into pick-and-place steps in the order they must be done. [...] For every step also point at the object to pick and at its place, in this image. Answer with JSON only [...]
\par}\smallskip\noindent
and the answer, with the points in thousandths of the image size:
\par\smallskip{\footnotesize\ttfamily\raggedright\noindent
\{"plan":[\{"pick":"yellow block", "pick\_point":[549,433], "place":"orange container", "place\_point":[481,475]\},\\
\{"pick":"blue block", "pick\_point":[681,492], "place":"orange container", "place\_point":[481,475]\}]\}
\par}\smallskip\noindent
For the run of \cref{fig:supp_src_vlm2} the task is ``put the vegetables on the blue plate'' and the answer is
\par\smallskip{\footnotesize\ttfamily\raggedright\noindent
\{"plan":[\{"pick":"red tomato", "pick\_point":[626,424], "place":"blue plate", "place\_point":[549,536]\},\\
\{"pick":"yellow pepper", "pick\_point":[472,424], "place":"blue plate", "place\_point":[549,536]\},\\
\{"pick":"orange carrot", "pick\_point":[731,486], "place":"blue plate", "place\_point":[549,536]\}]\}
\par}\smallskip

\Cref{fig:supp_vlm_base} compares the base policy with the steered one on a scene of this kind.

\begin{center}
\includegraphics[width=\linewidth]{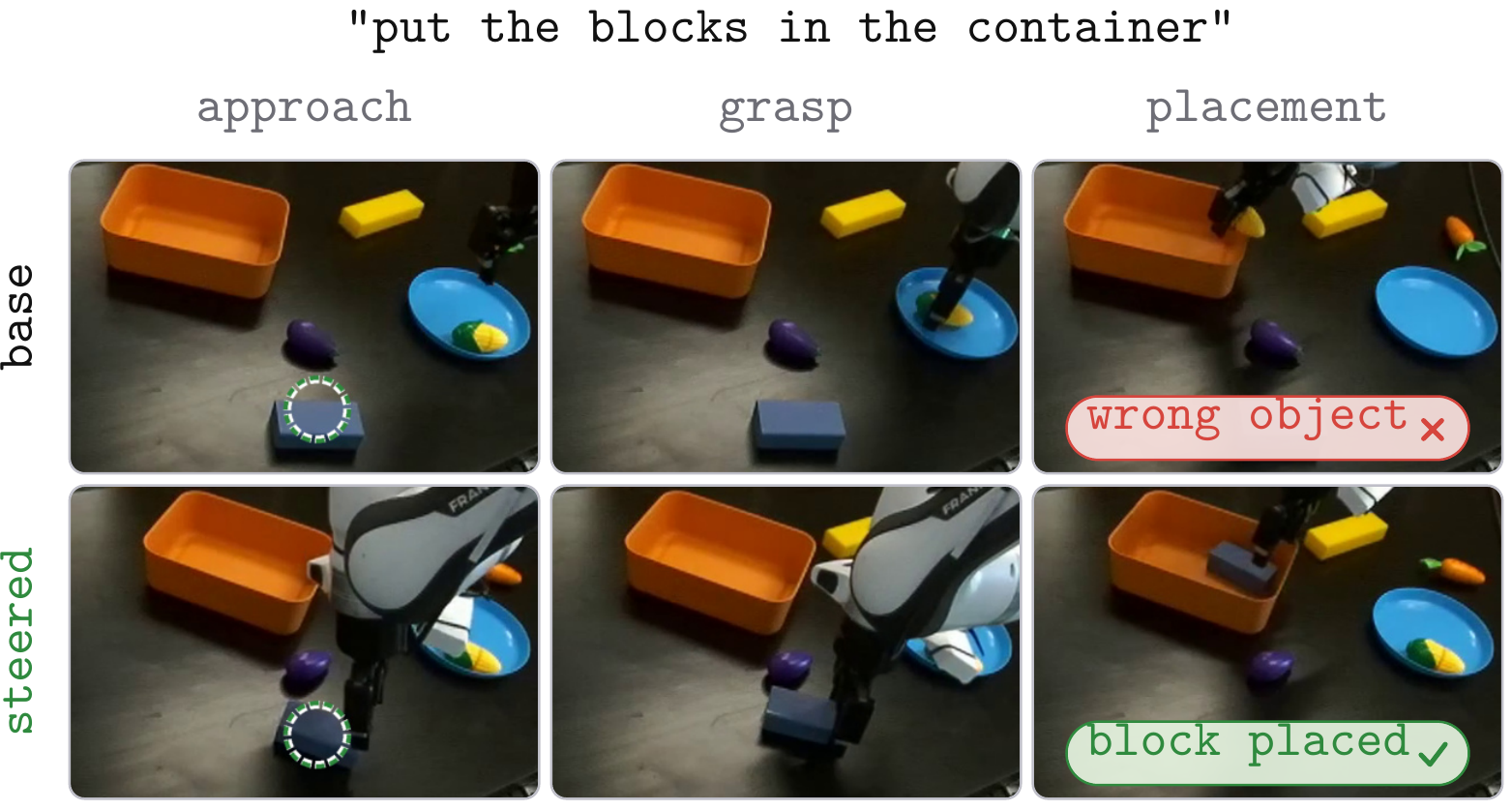}
\captionsetup{hypcap=false}
\captionof{figure}{\textbf{Base vs.\ steered with VLM targets} (same layout). The base policy takes the corn from the plate and puts it in the container; steered toward the block the VLM points to (dashed ring), it puts the block in.}
\captionsetup{hypcap=true}
\label{fig:supp_vlm_base}
\end{center}

\paragraph{Direction keys.}
A person presses one key per action chunk (\cref{fig:supp_src_keys}): W/S move the target forward/back, A/D left/right and Q/E up/down, in the robot base frame.
The target for that call is the end of the policy's unsteered chunk plus one step along the key's axis.
Without a key the call is not steered.
Keys need no camera calibration.

\begin{center}
\includegraphics[width=\linewidth]{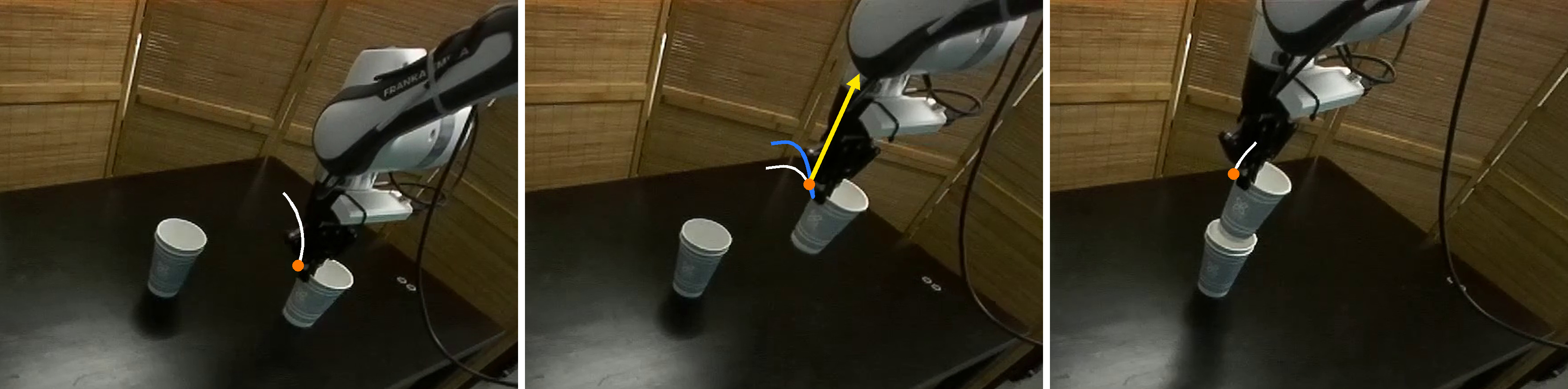}
\captionsetup{hypcap=false}
\captionof{figure}{\textbf{Direction keys.} ``Stack the cups.'' Middle: the policy's own chunk (white) heads sideways toward the other cup, and the up key (yellow arrow) lifts the steered chunk (blue) above it.}
\captionsetup{hypcap=true}
\label{fig:supp_src_keys}
\end{center}

\paragraph{Carry.}
Once the object is held, the target is the destination (\cref{fig:supp_src_carry}).
An operator can also select which target is active, the object or the destination.

\begin{center}
\includegraphics[width=\linewidth]{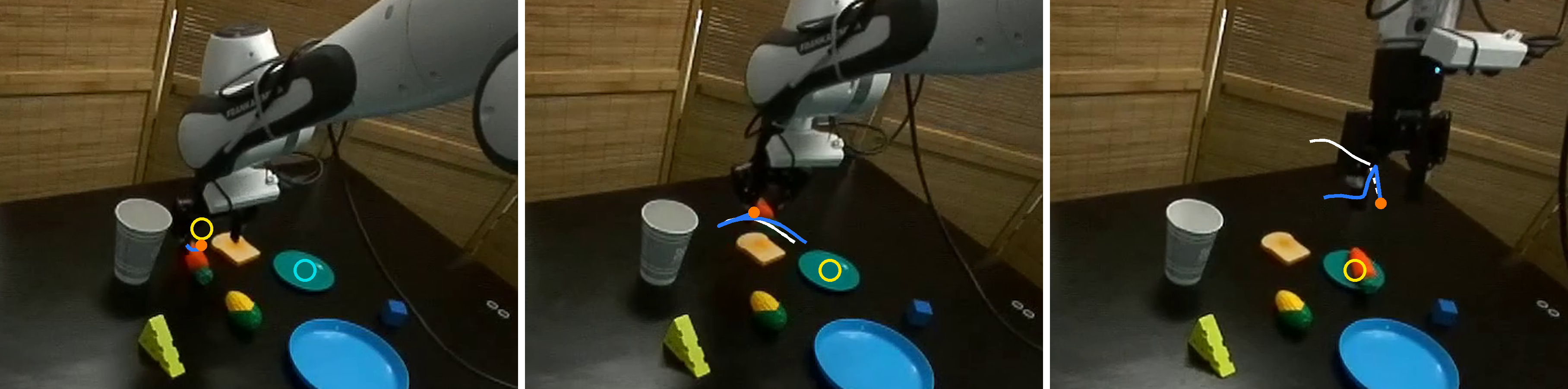}
\captionsetup{hypcap=false}
\captionof{figure}{\textbf{Carry.} ``Put the carrot on the green plate.'' Left: the target is above the carrot (yellow ring); the destination is not yet active (cyan ring). Middle and right: with the carrot held, the operator makes the destination the active target.}
\captionsetup{hypcap=true}
\label{fig:supp_src_carry}
\end{center}

\begin{figure}[t]
\centering
\includegraphics[width=\linewidth]{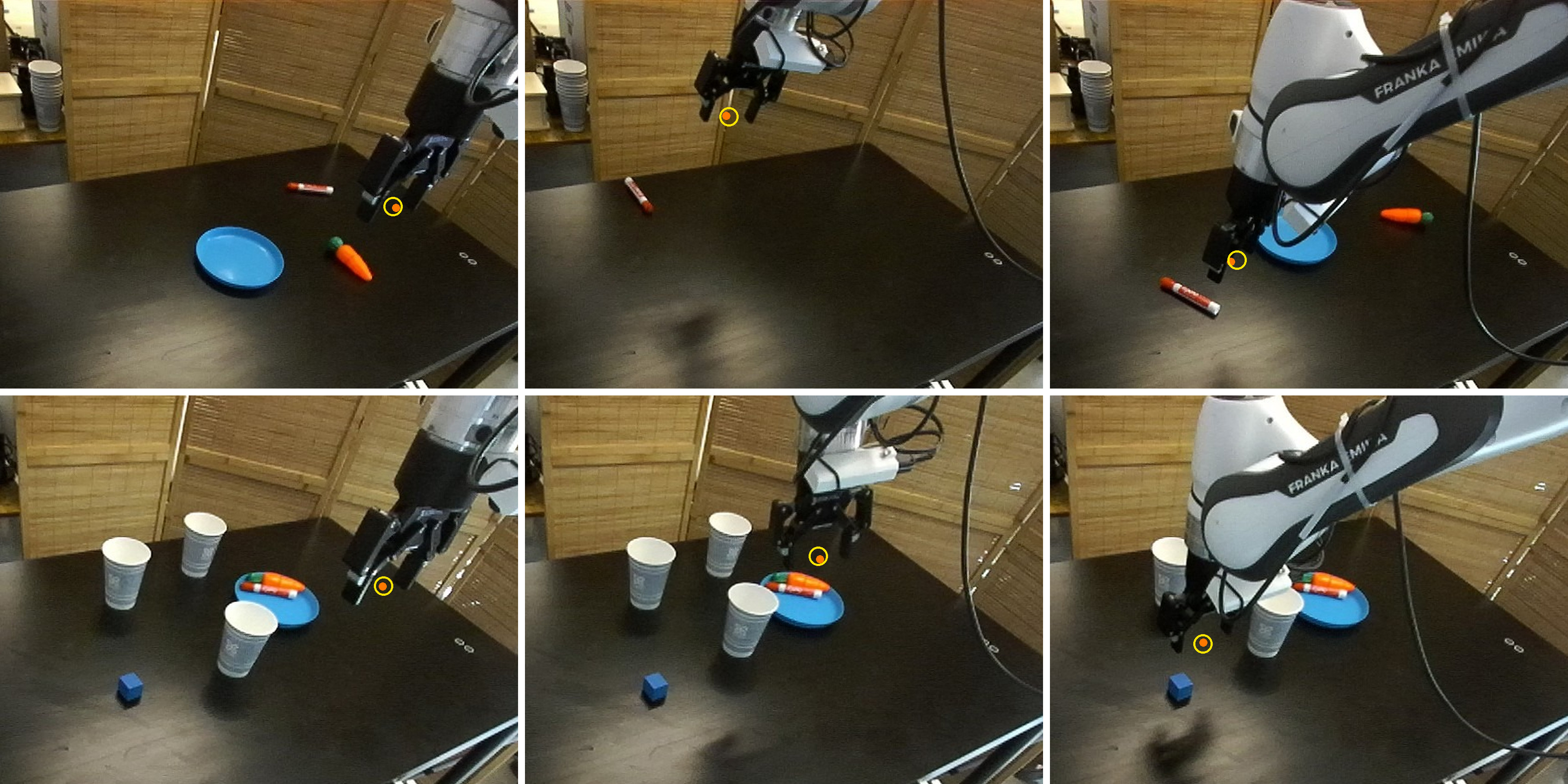}
\caption{\textbf{Example calibration clicks.} Yellow ring: the clicked point between the fingertips. Orange dot: the tool point from forward kinematics, projected into the image with the solved camera pose.}
\label{fig:supp_calib}
\end{figure}

\paragraph{Camera calibration.}
The exterior camera is calibrated without a board.
The arm is stopped at several poses, and at each a person clicks the point between the fingertips in the image (\cref{fig:supp_calib}).
The clicked pixels and the tool points from forward kinematics give the camera pose by PnP, with the camera's own intrinsics.
In total we clicked $37$ points.

\begin{figure}[t]
\centering
\includegraphics[width=\linewidth]{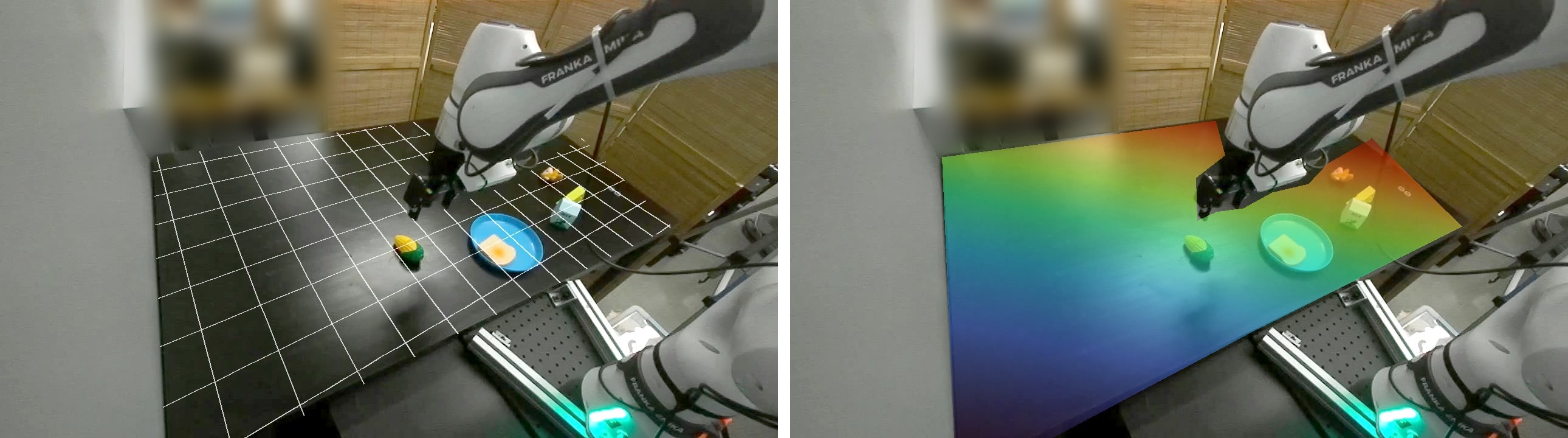}
\caption{\textbf{The table plane from the calibration.} Left: a grid on the plane, drawn along the axes of the robot base. Right: depth of the plane from the camera (blue near, red far).}
\label{fig:supp_table_plane}
\end{figure}

The calibrated pose places the table plane in the image (\cref{fig:supp_table_plane}).
A clicked or VLM-pointed pixel is lifted to 3D by intersecting its camera ray with this plane.

\section{Real-robot evaluation}
\label{sec:supp_real}

\paragraph{Setup.}
The robot is a Franka arm with a Robotiq gripper in the DROID configuration, with one exterior camera and one wrist camera.
The policy is \pifive-DROID with joint-velocity actions at $15$\,Hz, and its weights are unchanged.
\ours uses the grad selector with the default dictionary ($N{=}64$, $K{=}4$).
FRS~\cite{tang2026frs} is our implementation, run on the same policy server and given the same kinds of target as \ours (\cref{sec:supp_targets}).

\paragraph{Protocol.}
A layout is one arrangement of the objects on the table.
We report only layouts on which the policy alone, FRS and \ours were all run, which gives $47$, $41$ and $45$ trials; a method can have more than one trial on a layout.
An author marked each trial after the run as a success, a partial success (part of the task done, such as one of two objects placed) or a failure, knowing the method.
The score counts a partial success as half.
Counting full successes only, the three methods reach $28\%$, $34\%$ and $67\%$.

\paragraph{Results per task.}
\Cref{tab:supp_real} lists every task.
\ours has the highest score on eight of the ten tasks.
The policy alone is ahead on ``put the fork and spoon on the blue plate'', which it already solves, and ties with \ours on ``put the blocks in the container''.
\Cref{fig:supp_real_far,fig:supp_real_stack} show one layout of each cup task.

\begin{center}
\small
\setlength{\tabcolsep}{4pt}
\resizebox{\linewidth}{!}{%
\begin{tabular}{@{}lccc@{}}
\toprule
Task & \pifive & FRS & \ours \\
\midrule
blocks in the container & \fst{75} (4) & 62 (4) & \fst{75} (6) \\
blue cube in the cup & 42 (6) & 80 (5) & \fst{100} (3) \\
carrot on the green plate & 8 (6) & 50 (4) & \fst{75} (4) \\
corn on the blue plate & 0 (4) & 0 (4) & \fst{90} (5) \\
corn on the green plate & 0 (2) & 25 (2) & \fst{50} (2) \\
fork and spoon on the blue plate & \fst{100} (3) & 83 (3) & 75 (4) \\
fork in the bowl & 38 (4) & 33 (3) & \fst{83} (3) \\
monkey on the blue plate & 40 (5) & 33 (3) & \fst{100} (3) \\
\midrule
pick and place, all & 37 (34) & 48 (28) & \fst{82} (30) \\
pick up the furthest cup & 0 (6) & 14 (7) & \fst{62} (8) \\
stack the cups & 36 (7) & 67 (6) & \fst{86} (7) \\
\midrule
all tasks & 32 (47) & 45 (41) & \fst{79} (45) \\
\bottomrule
\end{tabular}}
\captionsetup{hypcap=false}
\captionof{table}{\textbf{Real-robot score per task} (\%), with the number of trials in parentheses. A partial success counts as half.}
\captionsetup{hypcap=true}
\label{tab:supp_real}
\end{center}

\begin{center}
\includegraphics[width=\linewidth]{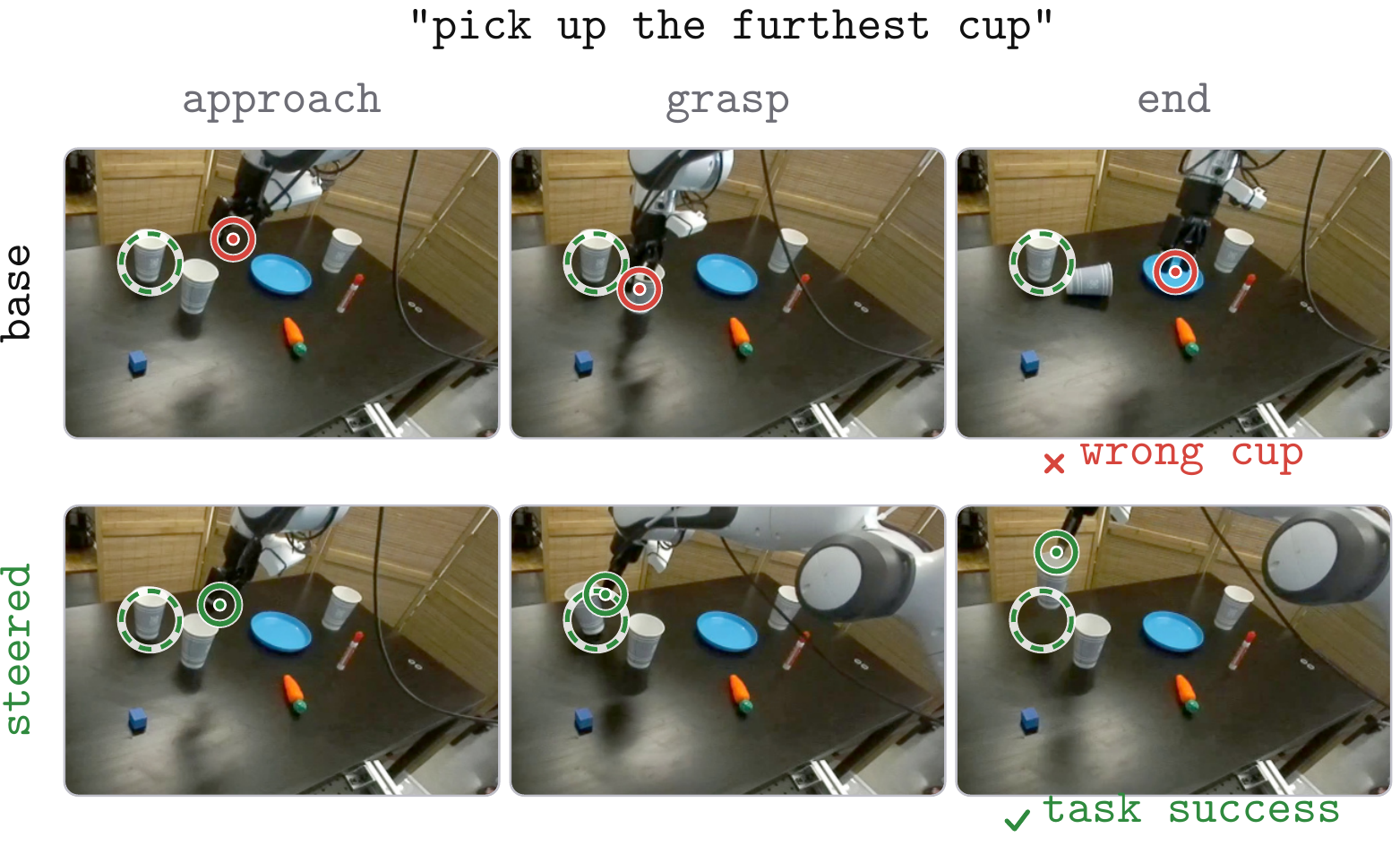}
\captionsetup{hypcap=false}
\captionof{figure}{\textbf{Furthest cup, base vs.\ steered} (same layout). Dashed ring: the furthest cup. The base policy goes to a nearer cup and knocks it over; steered toward the clicked cup, it lifts the furthest one.}
\captionsetup{hypcap=true}
\label{fig:supp_real_far}
\end{center}

\begin{center}
\includegraphics[width=\linewidth]{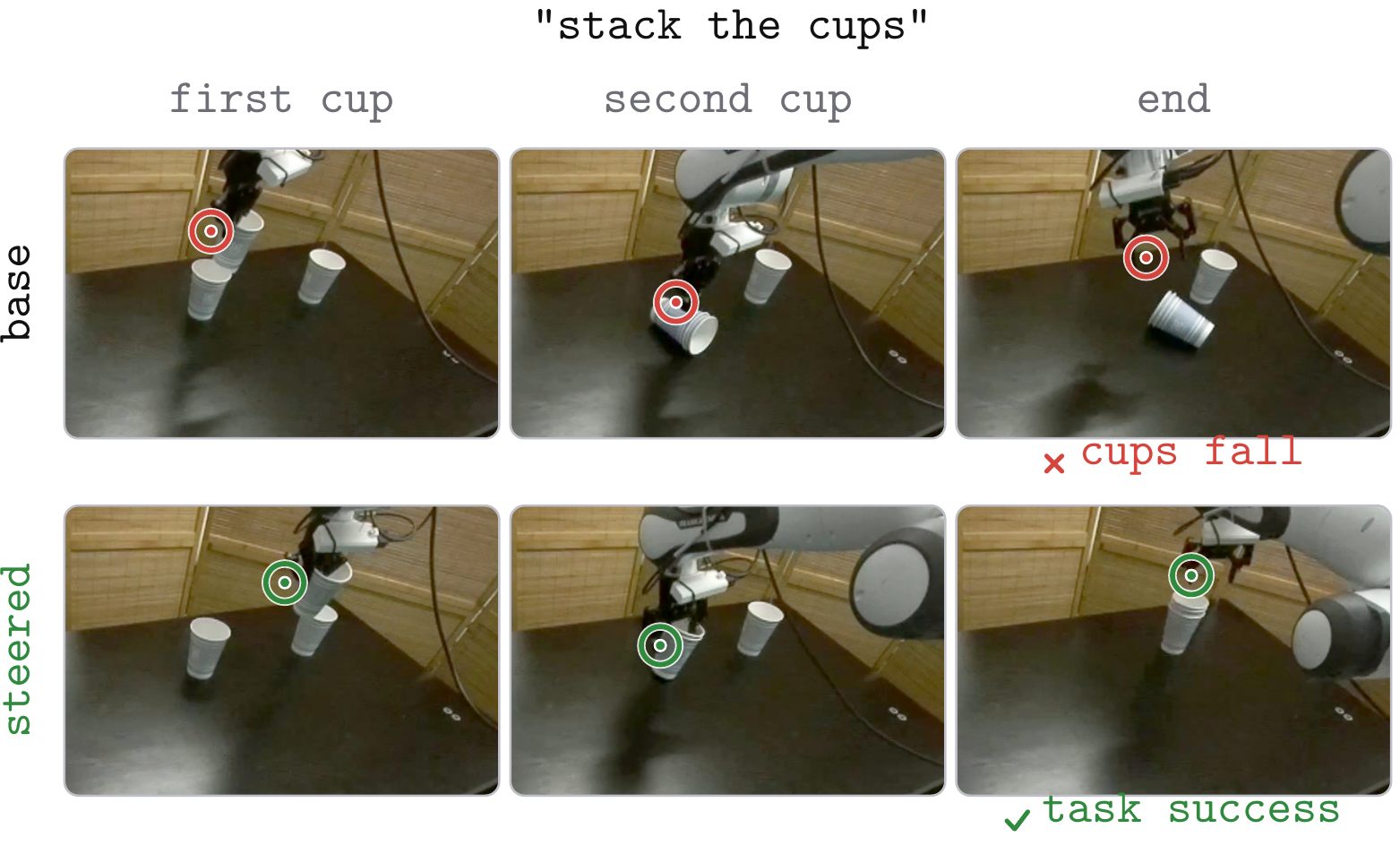}
\captionsetup{hypcap=false}
\captionof{figure}{\textbf{Cup stacking, base vs.\ steered} (same layout). The base policy knocks the cups over; steered with direction keys, it stacks all three.}
\captionsetup{hypcap=true}
\label{fig:supp_real_stack}
\end{center}

\section{Vision-language model queries}
\label{sec:supp_vlm}

On the real robot the vision-language model is Qwen3.8-27B~\cite{qwen38}.
Each query is one user message with the camera image and a text prompt, without a system prompt, at temperature zero.
\Cref{tab:supp_vlm_prompts} lists the prompts.

\paragraph{Plan and point.}
At the start of a task the model splits the instruction into pick-and-place steps and points to the object and the place of each step in the same answer (examples in \cref{sec:supp_targets}).

\paragraph{Point.}
During the task the model is asked again to point to the same objects and places, so that the targets follow objects that have moved.

\paragraph{State.}
The model sees the exterior and wrist images and reports where the object of each step is: on the table, in the gripper, or at its place.
This moves the plan to the next step.

\paragraph{Height.}
The model estimates how tall each destination stands above the table, and the destination target is raised by that amount.

\begin{table*}[t]
\centering
\scriptsize
\setlength{\tabcolsep}{4pt}
\begin{tabular}{@{}p{0.07\linewidth}p{0.10\linewidth}p{0.62\linewidth}p{0.15\linewidth}@{}}
\toprule
Query & When & Prompt (verbatim) & Use of the answer \\
\midrule
\raggedright Plan and point & \raggedright once at the start of a task & \raggedright \texttt{You plan for a robot arm with one two-finger gripper working on a table. The image is the side camera view of the table.} \newline \texttt{Task: "\{\textit{task}\}"} \newline \texttt{Break the task into pick-and-place steps in the order they must be done. One step = pick up one object and put it at one place.} \newline \texttt{- Name things the way a person would point at them in this image: colour + object ("red block", "blue plate", "paper cup").} \newline \texttt{- "place" is the object or container that the picked object goes on or in. If the task only asks to pick something up, use null.} \newline \texttt{- Stacking N objects takes N-1 steps: the bottom object stays where it is and is never picked.} \newline \texttt{- Only steps the task asks for. Do not add steps for other objects on the table.} \newline \texttt{- When the task leaves the order free (e.g. several objects that go to the same plate), pick the object farthest from the camera first (the one highest up in the image) and the nearest one last. Never change an order the task needs (stacking, "first ... then ...").} \newline \texttt{- "say" is what you would tell the robot for this one step alone: one short plain sentence, e.g. "put the blue cube in the paper cup" or "pick up the fork".} \newline \texttt{For every step also point at the object to pick and at its place, in this image.} \newline \texttt{Answer with JSON only, no other text: \{"plan":[\{"pick":"<object>","pick\_point":[x, y],"place":"<object or container, or null>","place\_point":[x, y],"say":"<instruction for this step>"\}]\}} & \raggedright ordered steps; each point is converted to a pixel and lifted to a 3D target \tabularnewline
\addlinespace
\raggedright Point & \raggedright during the task & \raggedright \texttt{The image is the side camera view of a table with a robot arm. Task: "\{\textit{task}\}"} \newline \texttt{The things below are objects on the table that the task is about, never a part of the robot.} \newline \texttt{Point to each of these things in the image:} \newline \texttt{\{\textit{names}\}} \newline \texttt{Answer with JSON only, one entry per thing, in the same order: [\{"label": "<thing>", "point\_2d": [x, y]\}]. Use "point\_2d": null for a thing that is not in the image.} & \raggedright updated targets for the objects and places of the plan \tabularnewline
\addlinespace
\raggedright State & \raggedright during the task, exterior and wrist image & \raggedright \texttt{You watch a robot arm with a two-finger gripper doing a task on a table.} \newline \texttt{Image 1 is the side camera. Image 2 is the wrist camera, looking down past the two gripper fingers at the table below.} \newline \texttt{Task: "\{\textit{task}\}"} \newline \texttt{Steps, in order:} \newline \texttt{\{\textit{steps}\}} \newline \texttt{The robot's own sensor says the gripper is \{\textit{grip}\} right now.\{\textit{grip\_note}\}} \newline \texttt{For the object of every step, say where it is now:} \newline \texttt{- "table": it rests on the table (or anywhere that is not its place) and the gripper does not hold it. Seeing it on the table next to or below the gripper counts as "table".} \newline \texttt{- "gripper": it is clamped between the two fingers, lifted off the table or just being lifted.} \newline \texttt{- "place": it rests at the place of its step and the gripper has let go of it. If the place is a container (cup, bowl, box, bin, mug) and the object can no longer be seen on the table or in the gripper, it is inside the container: "place".} \newline \texttt{Find each object in image 1 first; use image 2 to confirm. Answer with JSON only, no other text:} \newline \texttt{\{"where":[\{\textit{flags}\}], "other":<true if the gripper clamps an object that belongs to no step, else false>, "why":"<at most 12 words>"\}} & \raggedright moves the plan to the next step \tabularnewline
\addlinespace
\raggedright Height & \raggedright once per destination & \raggedright \texttt{The image is a side camera view of a table with a robot arm. For each of these things on the table, estimate how tall it stands above the table surface, in centimetres (a flat plate is about 2, a block about 3, a bowl about 6, a paper cup about 10):} \newline \texttt{\{\textit{names}\}} \newline \texttt{Answer with JSON only, no other text: \{"heights\_cm":[\{\textit{slots}\}]\}} & \raggedright height of the destination target \tabularnewline
\bottomrule
\end{tabular}
\caption{\textbf{Prompts sent to the vision-language model.} Each query is one user message with the image and this text; no system prompt is used. Italic names in braces are filled at run time: the task, the step list, the object names and the gripper reading.}
\label{tab:supp_vlm_prompts}
\end{table*}

\section{Targets from a pointing model in simulation}
\label{sec:supp_molmo}
The simulated experiments take the steering target from the simulator's object poses.
Here we obtain it from a pointing model instead.
At every policy call, Molmo-7B-D~\cite{deitke2024molmo} points at the object to grasp, or at the place once the object is lifted, and the point is lifted to 3D with the simulator's depth and camera calibration.
On Hard-62 with \pifive-LIBERO, \ours\ (grad) reaches $17.1\%$ with these targets against $5.0\%$ for the base policy, and keeps most of the $20.3\%$ it reaches with simulator targets (\cref{tab:supp_molmo}).

\vspace{2pt}
\begin{center}
\small
\begin{tabular}{lcc}
\toprule
 & \multicolumn{2}{c}{Target from} \\
\cmidrule(lr){2-3}
Method & Molmo & Simulator \\
\midrule
Base policy (\pifive) & \multicolumn{2}{c}{$5.0$} \\
+ FRS & $6.5$ & $5.5$ \\
+ \ours\ (lookup) & $11.3$ & $16.6$ \\
+ \ours\ (grad) & $\mathbf{17.1}$ & $\mathbf{20.3}$ \\
\bottomrule
\end{tabular}
\captionof{table}{\textbf{Steering with targets from a pointing model.} Hard-62 success (\%) with \pifive-LIBERO, ten episodes per task, same initial states in both columns.}
\label{tab:supp_molmo}
\end{center}

\section{Baseline implementations}
\label{sec:supp_baselines}

\paragraph{Direct reaching with \pifive.}
\label{sec:supp_reach}
Reach moves the gripper straight toward the target during approach and carry. Within $8$\,cm of the target it returns control to the policy. On LIBERO, Reach divides the current Cartesian target error by $5$\,cm, clips each component to $[-1,1]$ and updates the translation command at each executed step. It holds orientation fixed and uses the policy's gripper command.
On RoboLab, Reach maps Cartesian error to joint increments once per policy call with a damped Jacobian pseudo-inverse, spreads the increments over $15$ steps and holds the gripper at its current reading.

\begin{figure}[!htbp]
\centering
\includegraphics[width=\linewidth]{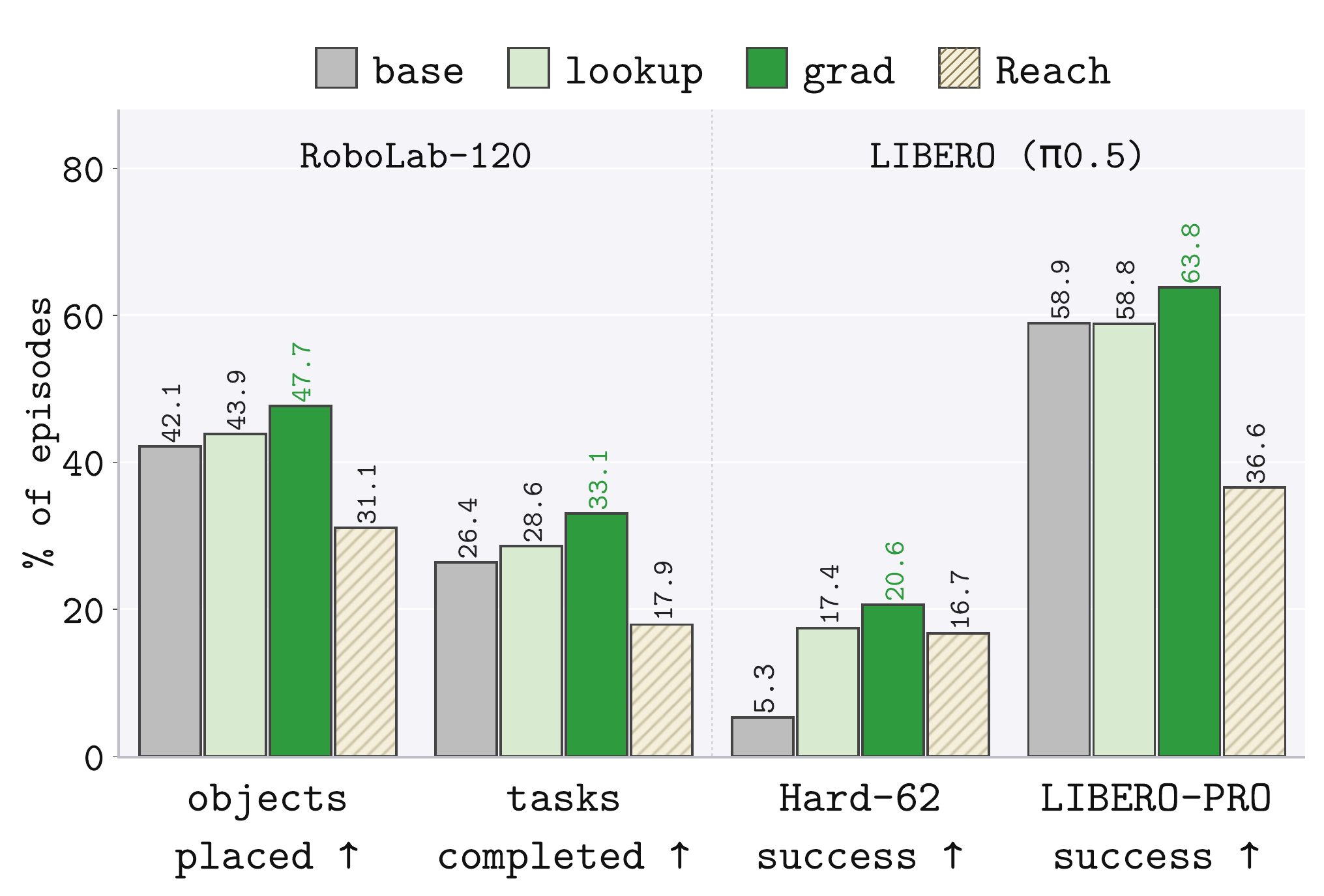}
\caption{\textbf{Direct reach on RoboLab-120 and LIBERO.} Per-episode outcomes on RoboLab-120 and \pifive success on Hard-62 and LIBERO-PRO. Reach falls below grad on all four, and below the base policy on RoboLab and LIBERO-PRO.}
\label{fig:reach_carry}
\end{figure}

On RoboLab, Reach places $31.1\%$ of objects and completes $17.9\%$ of tasks overall. The corresponding object and task rates by complexity are $26.2/16.9\%$ (Simple), $42.3/24.4\%$ (Moderate) and $24.0/6.9\%$ (Complex).
Paired sign tests over the $720$ task-seed episodes favor both the base policy ($p=1.3\times10^{-6}$) and grad ($p=4.2\times10^{-17}$) over Reach. \Cref{fig:reach_carry} shows the outcomes.

\paragraph{FRS on flow policies.}
We follow the reference interfaces of FRS~\cite{tang2026frs}, using the direction to the shared step target in place of a human or VLM command.
On DROID, a reference of five $7.5$\,mm Cartesian steps is converted to joint targets with a damped tool-Jacobian pseudo-inverse while holding orientation, then tiled over the chunk.
Padding dimensions are set to $-1$.
On LIBERO, the direction is written into a normalized Cartesian reference chunk.
The reference is integrated backward through the frozen flow and then denoised normally.
For DROID, only the recovered noise for the first five action rows is retained; later rows, padding and gripper coordinates receive fresh noise.
The VLA policies use $10$ reverse Euler steps; FastWAM uses its own shifted sampling grid.

\paragraph{FRS on Cosmos Policy.}
There is no flow-matching action expert to reverse.
Our adaptation starts from the base sample at noise level $4$, replaces its action frame with the reference action plus the estimated residual noise, and approximately reverses the diffusion path to noise level $80$.
The resulting action-frame noise is inserted into the original call's noise tensor; the other frames keep their original noise.
The native sampler then generates the chunk.
The WAM table identifies this as an approximate diffusion adaptation of FRS.

\paragraph{Flow Control.}
For LIBERO, we write $1.5$ times the unit target direction into the translation coordinates of the first $10$ noise rows and set their rotation coordinates to zero.
Other coordinates retain the call's sampled noise.
FastWAM uses the same command in its action-noise tensor.
For Cosmos Policy, it is written into every tiled copy of the action frame before the initial noise is scaled by $80$.
The policy then samples with its native schedule~\cite{kao2026flowcontrol}.

\begin{figure*}[t]
\centering
\includegraphics[width=0.82\textwidth]{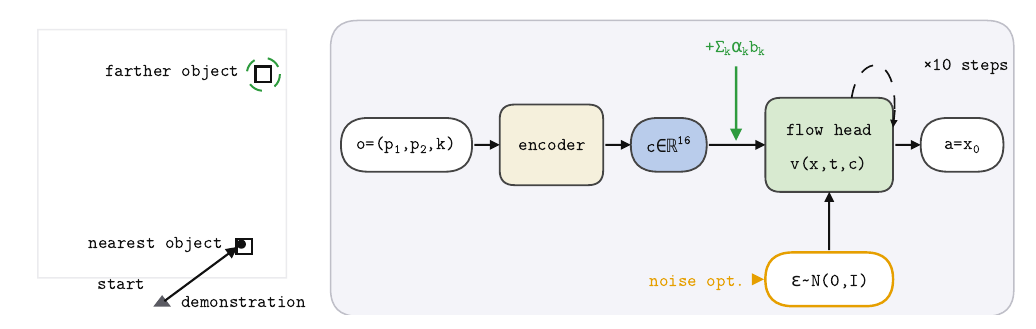}\\[4pt]
\parbox{0.82\textwidth}{\small \textbf{The 2D toy policy of \cref{sec:supp_toy2d}.} Left: a training scene, with the demonstration ending at the nearest object. Right: the policy. Learned directions are added to the conditioning at each of the ten sampler steps (green); noise optimization changes the starting noise (orange).}
\end{figure*}

\paragraph{AFI.}
Our simplified LIBERO reimplementation follows the trap detection, rollback and waypoint-search structure of AFI~\cite{xu2025afi}.
It receives the same simulator target as other target-based methods; other movable-object positions provide the obstacle terms.
The affordance cost combines distance to the target with repulsion within $5$\,cm of other objects.
A trap is detected when the gripper moves less than $2$\,cm over six calls while far from the target, or closes for two consecutive calls during an incorrect approach.
The controller returns to a recent low-cost position, samples nearby waypoints and ranks policy-generated chunks by accumulated field cost.
It samples $60$ candidate waypoint positions within $10$\,cm and evaluates eight chunks at each retained waypoint.
After executing a selected chunk, it waits three calls before another recovery, with at most two recoveries per phase visit.
This version replaces the paper's VLM and depth-based geometry with simulator information and the shared task sequencer.

\paragraph{COAST.}
Our reimplementation fits separate conceptors to successful and failed base-policy rollout activations~\cite{miao2026coast}.
A contrastive conceptor keeps the success-associated subspace while suppressing the failure-associated subspace.
At each denoising step it filters the action expert's residual activations at a selected layer.
On LIBERO, this is the third of four evenly spaced recorded layers, counted from $0$: layer $11$ of $18$ for \pifive and \pizero (the \pifive setting of~\cite{miao2026coast}), block $19$ of $30$ in FastWAM's action expert, and the action-frame tokens after DiT block $18$ of $28$ in Cosmos Policy.
All four policies use the conceptor aperture $\alpha=0.5$ that~\cite{miao2026coast} use for \pifive.
The fit uses separate conceptors per denoising step.
With fewer than three successful fitting rollouts, it uses the complement of the failure conceptor; with fewer than three failures, it uses the success conceptor alone.
COAST is applied at every policy call and receives no target point.

\section{Initial-noise and transfer studies}
\label{sec:supp_initial_noise}

\subsection{Two-dimensional toy policy}
\label{sec:supp_toy2d}
\paragraph{Training and evaluation.}
The observation contains two object positions and a one-hot instruction naming one of them.
Object positions are drawn within $\pm15$\,cm, and the gripper starts at $(0,-18)$\,cm.
Each demonstration ends at the nearest object, with isotropic Gaussian noise of standard deviation $0.4$\,cm.
Thus, the demonstrations never select the instructed object when it is farther away, although reaching positions elsewhere in the workspace occurs in other scenes.
We train one policy with one training seed using fresh demonstrations at each of $12{,}000$ steps, with batches of $1{,}024$.
The trained policy has an endpoint spread of $0.09$\,cm.

We evaluate $300$ calls where the instruction names the farther object.
Each call receives an external target $6$\,cm from the gripper toward that object and succeeds if its predicted endpoint is within $2$\,cm of this target.
Methods use the same scenes and matched initial noise.
The test panel of \cref{fig:toy2d} shows $300$ noise draws in one scene at five times the training panels' scale, with arrows from the gripper start to the mean endpoint of the policy alone and of \ours.
The plot aggregates the separate $300$-call evaluation.
This is a controlled reaching model with a deliberately biased demonstrator.

\paragraph{Policy and direction discovery.}
An MLP encoder with widths $6\to64\to64\to16$ maps the observation to conditioning $c$.
A velocity MLP with three $192$-unit hidden layers predicts a two-dimensional field conditioned on $c$ and time.
For a demonstrated endpoint $a$ and Gaussian noise $\varepsilon$, rectified-flow training uses $x_t=t\varepsilon+(1-t)a$ and velocity target $\varepsilon-a$.
The sampler starts at $x_1=\varepsilon\sim\mathcal{N}(0,I_2)$ and takes ten Euler steps, $x\leftarrow x-0.1v_\theta(x,t,c)$.
Its differentiability permits optimization of either the starting noise or steering strengths. The diagram at the top of this page shows the policy and where each method acts.
We learn six steering directions on training observations with the contrastive objective of \cref{sec:objective}.
The objective compares endpoint changes produced with and without each direction under the same noise.
It encourages consistent changes across observations and distinct changes across directions, without targets, rewards or motion labels.
At inference, as in the robot experiments, we add the weighted directions at every denoising step and take two gradient steps on their strengths, starting from zero.

\paragraph{Baselines and computation.}
Best-of-$1{,}024$ sampling returns the candidate closest to the supplied target.
Noise optimization takes Adam steps through the frozen sampler to minimize endpoint error.
FRS approximately reverses the flow from the target and samples forward from the recovered noise.
One sampler pass is a full ten-step denoising run. A backward pass through that run counts as one additional pass.
Our two gradient steps use five passes, including the initial evaluation.
The $1{,}000$-step noise search uses $2{,}001$ passes, about $400$ times as many.
These counts describe computation in the toy; \cref{tab:supp_toy2d} gives each method's success. Robot wall-clock timings are measured separately in \cref{tab:noise_search}.

\begin{table}[!htbp]
\centering
\small
\begin{tabularx}{\linewidth}{@{}Xr@{}}
\toprule
Method & Success (\%) \\
\midrule
Policy alone & 0.3 \\
Best of $1{,}024$ samples & 3.0 \\
FRS & 1.0 \\
Noise optimization, $100$ steps & 25.0 \\
Noise optimization, $1{,}000$ steps & \fst{89.7} \\
Constrained noise optimization, $300$ steps & 7.0 \\
Learned steering directions, two steps & \snd{88.0} \\
\bottomrule
\end{tabularx}
\caption{\textbf{Toy target-reaching success.} Results over $300$ calls for one trained policy. Constrained search uses $\|\varepsilon\|\le5.26$.}
\label{tab:supp_toy2d}
\end{table}

On successful noise-search calls, the median recovered noise norm is $11.6$.
For a two-dimensional standard Gaussian, $\Pr(\|\varepsilon\|>r)=\exp(-r^2/2)$, giving approximately $6\times10^{-30}$ at $r=11.6$.
Constraining the norm to $5.26$, which contains approximately $99.9999\%$ of the prior mass, yields $7.0\%$ success after $300$ steps.
The constrained and unconstrained results use different iteration budgets.
In the wider-policy variant from the toy study, with $2.7$\,cm demonstration noise and $2$\,cm policy spread, noise optimization reaches $100\%$ and learned directions reach $92\%$.
The difficulty therefore depends on the learned conditional distribution.

\subsection{Initial-noise search in robot policies}
For a fixed observation $o$, the sampler is a deterministic map $\Phi_o$ from the starting noise $\varepsilon$ to an action chunk.
We search for a target-reaching endpoint by minimizing
\begin{equation}
\varepsilon^\star = \arg\min_{\varepsilon}\, \big\lVert e\big(\Phi_o(\varepsilon)\big) - p^\star \big\rVert_2^2 ,
\label{eq:noise_search}
\end{equation}
where $e(\cdot)$ is the chunk endpoint and $p^\star$ is the per-call target.
For \pifive-LIBERO, Adam differentiates through all ten sampler steps from the call's initial noise.

\begin{figure}[!tb]
\centering
\includegraphics[width=\linewidth]{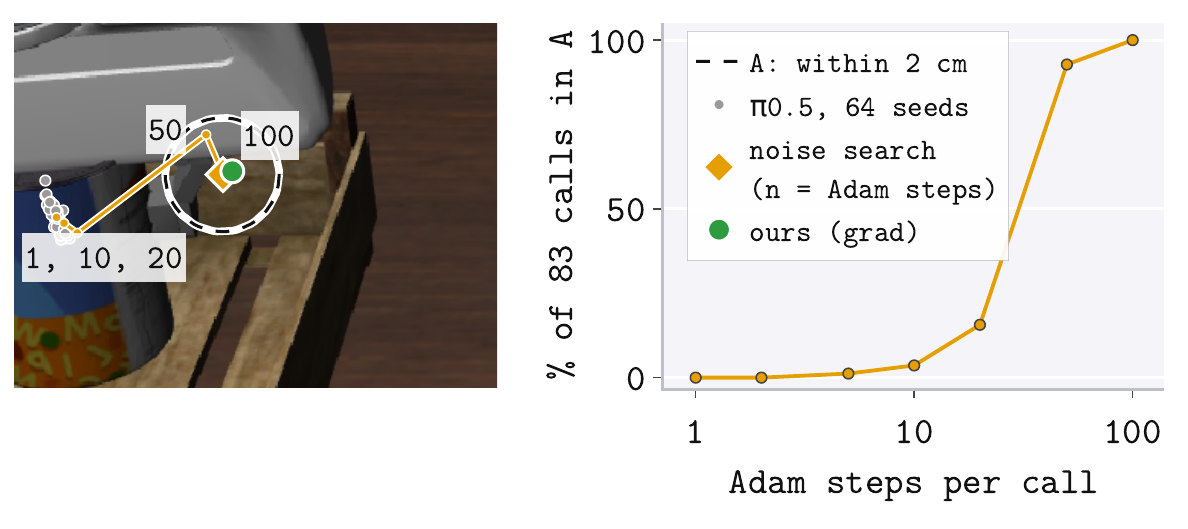}
\caption{\textbf{A rare action recovered by noise search.} Left: one call of LIBERO-90. All $64$ unsteered noise seeds end about $6$\,cm from the target (gray; ring: $2$\,cm). Noise search (orange, numbers give Adam steps) and grad (green) both reach it. Right: share of the $83$ calls whose endpoint reaches the target after a given number of Adam steps.}
\label{fig:noise_rare}
\end{figure}

\paragraph{Offline diagnostics.}
For $205$ recorded failing calls, subset simulation estimates the probability that Gaussian starting noise yields a chunk ending within $2$\,cm of the target.
The median estimated probability is about $10^{-22}$.
On the $83$ calls selected for the transfer study below, noise optimization reaches the target in $0\%$ of calls after two Adam steps, $16\%$ after $20$, $93\%$ after $50$ and $100\%$ after $100$ (\cref{fig:noise_rare}).
These calls were selected for successful grad endpoints.
The median optimized-noise norm is $37.7$, compared with $17.9$ for Gaussian draws of the same dimension.
The selected calls and the subset-simulation study use separate evaluation sets.

\begin{figure}[!htbp]
\centering
\includegraphics[width=0.88\linewidth]{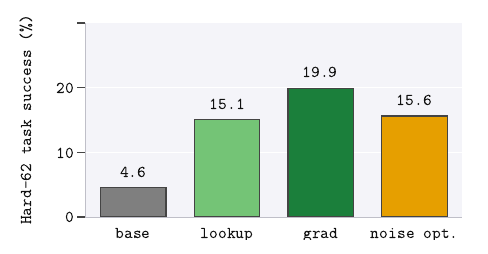}
\caption{\textbf{Noise search in closed loop.} Success on the $124$ Hard-62 scenes used in \cref{tab:noise_search}.}
\label{fig:supp_noise_search}
\end{figure}

\paragraph{Closed-loop evaluation and timing.}
All steered methods share the same target and the same rule for when to steer in this Hard-62 comparison (\cref{fig:supp_noise_search}). Noise search takes $100$ Adam steps (learning rate $0.05$) on the call's $320$ initial noise coordinates and keeps the best iterate. Grad takes two normalized steps on the direction strengths.
For base, lookup, grad and noise search, respectively, each call uses $1/0$, $3/0$, $5/2$ and $104/101$ forward/backward sampler passes. Each forward pass runs the ten-step sampler.
We interleave the four methods on the same $20$ recorded calls and report medians on the final $12$ uncontended calls. Timing runs from the transformed observation to the action chunk on the host, including the base pass for each steered method. It excludes compilation, communication and simulator time.
Learning the dictionary once for \pifive-LIBERO takes about two hours on eight RTX 4090 GPUs, excluding the subsequent effect readout.

\subsection{Transfer across scenes}
\label{sec:supp_transfer}
\begin{figure}[!htbp]
\centering
\includegraphics[width=\linewidth]{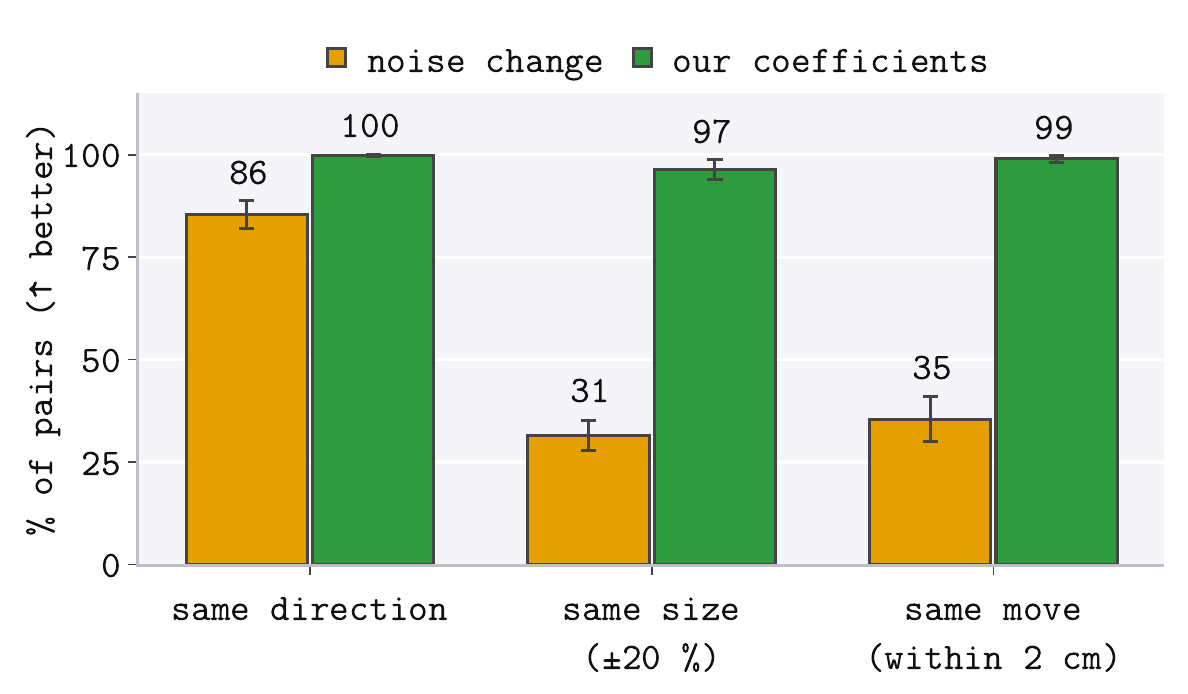}
\caption{\textbf{Transfer of endpoint displacements across tasks.} Each bar reports the share of pairs meeting the criterion below it.}
\label{fig:xscene}
\end{figure}

We use $83$ recorded calls from $27$ tasks where the base policy fails, selecting the first seed for which grad ends within $2$\,cm of the target.
For each source call, we fit a noise offset with $100$ Adam steps or direction coefficients with two gradient steps.
We apply the fitted change unchanged to recipient observations from different tasks, using the source call's starting noise for both the steered and unsteered predictions.
The recipient target is unused.
For each of $6{,}616$ ordered source-recipient pairs, we compare the induced endpoint displacement with its source-scene value.
Direction agreement requires cosine similarity above $0.9$, magnitude agreement requires a ratio in $[0.8,1.2]$, and displacement agreement requires an error below $2$\,cm.
Learned coefficients and noise offsets preserve direction in $99.9\%$ and $85.5\%$ of pairs, magnitude in $96.5\%$ and $31.4\%$, and displacement in $99.1\%$ and $35.5\%$, respectively (\cref{fig:xscene}).
Uncertainty is estimated with $2{,}000$ two-way bootstrap resamples over source and recipient tasks.
These measurements assess transfer of predicted motion on the selected calls.

\section{Direction readout and motion metrics}
\label{sec:supp_readout}

\paragraph{Measured effects.}
For each direction, we predict a path on $16$ held-out observations at strength $1$ and subtract the unsteered path generated from the same noise.
The lookup server stores the mean endpoint displacement $\vect{m}_k$ over observations.
Its usability filter requires $\|\vect{m}_k\|\ge10$\,mm and mean pairwise cosine similarity of the displacement vectors at least $0.85$.
For LIBERO, the filter is applied to commanded displacements; stored effects are then scaled for the executed prefix and controller gain.
The WAM servers read out the displacement at the scored action step directly.

\paragraph{Coverage.}
The coverage diagnostic measures the fraction of 3D unit directions within $25^\circ$ of a usable direction's mean effect.
The diagnostic implementation estimates this fraction with $2{,}000$ uniformly sampled sphere directions.
Its gain statistic is the median displacement norm over observations, and its consistency statistic is the mean cosine with the direction's mean effect.
These diagnostic statistics differ from the lookup server's mean-vector norm and pairwise consistency filter.

\begin{figure*}[!tbp]
\centering
\includegraphics[width=\linewidth]{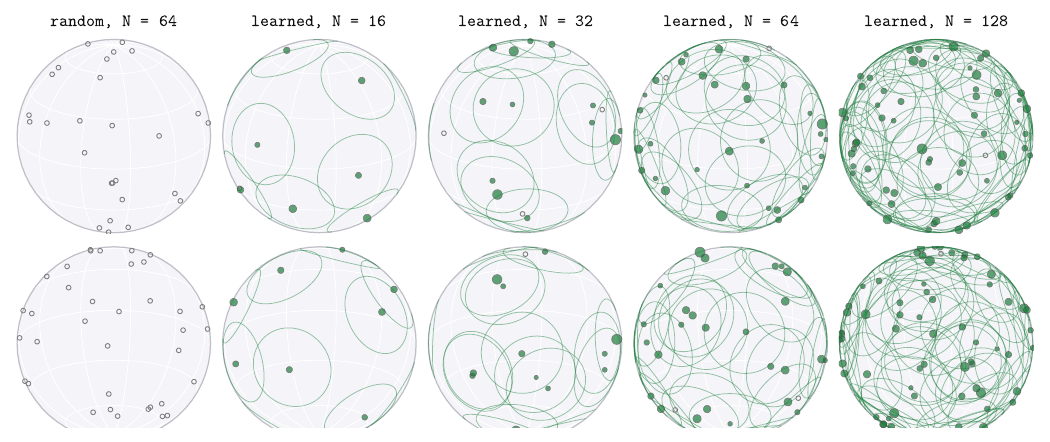}
\caption{\textbf{Directions of gripper motion on the unit sphere, \pifive-LIBERO.} Each dot is the mean endpoint displacement of one direction at strength $1$ over $16$ held-out observations, normalized to unit length. The two rows show opposite halves of the sphere. Filled dots are usable directions, with dot size increasing with the median move, and each ring bounds the directions within $25^\circ$ of one. Open dots are directions that fail the usability filter. Random directions have the norm of a learned direction; one of three draws is shown.}
\label{fig:supp_coverage}
\end{figure*}

\paragraph{Coverage of learned and random dictionaries.}
\Cref{tab:supp_coverage,fig:supp_coverage} report these diagnostics for \pifive-LIBERO.
At the norm of a learned direction, a random direction moves the gripper by a median of $2.8$\,mm, so none of the $64$ passes the $10$\,mm threshold and coverage is zero for all three random draws.
Learned directions move the gripper by $21$ to $30$\,mm at the same norm, and at least $87\%$ of them are usable for every $N$.
Coverage by a single direction grows with the dictionary, from $64\%$ at $N=16$ to $98\%$ at $N=64$ and $100\%$ at $N=128$.
A non-negative combination of usable directions reaches every sampled direction within $25^\circ$ for all four learned dictionaries.

\begin{table}[!htbp]
\centering
\small
\setlength{\tabcolsep}{4pt}
\begin{tabular}{@{}lrrrrr@{}}
\toprule
 & & & Median & \multicolumn{2}{c}{Coverage (\%)} \\
\cmidrule(l){5-6}
Directions & $N$ & Usable & move (mm) & one & mix \\
\midrule
Random & 64 & 0 & 2.8 & 0 & 0 \\
\midrule
Learned & 16 & 15 & 30.0 & 64 & 100 \\
Learned & 32 & 28 & 21.0 & 71 & 100 \\
Learned & \textbf{64} & 60 & 28.3 & 98 & 100 \\
Learned & 128 & 126 & 29.2 & 100 & 100 \\
\bottomrule
\end{tabular}
\caption{\textbf{Usable directions and coverage on \pifive-LIBERO.} Usable: directions with median move at least $10$\,mm and consistency at least $0.85$ over $16$ held-out observations at strength $1$. Coverage: share of $2{,}000$ sampled 3D directions within $25^\circ$ of one usable direction (one) or of a non-negative combination of usable directions (mix). Random directions have the norm of a learned direction, and their row averages three draws; bold marks the default $N$.}
\label{tab:supp_coverage}
\end{table}

\paragraph{Path roughness (jerk proxy).}
\label{sec:supp_smoothness}
For a predicted end-effector path with $L$ positions $\vect{p}_1,\ldots,\vect{p}_L$, we compute
\begin{equation}
J=\frac{1}{L-2}\sum_{t=1}^{L-2}
\left\|\vect{p}_{t+2}-2\vect{p}_{t+1}+\vect{p}_t\right\|_2.
\end{equation}
The score averages the norms of second position differences over the full predicted path, including any unused suffix.
It does not apply time scaling and is a path-roughness proxy, not physical jerk in $\mathrm{m}/\mathrm{s}^3$.
For each steered call, we divide the steered score by the unsteered score under the same sampling noise.
The reported table entry is the median of these ratios over steered calls.
A value above $1$ indicates greater predicted-path roughness than the unsteered policy under this metric.

\makeatletter
\setlength{\@dblfptop}{0pt}
\setlength{\@dblfpsep}{12pt}
\makeatother
\begin{table*}[t]
\section{Full benchmark results}
\label{sec:supp_benchmarks}
\noindent \Cref{tab:supp_libero_vla,tab:supp_libero_wam} give the results per policy and perturbation, and \cref{tab:supp_hard62} lists the Hard-62 tasks.\par\medskip
\centering
\small
\setlength{\tabcolsep}{3.5pt}
\begin{tabular}{@{}l cccccc cccccc@{}}
\toprule
 & \multicolumn{6}{c}{\pifive} & \multicolumn{6}{c}{\pizero} \\
\cmidrule(lr){2-7}\cmidrule(lr){8-13}
 & LIBERO & \multicolumn{5}{c}{LIBERO-PRO} & LIBERO & \multicolumn{5}{c}{LIBERO-PRO} \\
\cmidrule(lr){2-2}\cmidrule(lr){3-7}\cmidrule(lr){8-8}\cmidrule(lr){9-13}
Method & \hardsix & Swap & Object & Lang. & Task & Mean & \hardsix & Swap & Object & Lang. & Task & Mean \\
\midrule
Base policy & 5.3 & 26.6 & \snd{86.5} & \snd{96.4} & 26.0 & 58.9 & 2.2 & 5.0 & \fst{79.0} & \fst{88.1} & 21.1 & \fst{48.3} \\
+ FRS~\cite{tang2026frs} & 7.2 & 31.6 & 80.6 & 91.5 & 31.2 & 58.8 & 3.5 & \snd{8.6} & 50.3 & 49.5 & 19.9 & 32.1 \\
+ Flow Control~\cite{kao2026flowcontrol} & 7.3 & 30.4 & 85.2 & \fst{97.8} & 28.6 & \snd{60.5} & 3.9 & 6.1 & 73.9 & \snd{87.0} & \snd{23.9} & \snd{47.7} \\
+ AFI~\cite{xu2025afi} & 8.1 & 27.1 & 78.2 & 90.8 & 23.8 & 55.0 & 3.7 & 5.6 & \snd{74.0} & 83.5 & 19.2 & 45.6 \\
+ COAST~\cite{miao2026coast} & 6.5 & 26.1 & \fst{87.9} & 96.2 & 25.9 & 59.0 & 3.5 & 5.1 & 69.9 & 82.1 & 20.5 & 44.4 \\
+ \ours (lookup) & \snd{17.4} & \snd{37.4} & 71.9 & 79.9 & \snd{45.9} & 58.8 & \snd{5.8} & 8.5 & 43.2 & 38.6 & 21.4 & 27.9 \\
+ \ours (grad) & \fst{20.6} & \fst{45.6} & 73.2 & 84.1 & \fst{52.1} & \fst{63.8} & \fst{8.2} & \fst{20.5} & 64.2 & 71.6 & \fst{34.2} & \snd{47.7} \\
\bottomrule
\end{tabular}
\caption{\textbf{Task success with VLA policies.} Entries are success rates (\%) on Hard-62 and LIBERO-PRO. Both policies use the same task sets. Each policy is evaluated with the same observation pipeline and seeded noise stream across methods. AFI and COAST are our reimplementations and use their own steering rules. Dark and light green mark the best and second-best values in each column. Lookup uses a strength cap of $20$ for both policies.}
\label{tab:supp_libero_vla}
\end{table*}

\begin{table*}[t]
\centering
\scriptsize
\setlength{\tabcolsep}{3pt}
\begin{tabular}{@{}rll rll@{}}
\toprule
ID & Scene & Instruction & ID & Scene & Instruction \\
\midrule
0 & K10 & close the top drawer of the cabinet & 43 & K9 & put the white bowl on top of the cabinet \\
1 & K10 & close the top drawer of the cabinet and put the black bowl on top of it & 44 & K9 & turn on the stove \\
3 & K10 & put the butter at the back in the top drawer of the cabinet and close it & 45 & K9 & turn on the stove and put the frying pan on it \\
4 & K10 & put the butter at the front in the top drawer of the cabinet and close it & 48 & L1 & pick up the ketchup and put it in the basket \\
5 & K10 & put the chocolate pudding in the top drawer of the cabinet and close it & 49 & L1 & pick up the tomato sauce and put it in the basket \\
6 & K1 & open the bottom drawer of the cabinet & 52 & L2 & pick up the milk and put it in the basket \\
7 & K1 & open the top drawer of the cabinet & 53 & L2 & pick up the orange juice and put it in the basket \\
8 & K1 & open the top drawer of the cabinet and put the bowl in it & 58 & L3 & pick up the ketchup and put it in the tray \\
11 & K2 & open the top drawer of the cabinet & 59 & L3 & pick up the tomato sauce and put it in the tray \\
13 & K2 & put the black bowl at the front on the plate & 62 & L4 & pick up the salad dressing and put it in the tray \\
14 & K2 & put the middle black bowl on the plate & 63 & L4 & stack the left bowl on the right bowl and place them in the tray \\
15 & K2 & put the middle black bowl on top of the cabinet & 64 & L4 & stack the right bowl on the left bowl and place them in the tray \\
16 & K2 & stack the black bowl at the front on the black bowl in the middle & 65 & L5 & put the red mug on the left plate \\
17 & K2 & stack the middle black bowl on the back black bowl & 66 & L5 & put the red mug on the right plate \\
18 & K3 & put the frying pan on the stove & 69 & L6 & put the chocolate pudding to the left of the plate \\
21 & K3 & turn on the stove and put the frying pan on it & 71 & L6 & put the red mug on the plate \\
23 & K4 & close the bottom drawer of the cabinet and open the top drawer & 73 & S1 & pick up the book and place it in the front compartment of the caddy \\
25 & K4 & put the black bowl on top of the cabinet & 74 & S1 & pick up the book and place it in the left compartment of the caddy \\
26 & K4 & put the wine bottle in the bottom drawer of the cabinet & 75 & S1 & pick up the book and place it in the right compartment of the caddy \\
27 & K4 & put the wine bottle on the wine rack & 76 & S1 & pick up the yellow and white mug and place it to the right of the caddy \\
28 & K5 & close the top drawer of the cabinet & 78 & S2 & pick up the book and place it in the front compartment of the caddy \\
31 & K5 & put the black bowl on top of the cabinet & 79 & S2 & pick up the book and place it in the left compartment of the caddy \\
32 & K5 & put the ketchup in the top drawer of the cabinet & 80 & S2 & pick up the book and place it in the right compartment of the caddy \\
34 & K6 & put the yellow and white mug to the front of the white mug & 81 & S3 & pick up the book and place it in the front compartment of the caddy \\
35 & K7 & open the microwave & 83 & S3 & pick up the book and place it in the right compartment of the caddy \\
36 & K7 & put the white bowl on the plate & 84 & S3 & pick up the red mug and place it to the right of the caddy \\
37 & K7 & put the white bowl to the right of the plate & 85 & S3 & pick up the white mug and place it to the right of the caddy \\
39 & K8 & turn off the stove & 86 & S4 & pick up the book in the middle and place it on the cabinet shelf \\
40 & K9 & put the frying pan on the cabinet shelf & 87 & S4 & pick up the book on the left and place it on top of the shelf \\
41 & K9 & put the frying pan on top of the cabinet & 88 & S4 & pick up the book on the right and place it on the cabinet shelf \\
42 & K9 & put the frying pan under the cabinet shelf & 89 & S4 & pick up the book on the right and place it under the cabinet shelf \\
\bottomrule
\end{tabular}
\caption{\textbf{The Hard-62 tasks.} LIBERO-90 tasks on which the reference \pifive-LIBERO checkpoint succeeds at most $40\%$ of the time, with the task list taken from FRS~\cite{tang2026frs}. ID is the LIBERO-90 task index. K, L and S denote the kitchen, living-room and study scenes.}
\label{tab:supp_hard62}
\end{table*}

\begin{table*}[t]
\centering
\small
\setlength{\tabcolsep}{3.5pt}
\begin{tabular}{@{}l cccccc cccccc@{}}
\toprule
 & \multicolumn{6}{c}{FastWAM} & \multicolumn{6}{c}{Cosmos Policy} \\
\cmidrule(lr){2-7}\cmidrule(lr){8-13}
 & LIBERO & \multicolumn{5}{c}{LIBERO-PRO} & LIBERO & \multicolumn{5}{c}{LIBERO-PRO} \\
\cmidrule(lr){2-2}\cmidrule(lr){3-7}\cmidrule(lr){8-8}\cmidrule(lr){9-13}
Method & \hardsix & Swap & Object & Lang. & Task & Mean & \hardsix & Swap & Object & Lang. & Task & Mean \\
\midrule
Base policy & 1.3 & 0.6 & \fst{79.2} & \snd{81.2} & 14.6 & 43.9 & 2.7 & 3.2 & \fst{83.4} & 95.0 & 23.5 & \snd{51.3} \\
+ FRS~\cite{tang2026frs} & 1.6 & 1.0 & 76.6 & \fst{83.8} & 15.1 & \snd{44.1} & 2.5 & 3.0 & \snd{82.6} & \snd{95.1} & 24.0 & 51.2 \\
+ Flow Control~\cite{kao2026flowcontrol} & 1.2 & 0.9 & \snd{77.8} & 80.7 & 14.8 & 43.5 & 3.9 & 3.8 & 81.0 & 93.1 & 26.4 & 51.1 \\
+ AFI~\cite{xu2025afi} & 3.6 & 0.6 & 72.8 & 76.8 & 12.4 & 40.6 & 3.4 & 2.6 & 77.6 & 90.0 & 23.8 & 48.5 \\
+ COAST~\cite{miao2026coast} & 1.7 & 0.6 & 77.0 & 80.0 & 13.2 & 42.7 & 2.3 & 3.5 & 82.3 & \fst{95.5} & 23.5 & 51.2 \\
+ \ours (lookup) & \snd{8.1} & \snd{15.5} & 61.7 & 64.6 & \snd{26.9} & 42.2 & \snd{6.1} & \snd{13.4} & 62.1 & 71.1 & \snd{31.0} & 44.4 \\
+ \ours (grad) & \fst{10.1} & \fst{20.9} & 66.9 & 69.8 & \fst{31.8} & \fst{47.3} & \fst{10.3} & \fst{24.2} & 69.5 & 79.5 & \fst{39.5} & \fst{53.2} \\
\bottomrule
\end{tabular}
\caption{\textbf{Task success with world action models.} Entries are success rates (\%) on the same task sets and evaluation schedule as \cref{tab:supp_libero_vla}. FastWAM uses flow matching; Cosmos Policy uses EDM diffusion. The Cosmos FRS result uses an approximate reversal of the diffusion sampler. All methods use each model's native action normalization. AFI and COAST are our reimplementations and use their own steering rules. Dark and light green mark the best and second-best values per column.}
\label{tab:supp_libero_wam}
\end{table*}

\begin{table*}[!tbp]
\section{Full ablation results}
\label{sec:supp_ablations}
\begin{minipage}[t]{0.48\textwidth}
\vspace{0pt}
\centering
\small
\setlength{\tabcolsep}{3pt}
\resizebox{\linewidth}{!}{%
\begin{tabular}{@{}llc ccccc@{}}
\toprule
 & & LIBERO & \multicolumn{5}{c}{LIBERO-PRO} \\
\cmidrule(lr){3-3}\cmidrule(lr){4-8}
Selector & Directions & \hardsix & Swap & Object & Lang. & Task & Mean \\
\midrule
grad & 16 & 14.0 & 41.1 & \fst{78.2} & \fst{89.8} & 44.6 & 63.4 \\
grad & 32 & 17.0 & \snd{44.8} & \snd{75.5} & \snd{87.0} & 46.9 & \snd{63.5} \\
grad & \textbf{64} & \snd{20.6} & \fst{45.6} & 73.2 & 84.1 & \fst{52.1} & \fst{63.8} \\
grad & 128 & \fst{21.5} & 43.1 & 70.1 & 81.2 & \snd{50.7} & 61.3 \\
\midrule
lookup & 16 & 13.9 & \snd{33.4} & \snd{66.5} & \snd{78.7} & 41.4 & \snd{55.0} \\
lookup & 32 & 13.8 & 27.8 & 64.4 & 78.5 & \snd{45.1} & 53.9 \\
lookup & \textbf{64} & \fst{17.4} & \fst{37.4} & \fst{71.9} & \fst{79.9} & \fst{45.9} & \fst{58.8} \\
lookup & 128 & \snd{16.4} & 31.2 & 65.1 & 76.1 & 40.9 & 53.3 \\
\bottomrule
\end{tabular}}
\caption{\textbf{Full direction-count ablation.} Task success (\%) for \pifive-LIBERO with lookup and grad. Each dictionary is trained separately with the same temporal basis and regularizer. The PRO mean averages the four perturbation columns, each of which averages four suites. Bold marks the default count.}
\label{tab:supp_abl_dirs}
\end{minipage}\hfill
\begin{minipage}[t]{0.48\textwidth}
\vspace{0pt}
\centering
\small
\setlength{\tabcolsep}{3pt}
\resizebox{\linewidth}{!}{%
\begin{tabular}{@{}llc ccccc@{}}
\toprule
 & & LIBERO & \multicolumn{5}{c}{LIBERO-PRO} \\
\cmidrule(lr){3-3}\cmidrule(lr){4-8}
Selector & $K$ & \hardsix & Swap & Object & Lang. & Task & Mean \\
\midrule
grad & 2 & \fst{21.3} & 42.9 & 69.7 & 81.7 & \snd{50.7} & 61.3 \\
grad & \textbf{4} & \snd{20.6} & \fst{45.6} & \fst{73.2} & \fst{84.1} & \fst{52.1} & \fst{63.8} \\
grad & 8 & 20.1 & \snd{44.4} & \snd{72.6} & \fst{84.1} & 48.8 & \snd{62.5} \\
\midrule
lookup & 2 & \snd{16.3} & \fst{38.4} & \snd{70.0} & \snd{79.5} & \fst{49.1} & \fst{59.2} \\
lookup & \textbf{4} & \fst{17.4} & \snd{37.4} & \fst{71.9} & \fst{79.9} & \snd{45.9} & \snd{58.8} \\
lookup & 8 & 16.1 & 33.1 & 67.0 & 76.5 & 44.1 & 55.2 \\
\bottomrule
\end{tabular}}
\caption{\textbf{Full temporal DCT ablation.} Task success (\%) for \pifive-LIBERO. $K$ is the number of temporal cosine components per direction, including the constant component. Each row uses $64$ directions; bold marks $K=4$, the setting used in the main experiments.}
\label{tab:supp_abl_k}
\end{minipage}

\medskip
\centering
\small
\setlength{\tabcolsep}{5pt}
\begin{tabular}{@{}llrrrrrrr@{}}
\toprule
 & & LIBERO & \multicolumn{5}{c}{LIBERO-PRO} & \\
\cmidrule(lr){3-3}\cmidrule(lr){4-8}
Regularizer & Setting & \hardsix & Swap & Object & Lang. & Task & Mean & Jerk proxy $\times$ \\
\midrule
Off & None & 13.8 & 31.8 & 64.1 & \snd{76.1} & \snd{43.4} & 53.8 & 14.0 \\
\textbf{On} & \textbf{Default} & \fst{17.4} & \fst{37.4} & \fst{71.9} & \fst{79.9} & \fst{45.9} & \fst{58.8} & \snd{5.8} \\
On & Stronger & \snd{15.1} & \snd{32.4} & \snd{67.0} & 76.0 & 42.5 & \snd{54.5} & \fst{5.3} \\
\bottomrule
\end{tabular}
\caption{\textbf{Full motion-regularization ablation.} Task success (\%) for \pifive-LIBERO with lookup. Off removes the regularizer; the default and stronger settings use weights $w=5$ and $w=20$. Increasing the weight from $5$ to $20$ further lowers the smoothness proxy but reduces success on both benchmarks.}
\label{tab:supp_abl_w}

\end{table*}

\paragraph{Dictionary size and temporal components.}
A larger dictionary offers more distinct motions, but lookup applies one direction per call, so extra directions help it only if they add useful motions (\cref{tab:supp_abl_dirs}).
A larger $K$ lets a direction's offset vary more across the action steps of a chunk.
$K=2$ is ahead of our default $K=4$ by less than one point in grad's Hard-62 success and lookup's LIBERO-PRO mean, and $K=8$ gives the lowest lookup success on both benchmarks (\cref{tab:supp_abl_k}).

\clearpage
\clearpage

\begin{figure*}[p]
\begin{minipage}{\textwidth}
\centering
\includegraphics[width=\textwidth]{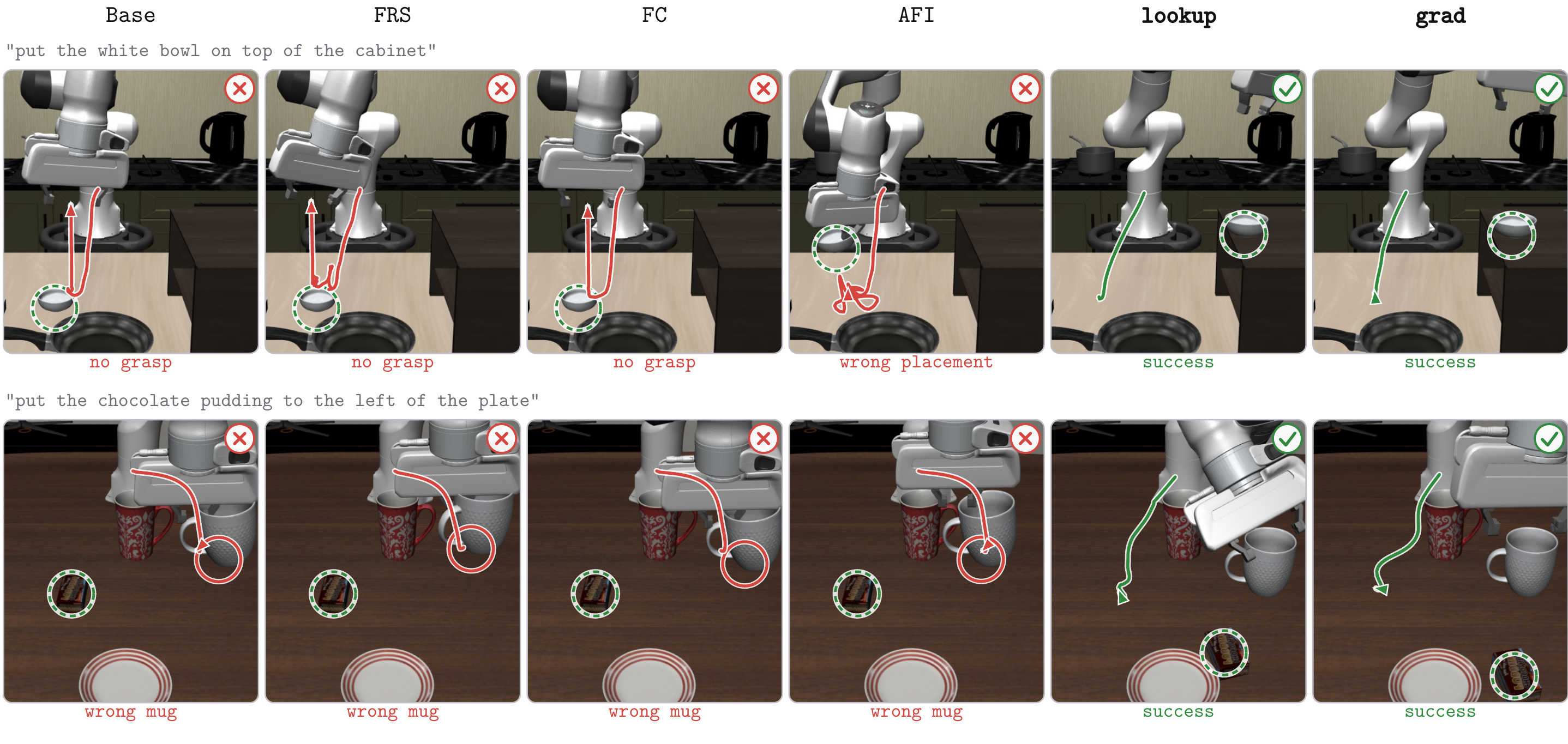}
\captionof{figure}{\textbf{FastWAM comparisons.} Two Hard-62 tasks, all methods from the same initial state. COAST is not shown: its rollouts were not re-rendered.}
\label{fig:supp_comparison_fastwam}
\end{minipage}
\medskip

\begin{minipage}{\textwidth}
\centering
\includegraphics[width=0.86\textwidth]{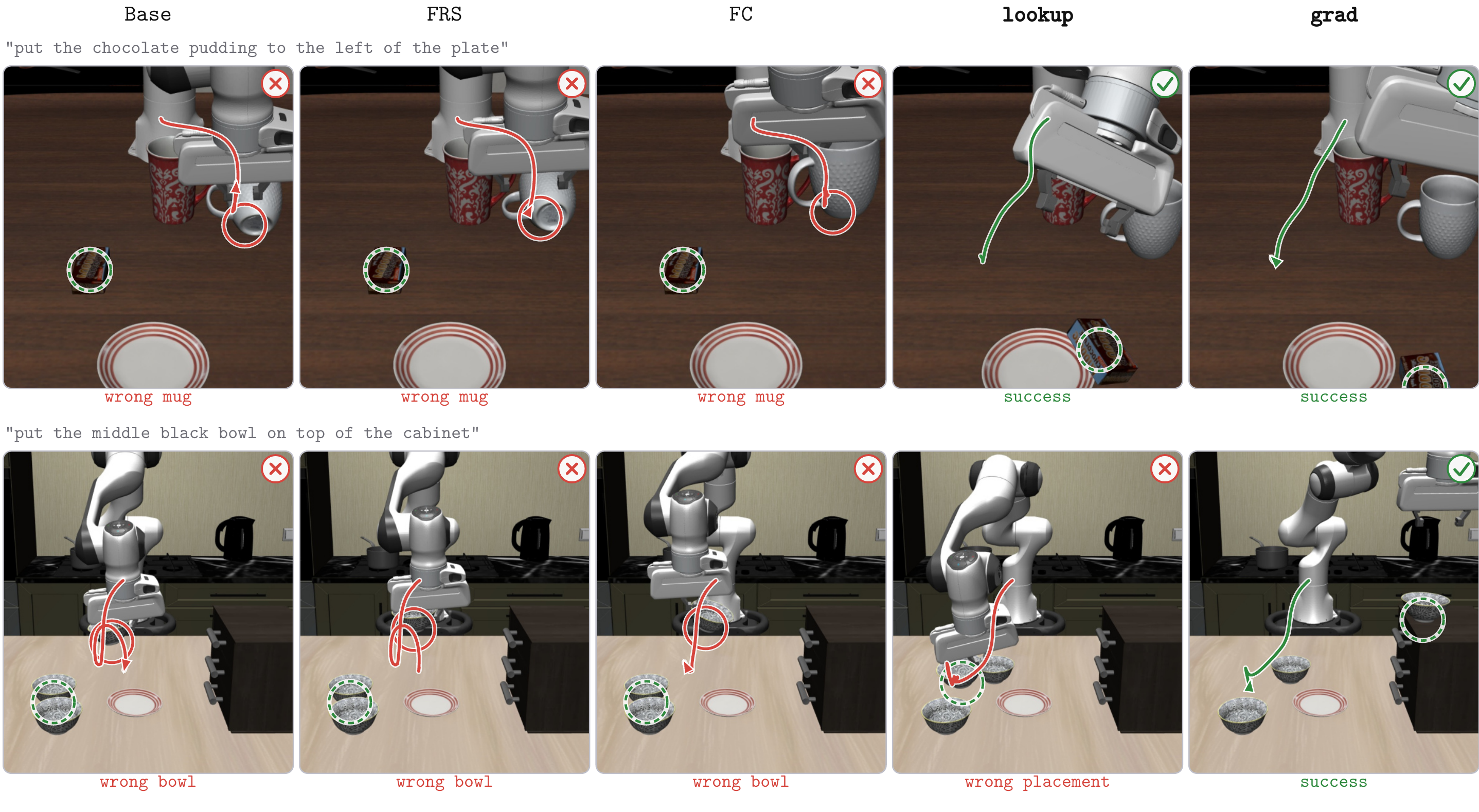}
\captionof{figure}{\textbf{Cosmos Policy comparisons.} Two Hard-62 tasks, all methods from the same initial state. AFI and COAST are not shown: their rollouts were not re-rendered.}
\label{fig:supp_comparison_cosmos}
\end{minipage}
\end{figure*}
\clearpage

\begin{figure*}[p]
\begin{minipage}{\textwidth}
\centering
\includegraphics[width=0.88\textwidth]{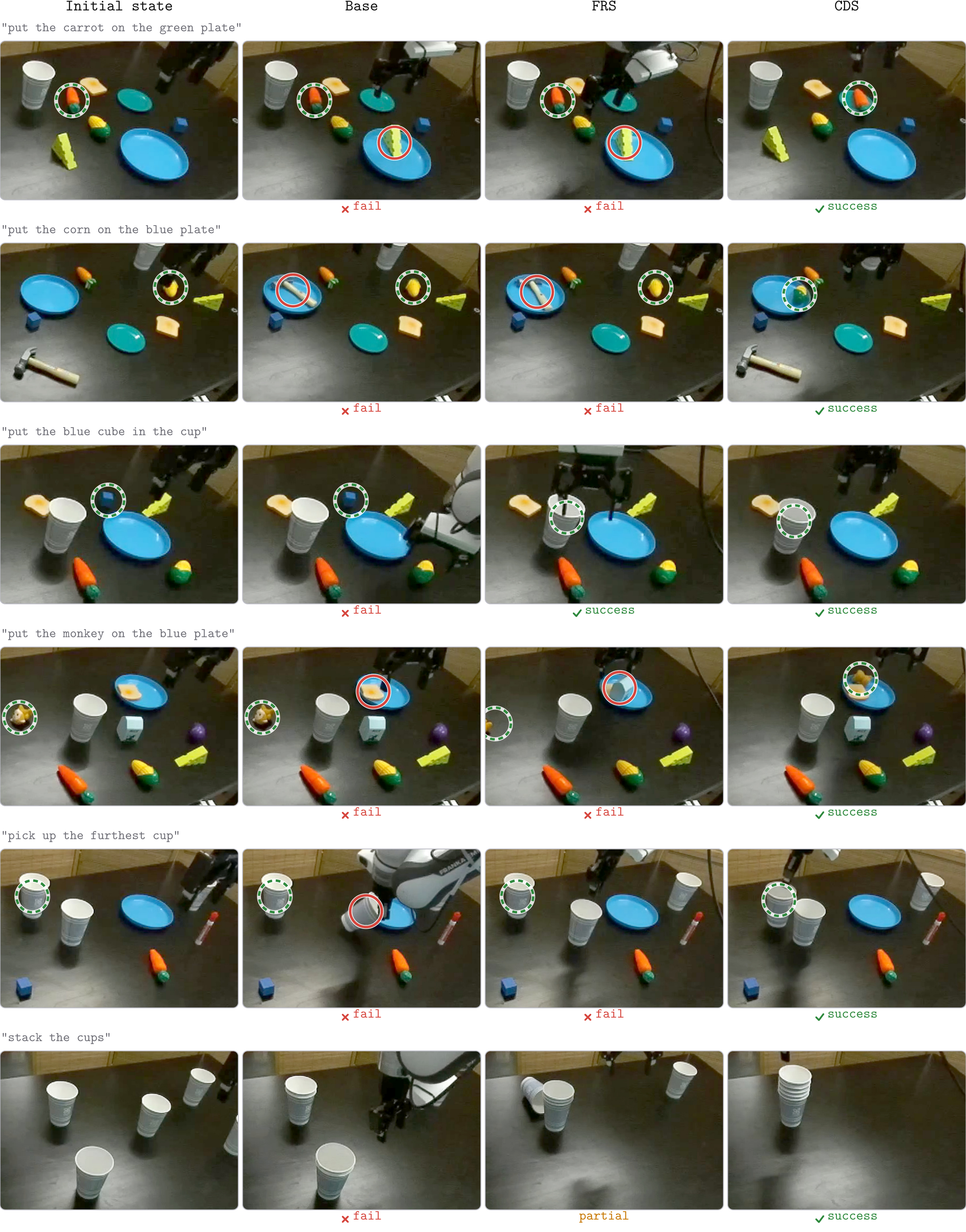}
\captionof{figure}{\textbf{Real-robot comparisons.} Six object layouts with \pifive-DROID, one per row: the initial scene, then the last frame of the run with the policy alone, with FRS and with \ours, and the outcome marked after each run (\cref{sec:supp_real}). Dashed green rings mark the instructed object, red circles a wrong object that was moved.}
\label{fig:supp_comparison_real}
\end{minipage}
\end{figure*}
\clearpage

\begin{figure*}[p]
\begin{minipage}{\textwidth}
\centering
\includegraphics[width=\textwidth]{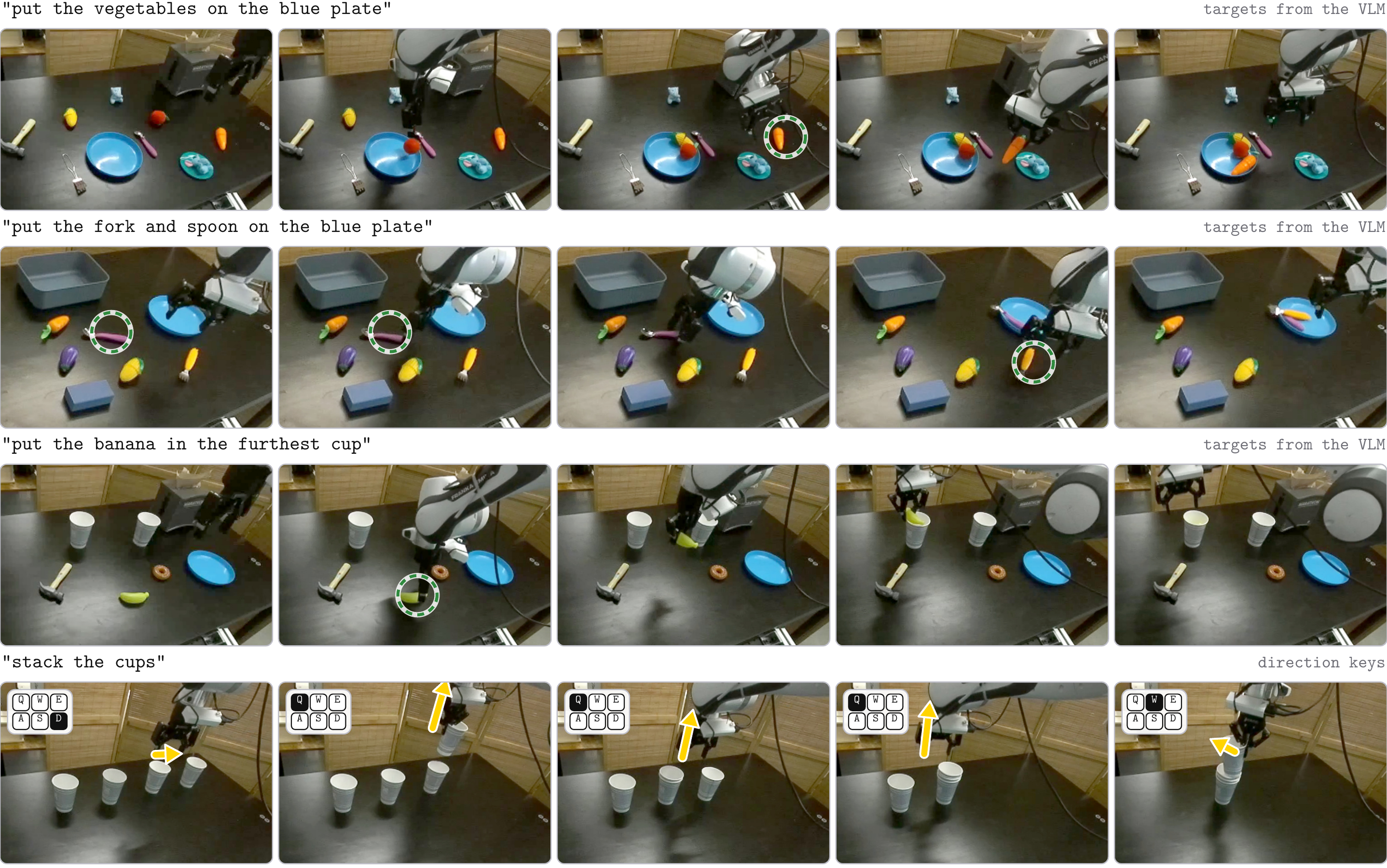}
\captionof{figure}{\textbf{\ours\ on the real robot.} Four runs with \pifive-DROID, five moments each from start to end. Dashed green rings mark the target of the current policy call when the vision-language model supplies one; the cups are stacked with direction keys, shown with the pressed key and a yellow arrow along its axis.}
\label{fig:supp_real_ours}
\end{minipage}
\end{figure*}
\clearpage

\end{document}